\documentclass[11pt,authoryear]{elsarticle}
\usepackage{cancel}
\usepackage{caption}
\usepackage{color}
\usepackage{lscape}
\usepackage{afterpage}
\usepackage{pstricks}
\usepackage{pst-plot}
\usepackage{longtable}
\usepackage{dcolumn}
\usepackage{pst-node}
\usepackage{amsmath}
\usepackage{amssymb}
\usepackage{amsthm}
\usepackage{multirow}
\usepackage{colortab}
\usepackage{color}
\usepackage{rotating}
\usepackage{array}
\usepackage{nth}
\usepackage{amsmath}
\usepackage{makecell}
\usepackage{longtable}
\usepackage{textcomp}
\usepackage{pstricks, pst-node}
\usepackage{pst-all}
\usepackage{algorithm}
\usepackage{algpseudocode}
\usepackage{url}
\usepackage{soul}
\usepackage{hyperref}
\psset{arrows=->, labelsep=3pt, mnode=circle}
\usepackage{lineno}
\usepackage{booktabs}
\usepackage{tabularx}
\usepackage{subcaption}
\usepackage{eurosym}
\usepackage{bm}
\usepackage[normalem]{ulem}
\usepackage{nth}
\usepackage{csquotes}
\usepackage{tikz}
\usetikzlibrary{decorations.pathreplacing,calligraphy}
\usetikzlibrary{shapes.geometric, arrows.meta, positioning}
\tikzstyle{startstop} = [ellipse, draw, minimum width=2.5cm, minimum height=1cm, align=center]
\tikzstyle{process} = [rectangle, draw, minimum width=2.5cm, minimum height=1cm, align=center]
\tikzstyle{decision} = [diamond, draw, aspect=2, align=center, inner sep=1pt]
\tikzstyle{arrow} = [thick, ->, >=stealth]
\usetikzlibrary{calc,positioning,fit}
\usetikzlibrary{backgrounds}
\usepackage{etoolbox}
\newcommand{\CustomState}[2]{%
  \par\noindent
  \hbox to \algorithmicindent{\hss}%
  \textbf{#1} #2\par
}

\usepackage{titlesec}

\titleformat{\paragraph}[block]
  {\normalfont\normalsize\bfseries} 
  {}{0pt}{} 
\newfont{\rams}{msbm10 scaled\magstep1}

\newenvironment{resumeT}{\begin{list}{}{\setlength{\rightmargin}{\leftmargin}}\item[]
{\centering {\bf \it~~~}
\par}\item[]\ignorespaces}{\unskip\end{list}}

\newtheorem{defn}{Definition}[section]

\usepackage{longtable}
\usepackage{lipsum}
\usepackage{titlesec}
\titleformat{\subsection}
  {\bfseries\normalsize}   
  {\thesubsection}         
  {1em}                    
  {}

\algnewcommand{\algorithmicsteps}{\textbf{Algorithm steps:}}
\algnewcommand{\Steps}{\item[\algorithmicsteps]}
\begin{document}
\title{An Explainable and Interpretable Composite Indicator Based on Decision Rules}

\author[Eco]{\rm Salvatore Corrente}
\ead{salvatore.corrente@unict.it}
\author[Eco]{\rm Salvatore Greco}
\ead{salgreco@unict.it}
\author[Poz]{\rm Roman S{\l}owi\'{n}ski}
\ead{roman.slowinski\string@cs.put.poznan.pl}
\author[Eco]{\rm Silvano Zappal\`{a}}
\ead{silvano.zappala@phd.unict.it}

\address[Eco]{Department of Economics and Business, University of Catania, Corso Italia, 55, 95129  Catania, Italy}
\address[Poz]{Institute of Computing Science, Pozna\'{n}, University of Technology, 60-965 Pozna\'{n}, and Systems Research Institute, Polish Academy of Sciences, 01-447 Warsaw, Poland}

\date{}
\maketitle

\vspace{-1cm}

\begin{resumeT}
{\large {\bf Abstract:}} Composite indicators are widely used to score or classify units evaluated on multiple criteria. Their construction typically involves aggregating criteria evaluations, a common practice in Multiple Criteria Decision Aiding (MCDA). Beyond producing a final score or classification, however, ensuring explainability, interpretability, and transparency is crucial. This paper proposes a novel framework for constructing explainable and interpretable composite indicators using if–then decision rules. We explore four scenarios: (i) decision rules explaining classifications derived from the sum of ordinal indicator codes; (ii) interpretation of an opaque numerical composite indicator used to classify units into quantiles; (iii) construction of a composite indicator from decision-maker preference information, given as classifications of reference units; and (iv) explanation of classifications generated by an existing MCDA method. To induce the rules from scored or classified units, we apply the Dominance-based Rough Set Approach. The resulting rules relate class assignments or scores to threshold conditions on indicator values in a clear and intelligible way, clarifying the underlying rationale and supporting the assessment of new units. Our main methodological contribution is the introduction of a decision-rule-based framework for constructing composite indicators. Moreover, the framework extends naturally to continuous composite indicators by treating each distinct score as an ordered class. This is enabled by a new algorithm that efficiently induces all minimal rules in a single run. Although this may yield many rules, explainability is preserved by showing only those satisfied by the unit of interest. Finally, the methodology can handle datasets with missing values, enhancing its practical applicability.
\end{resumeT}

\vspace{0,3cm}




\pagenumbering{arabic} 
\vspace{12pt}
\section{Introduction}\label{sec:Introduction}
Composite indicators (CIs) are nowadays widely used in different domains, ranging from countries' competitiveness \citep{schwab2019global} to well-being \citep{barrington2018measuring}, passing via healthcare \citep{haakenstad2022assessing} to higher education \citep{hazelkorn2015rankings} and cities' quality of life \citep{morais2011evaluation}. They assign a score to units evaluated on multiple elementary indicators corresponding to evaluation criteria \citep{ruiz2020mrp}. Relevant issues largely discussed for composite indicators are weighting and aggregation of criteria \citep{GrecoEtAl2019}. Regarding weighting, the overall assessment of the considered units heavily depends on the procedure used to assign weights to the criteria. The key questions are: who assigns these weights, and how are they determined? Very often, no weights are assigned to the different criteria, which amounts to assigning equal weights to all of them. In light of this, multiple weight vectors representing a plurality of perspectives \citep{GrecoEtAl2018}, or a different weight vector for each unit \citep{cherchye2001using}, have also been considered in the literature. With respect to the aggregation procedure, the basic problem is related to the compensatory or non-compensatory nature of the aggregation \citep{munda2009noncompensatory}, which implies the interpretation of the weights as trade-offs for compensatory approaches and coefficients of relative importance for non-compensatory approaches. A more recent discussion on the aims and scope of composite indicators has been developed, taking into account the broader context of mathematical models and quantitative evaluations in the socio-economic domain \citep{saltelli2023politics}. From this perspective, \citet{saltelli2020five} proposes five principles for ensuring methodological soundness, normative adequacy, and fairness in quantitative evaluations:

\begin{itemize}
	\item Mind the assumptions: check that the premises and basic hypotheses are reasonable within the considered domain.
	\item Mind the hubris: remember that complexity can be the enemy of relevance, and that composite indicators displaying an impressive number of elementary indicators often depend on a much smaller subset.
	\item Mind the framing: match purpose and context, ensuring sufficient transparency to evaluate both technical soundness and fairness.
	\item Mind the consequences: consider that quantification may backfire; provide full explanations to allow discussion and possible deconstruction of measurements, especially in relation to unintended or malicious effects.
	\item Mind the unknowns: acknowledge ignorance, inaccurate determinations, uncertainty, and imprecision, and avoid “quantifying at all costs”.
\end{itemize}

We address the above requirements by proposing a new methodology for constructing composite indicators based on logical ``\textit{if…, then…}'' decision rules, rather than the conventional numerical assessments obtained through aggregating elementary indicators. To the best of our knowledge, this is the first methodological proposal for decision-rule-based composite indicators. In our framework, decision rules provide both an explanation and an interpretation of the judgments produced by the composite indicators. This makes the approach a form of \enquote{glass box assessment}, in contrast to the common practice of \enquote{black box scoring}.
Our proposal contributes to the current debate on explainability and interoperability of Artificial Intelligence (AI) recommendations  \citep{christoph2020interpretable,rudin2019stop}. The idea is that most AI models are ``black boxes'', offering no explanation for how their outputs were generated. Explainability and interpretability have distinct meanings: explainability, typically associated with a post-hoc approach, refers to the ability to understand why an AI model produces a given output; interpretability, associated with an ante-hoc approach, concerns the comprehension of how a model generates its recommendations \citep{rudin2019stop,ali2023explainable}. 

We apply the concepts of post-hoc explainability and ante-hoc interpretability to composite indicators using a decision rule induction methodology. This approach expresses the relationships between elementary indicators (equivalent to evaluation criteria) and the composite indicator through decision rules of the form ``\textit{if…, then…}'', formulated in natural language. For example: \enquote{If country $a$ has a mean number of years of schooling not less than $11$, and Gross National Income (GNI) per capita not less than $3200$ dollars, then $a$ has at least medium human development}; or \enquote{If country $a$ has a life expectancy at birth not greater than $59$ years, and mean years of schooling not greater than $5$, then $a$ has low human development}.

In particular, we apply dominance-based decision rules derived through the Dominance-based Rough Set Approach (DRSA) \citep{GrecoMatarazzoSlowinski2001}. DRSA extends the original Rough Set concept \citep{pawlak1982rough, Pawlak1991} - initially proposed as a foundation for data reasoning methodologies - to the field of Multiple Criteria Decision Aiding (MCDA) (for a comprehensive collection of state-of-the-art surveys see \citealt{GrecoEhrgottFigueira2016}, while for a detailed discussion on a comprehensive taxonomy of MCDA problems and methods see \citealt{CinelliEtAl2020}). In a post-hoc explainability approach, decision rules are induced to justify the overall assessment provided by a composite indicator in terms of performance on specific criteria. Conversely, in an ante-hoc interpretability approach, the composite indicator is constructed by applying decision rules induced from overall evaluations elicited from the subject responsible for producing the indicator. The contribution that goes beyond a simple application of DRSA is an original procedure for selecting a minimal subset of decision rules explaining the multicriteria assessment without contradictions. While the DRSA literature has extensively explored decision rule induction, it has not addressed the problem of finding a consistent and minimal subset of decision rules from all possible decision rules derived from classification data. By consistency, we mean classification avoiding situations where the same unit is simultaneously assigned to at least a given class (e.g., `good') and to at most a lower class (e.g., `medium'), which would be clearly unacceptable.

The methodology we are proposing gives an adequate answer to the discussions on weighting and aggregation of criteria. Indeed, decision rules do not use any weight vector and do not apply any more or less compensatory aggregation procedure. Below, we explain how our decision-rule-based methodology answers the five requirements mentioned above for ensuring soundness, adequacy, and fairness.

\begin{itemize}
	\item Decision rules help to mind the assumptions, as they can be interpreted as basic ``cause-effect'' scenarios explaining the overall evaluation provided by the composite indicator.
	\item Decision rules help to mind the hubris, because even if a large number of elementary indicators is used, the explanation remains parsimonious, relying on a small set of meaningful indicators that actually drive the overall evaluation.
	\item Decision rules help to mind the framing, as their easily understandable syntax makes the reasoning behind the composite indicator fully transparent, allowing users to assess its soundness and fairness.
	\item Decision rules help to mind the consequences, as they provide explanations that support critical analysis of the composite indicator's evaluations.
	\item Decision rules help to mind the unknowns, since they do not require each elementary indicator to be expressed as a real number, and they allow for the inclusion of qualitative or ordinal evaluations; moreover, they can handle ``missing values'' in the classification data.
\end{itemize}

We can also say that our approach proposes a conception of composite indicators that goes beyond the mere assignment of numerical scores, shifting the focus to the reasons that justify the evaluations assigned to each unit, thereby addressing demands for transparency and accountability \citep{kuc2020quantitative}. 

We consider four scenarios for our approach: 
\begin{itemize}
	\item[(i)] decision rules explain the classification that results from partitioning the sum of numerical codes of elementary ordinal indicators;
	\item[(ii)] an ``obscure'' numerical composite indicator  is explained using decision rules; 
	\item[(iii)] given preference information provided by a Decision Maker (DM) in the form of classifications of some reference units (e.g., unit $a$ is assigned to Class $C_1$, while unit $b$ is assigned to Class $C_2$), a composite indicator is constructed using decision rules; 
    \item[(iv)] the classification of a set of units results from the application of an MCDA method such as ELECTRE-Score \citep{FigueiraGrecoRoy2022} (for a recent survey on MCDA with a perspective on the future developments see \citealt{greco2024fifty}), and is explained by decision rules. 
\end{itemize}
Observe that scenarios (i) and (iii) are ante-hoc interpretability-oriented, while scenarios (ii) and (iv) are post-hoc explainability-oriented.
We also note that the proposed methodology applies not only to discrete settings but also to continuous composite indicators and datasets with missing values. Explainability on demand is ensured by presenting the rules satisfied by the unit under analysis, while the generation of all possible rules remains computationally efficient thanks to a new algorithm that induces all minimal rules in a single run.

The paper is organized as follows.
Section \ref{sec:examples_intro} illustrates aggregation through composite indicators and the application of our approach across the four above mentioned scenarios, with supporting examples. Section \ref{sec:Background} presents the methodological foundations of the proposed framework and introduces a new algorithm that induces all decision rules in a single run. This algorithm efficiently handles continuous composite indicators by treating each distinct score as an ordered class. Section \ref{sec:algorithm} focuses on scenario (iv), discussing in detail the construction of decision-rule-based composite indicators from an ante-hoc interpretability perspective.
Section \ref{sec:Newalternative} describes how to classify new units and explain their classification. Section \ref{sec:Missing-value} discusses how the proposed rule induction algorithm can be extended to handle datasets with missing values, enabling the construction of composite indicators even in such cases in a natural and consistent way.  Section \ref{sec:Conclusions} concludes with a summary and a discussion of potential future developments.

\section{Multiple criteria evaluation of units and their aggregation by Composite Indicators}
\label{sec:examples_intro}
Let us consider a finite set of units $A$ evaluated across a set of elementary indicators interpreted as a consistent family of criteria $G=$$\left\{g_1,\ldots,g_m\right\}$ \citep{Roy1996}. The evaluation scales of criteria may be ordinal, interval, or ratio. In all cases, they take numerical values. A composite indicator $CI$ is a function aggregating the performances of unit $a \in A$ on criteria from $G$ to give a global numerical score to $a$. 

Denoting by $X_{j}$, $j=1,\ldots,m$, the set of possible evaluations unit $a$ can take on $g_j$, and by $g_j(a)$ the evaluation of $a$ on $g_j\in G$, a composite indicator is formally defined as a function   
$$
CI:X_1\times\cdots\times X_m\rightarrow\mathbb{R}
$$
so that $CI\left(g_1(a),\ldots,g_m(a)\right)$ is a score of $a\in A$ considering its performance on considered criteria. 

The way the performances of $a$ are aggregated to compute $CI\left(g_1(a),\ldots,g_m(a)\right)$ depends on several aspects, such as the degree of allowed compensation between criteria, the presence of interactions among criteria, and others. The resulting score can then be used to classify the units into quality classes, for example, by using some quantiles of the score distribution.

Independently on the type of aggregation used, an important aspect is the explainability and interpretability of the scoring, so that the user understands the relationship between the performances of a unit on particular criteria and its global score. In the following, we shall explain what we mean by an explainable and interpretable composite indicator in four possible scenarios, using some realistic examples.

\subsection{Scenario (i): Decision rules explain the classification that results from partitioning the sum of numerical codes of elementary ordinal indicators}
Typical examples of numerical scores obtained from an aggregation of numerical codes corresponding to elementary ordinal indicators are widely used medical scales. They are standardized tools used to assess and quantify various medical conditions, symptoms, or patient states. Very often these composite indicators aggregate the results of medical tests by simply summing numerical codes assigned to various test outcomes. They are intended to help healthcare professionals make quick decisions about patient conditions, for example, in a hospital emergency room \citep{chakraborty2023comprehensive}.

Let us consider a popular medical scale, the Glasgow Coma Scale (GCS) \citep{TeasdaleJennett1976}. It assesses a patient's level of consciousness after a brain injury based on the results of three tests: Eye Opening Response, Best Verbal Response, and Best Motor Response. Each test is evaluated on a different qualitative scale, as shown in Table \ref{tab:GCSscale}, and these assessments are then translated into numerical codes. The lower the code value, the more severe is the test response.
\begin{table}[h!]
    \centering
    \caption{Test's score of the Glasgow Comma Scale}
    \label{tab:GCSscale}
    \begin{tabular}{c|ccc}
    \toprule
    Numerical code&Eye opening&Verbal Response&Motor Response\\
    \midrule
    6&&&Obeys commands\\
    5&&Oriented&Localizes pain\\4&Spontaneous&Confused conversation&Withdraws from pain\\
    3&To sounds&Inappropriate words&Abnormal flexion\\
    2&To pain&Incomprehensible sounds&Abnormal extension\\1&None&None&None\\
    \bottomrule
    \end{tabular}
\end{table}

On the basis of the numerical grades given to the patient on the three tests, a final score (GCS score) equal to their sum is computed. Depending on the range of the GCS score, a final diagnosis about the severity of the patient's state is classified as shown in Table \ref{tab:example_second}.

\begin{table}[!h]
\caption{GCS score ranges and severity classes}
\label{tab:example_second}
\centering
\begin{tabular}{lll}
\toprule
GCS score & Severity class& Clinical implications\\
\midrule
13-15 & Mild & Observation, possible CT scan \\
9-12 & Moderate & Hospital admission, neurological monitoring \\
$\leqslant 8$ & Severe & ICU admission, possible intubation, high mortality risk\\
\bottomrule
\end{tabular}
\end{table}

Calculating the GCS score by summing numerical test grades assigns equal importance to all tests and introduces a compensatory effect among them. Consequently, a unit gain in one test compensates a unit loss in another. When medical personnel receive the global score and its numerical components, there is no clear explanation for why the patient is assigned to a specific severity class. We want to improve this situation by associating with the GCS score a verbal rule explanation.

In order to clarify our approach to explanatory composite indicators in the case of GCS, let us consider a set of patients assessed by medical doctors on the three tests, as presented in Table \ref{tab:GCSexample_evaluations}. The performances are not based on real patient data but were randomly generated for demonstration purposes.  Using the GCS scoring procedure, these patients were classified to the three classes of severity, displayed in the last column of the same table.
\begin{table}[h!]
	\centering
	\caption{Patients’ scores on the three GCS tests and DM’s classification. In the first three columns, the numerical codes of the test assessments (see Table \ref{tab:GCSscale}) are reported in parentheses. In the last column, the corresponding total score is shown in parentheses, translated into a severity class as described in Table \ref{tab:example_second}}
	\label{tab:GCSexample_evaluations}
	\begin{tabular}{cllll}
		\toprule
		Unit&Eye Opening&Verbal Response& Motor Response& Severity class\\
		\midrule
        $a_{1}$ & To pain (2) & Incomprehensible sounds (2) & Abnormal flexion (3) & Severe (7) \\ 
    $a_{2}$ & To sounds (3) & Confused conversation (4) & Localizes pain (5) & Moderate (12) \\ 
    $a_{3}$ & Spontaneous (4) & Inappropriate words (3) & Withdraws from pain (4) & Moderate (11) \\ 
    $a_{4}$ & To pain (2) & Incomprehensible sounds (2) & Abnormal extension (2) & Severe (6) \\ 
    $a_{5}$ & To sounds (3) & Incomprehensible sounds (2) & Abnormal extension (2) & Severe (7) \\ 
    $a_{6}$ & To pain (2) & Confused conversation (4) & Withdraws from pain (4) & Moderate (10) \\ 
    $a_{7}$ & Spontaneous (4) & Confused conversation (4) & Localizes pain (5) & Mild (13) \\ 
    $a_{8}$ & Spontaneous (4) & Inappropriate words (3) & Obeys commands (6) & Mild (13)  \\ 
    $a_{9}$ & To sounds (3) & Inappropriate words (3) & Withdraws from pain (4) & Moderate (10) \\ 
    $a_{10}$ & Spontaneous (4) & Inappropriate words (3) & Abnormal flexion (3) & Moderate (10) \\ 
    $a_{11}$ & Spontaneous (4) & Oriented (5) & Localizes pain (5) & Mild (14) \\ 
    $a_{12}$ & Spontaneous (4) & Oriented (5) & Obeys commands (6) & Mild (15) \\ 
    $a_{13}$ & To pain (2) & Inappropriate words (3) & Abnormal extension (2) & Severe (7) \\ 
    $a_{14}$ & To sounds (3) & Oriented (5) & Localizes pain (5) & Mild (13) \\ 
    $a_{15}$ & None (1) & Incomprehensible sounds (2) & Abnormal extension (2) & Severe (5) \\ 
    $a_{16}$ & To pain (2) & None (1) & Abnormal extension (2) & Severe (5) \\ 
    $a_{17}$ & To pain (2) & Inappropriate words (3) & Withdraws from pain (4) & Moderate (9) \\ 
    $a_{18}$ & Spontaneous (4) & Confused conversation (4) & Obeys commands (6) & Mild (14) \\ 
    $a_{19}$ & To pain (2) & Inappropriate words (3) & Localizes pain (5) & Moderate (10) \\ 
    $a_{20}$ & None (1) & None (1) & Abnormal extension (2) & Severe (4) \\ 
    $a_{21}$ & Spontaneous (4) & Oriented (5) & Localizes pain (5) & Mild (14) \\ 
    \midrule
    $a_{22}$& To pain (2) & Confused conversation (4) & Localizes pain (5) & ?  \\
    \bottomrule
    \end{tabular}
\end{table}
This decision table is a starting point for the generation of all possible \textit{\enquote{if\ldots, then\ldots}} decision rules using DRSA. The number of all possible rules is equal to $17$. A minimal subset of these rules covering all patients from Table \ref{tab:GCSexample_evaluations} is the following:\\[6mm]
\textbf{\enquote{at-least} rules}
\begin{enumerate}
    \item If \textit{Verbal Response} is not worse than \enquote{Inappropriate words}, and \textit{Motor Response} is not worse than \enquote{Abnormal flexion}, then the patient is assigned to a class not worse than \enquote{Moderate}; supported by classification examples: $\left\{a_{2}\right.$ $a_{3}$, $a_{6}$, $a_{7}$, $a_{8}$, $a_{9}$, $a_{10}$, $a_{11}$, $a_{12}$, $a_{14}$, $a_{17}$, $a_{18}$, $a_{19}$, $\left.a_{21}\right\}$,
    \item If \textit{Verbal Response} is not worse than \enquote{Oriented}, then the patient is assigned to a class not worse than \enquote{Mild}; supported by classification examples:  $\left\{a_{11}\right.$, $a_{12}$, $a_{14}$, $\left.a_{21}\right\}$,
    \item If \textit{Eye opening} is not worse than \enquote{Spontaneous}, and \textit{Motor Response} is not worse than \enquote{Localizes pain}, then the patient is assigned to a class not worse than \enquote{Mild}; supported by classification examples: $\left\{a_{7}\right.$, $a_{8}$, $a_{11}$, $a_{12}$, $a_{18}$, $\left.a_{21}\right\}$.
\end{enumerate}
\textbf{\enquote{at-most} rules}
\begin{enumerate}
    \item If \textit{Verbal Response} is not better than \enquote{Inappropriate words}, and \textit{Motor Response} is not better than \enquote{Localizes pain}, then the patient is assigned to a class not better than \enquote{Moderate}; supported by classification examples: $\left\{a_{1}\right.$, $a_{3}$, $a_{4}$, $a_{5}$, $a_{9}$, $a_{10}$, $a_{13}$, $a_{15}$, $a_{16}$, $a_{17}$, $a_{19}$, $\left.a_{20}\right\}$,
    \item If \textit{Eye opening} is not better than \enquote{To sounds}, and \textit{Verbal Response} is not better than \enquote{Confused conversation}, then the patient is assigned to a class not better than \enquote{Moderate}; supported by classification examples: $\left\{a_{1}\right.$, $a_{2}$, $a_{4}$, $a_{5}$, $a_{6}$, $a_{9}$, $a_{13}$, $a_{15}$, $a_{16}$, $a_{17}$, $a_{19}$, $\left.a_{20}\right\}$,
    \item If \textit{Motor Response} is not better than \enquote{Abnormal Extension}, then the patient is assigned to a class not better than \enquote{Severe}; supported by classification examples: $\left\{a_{4}\right.$, $a_{5}$, $a_{13}$, $a_{15}$, $a_{16}$, $\left.a_{20}\right\}$,
    \item If \textit{Eye opening} is not better than \enquote{To pain}, and \textit{Motor Response} is not better than \enquote{Abnormal flexion}, then the patient is assigned to a class not better than \enquote{Severe}; supported by classification examples: $\left\{a_{1}\right.$, $a_{4}$, $a_{13}$, $a_{15}$, $a_{16}$, $\left.a_{20}\right\}$.
\end{enumerate}

Let us assume now that a new patient, $a_{22}$, answering to the three tests as shown in the last row of Table \ref{tab:GCSexample_evaluations}, has to be evaluated. One can see which rules generated from Table \ref{tab:GCSexample_evaluations} are satisfied by them. In this case, from the whole set of possible rules, the following rules are satisfied by patient $a_{22}$:\\[1mm]
\textbf{\enquote{at-least} rules}:
\begin{enumerate}
    \item If \textit{Motor Response} is not worse than \enquote{Withdraws from pain}, then the patient is assigned to a class not worse than \enquote{Moderate},
    \item If \textit{Verbal Response} is not worse than \enquote{Confused conversation}, then the patient is assigned to a class not worse than \enquote{Moderate},
    \item If \textit{Verbal Response} is not worse than \enquote{Inappropriate words}, and \textit{Motor Response} is not worse than \enquote{Abnormal flexion}, then the patient is assigned to a class not worse than \enquote{Moderate}.
\end{enumerate}
\textbf{\enquote{at-most} rules}:
\begin{enumerate}
    \item If \textit{Eye opening} is not better than \enquote{To pain}, then the patient is assigned to a class not better than \enquote{Moderate},
    \item If \textit{Eye opening} is not better than \enquote{To sounds}, and \textit{Verbal response} is not better than \enquote{Confused conversation}, then the patient is assigned to a class not better than \enquote{Moderate}.
\end{enumerate}
Taking into consideration the minimal set of rules covering patients from Table \ref{tab:GCSexample_evaluations}, the following rules are satisfied by patient $a_{22}$:\\[1mm]
\textbf{\enquote{at-least} rules}:
\begin{enumerate}
    \item If \textit{Verbal response} is not worse than \enquote{Inappropriate words}, and \textit{Motor response} is not worse than \enquote{Abnormal flexion}, then the patient is assigned to a class not worse than \enquote{Moderate}.
\end{enumerate}
\textbf{\enquote{at-most} rules}:
\begin{enumerate}
    \item If \textit{Eye opening} is not better than \enquote{To sounds}, and \textit{Verbal response} is not better than \enquote{Confused conversation}, then the patient is assigned to a class not better than \enquote{Moderate}.
\end{enumerate}

The intersection of class assignments of these rules indicates the class \enquote{Moderate} for $a_{22}$. The rules satisfied by the analysed patient reveal, in an easily interpretable way, which conditions on the test results lead to the suggested class assignment.

\subsection{Scenario (ii): Explaining an obscure numerical composite indicator using decision rules}\label{sec:HSDIExample}
As an example of an obscure CI, we take the Human Development Index \citep{HDR}; for an interesting MCDA approach, see \citep{despotis2005measuring}. It is a CI contained in the Human Development Report (HDR) published for the first time in 1990. This index seeks to offer a comprehensive assessment of 193 countries worldwide across three key dimensions: Health, Education, and Standard of living. These dimensions are further represented by four specific indicators: Health is assessed using  Life expectancy at birth (years) ($g_1$); Education is evaluated through Expected years of schooling (years) ($g_2$) and Mean years of schooling (years) ($g_3$); and Standard of living is measured by Gross National Income (GNI) per capita (2017 PPP\$) ($g_4$). Once the minimum and maximum values are established for each criterion, the evaluations are standardized using min-max normalization. For the Education dimension, each unit's score is calculated as the average of the normalized values of its two associated indicators ($g_2$ and $g_3$). In the case of GNI, normalization is applied to the logarithm of the original values. The final score for each unit is then derived by taking the geometric mean of the normalized scores across the three dimensions. Based on the score, units are assigned to four different human development classes depending on the quartile they belong to: Class 4 - Very high human development (if their score is in the fourth quartile), Class 3 - High human development (if their score is in the third quartile), Class 2 - Medium human development (if their score is in the second quartile), Class 1 - Low human development (if their score is in the first quartile) (the technical notes on the HDI construction can be found at the following \href{https://hdr.undp.org/sites/default/files/2023-24_HDR/hdr2023-24_technical_notes.pdf}{link}).

With this example, we aim to illustrate how DRSA can easily explain the classification of the considered units based on their evaluations on the four criteria at hand. The data, drawn from 2023–2024, are presented in Table \ref{tab:hdi_index}. They can be downloaded from the HDR website via the following \href{https://hdr.undp.org/sites/default/files/2023-24_HDR/HDR23-24_Statistical_Annex_HDI_Table.xlsx?utm_source=chatgpt.com}{link}. By applying the DRSA, one gets a total of $83$ \enquote{at-least} and $90$ \enquote{at-most} rules. The same classifications can be explained by the following minimal subsets of $17$ \enquote{at-least} and $14$ \enquote{at-most} rules: \\[2mm]
\textbf{\enquote{at-least} rules}
\begin{enumerate}
    \item If $g_{2}(a)\geqslant11.018$, and $g_{4}(a)\geqslant3157.3586$, then $a$ is assigned to at least Class 2 ($d_{1}^{\geqslant} =2$),
    \item If $g_{1}(a)\geqslant58.059$, and $g_{3}(a)\geqslant7.2435$, then $a$ is assigned to at least Class 2 ($d_{2}^{\geqslant} =2$),
    \item If $g_{1}(a)\geqslant61.929$, and $g_{3}(a)\geqslant5.8443$, then $a$ is assigned to at least Class 2 ($d_{3}^{\geqslant} =2$),
    \item If $g_{1}(a)\geqslant63.728$, and $g_{3}(a)\geqslant5.6095$, and $g_{4}(a)\geqslant2801.7125$, then $a$ is assigned to at least Class 2 ($d_{4}^{\geqslant} =2$),
    \item If $g_{2}(a)\geqslant14.6195$, and $g_{4}(a)\geqslant10296.6496$, then $a$ is assigned to at least Class 3 ($d_{5}^{\geqslant} =3$),
    \item If $g_{2}(a)\geqslant13.0529$, and $g_{4}(a)\geqslant10813.9827$, then $a$ is assigned to at least Class 3 ($d_{6}^{\geqslant} =3$),
    \item If $g_{2}(a)\geqslant12.403$, and $g_{3}(a)\geqslant10.5207$, then $a$ is assigned to at least Class 3 ($d_{7}^{\geqslant} =3$),
    \item If $g_{2}(a)\geqslant12.4598$, and $g_{3}(a)\geqslant9.2482$, and $g_{4}(a)\geqslant9694.5228$, then $a$ is assigned to at least Class 3 ($d_{8}^{\geqslant} =3$),
    \item If $g_{1}(a)\geqslant65.913$, and $g_{4}(a)\geqslant14841.5784$, then $a$ is assigned to at least Class 3 ($d_{9}^{\geqslant} =3$),
    \item If $g_{1}(a)\geqslant71.294$, and $g_{3}(a)\geqslant8.577$, then $a$ is assigned to at least Class 3 ($d_{10}^{\geqslant} =3$),
    \item If $g_{1}(a)\geqslant70.962$, and $g_{3}(a)\geqslant8.8476$, and $g_{4}(a)\geqslant9242.0823$, then $a$ is assigned to at least Class 3 ($d_{11}^{\geqslant} =3$),
    \item If $g_{3}(a)\geqslant11.5048$, and $g_{4}(a)\geqslant19494.0089$, then $a$ is assigned to at least Class 4 ($d_{12}^{\geqslant} =4$),
    \item If $g_{1}(a)\geqslant70.116$, and $g_{4}(a)\geqslant26991.8496$, then $a$ is assigned to at least Class 4 ($d_{13}^{\geqslant} =4$),
    \item If $g_{1}(a)\geqslant73.246$, and $g_{3}(a)\geqslant12.2412$, then $a$ is assigned to at least Class 4 ($d_{14}^{\geqslant} =4$),
    \item If $g_{1}(a)\geqslant76.996$, and $g_{2}(a)\geqslant15.9345$, then $a$ is assigned to at least Class 4 ($d_{15}^{\geqslant} =4$),
    \item If $g_{1}(a)\geqslant79.236$, and $g_{2}(a)\geqslant15.5123$, then $a$ is assigned to at least Class 4 ($d_{16}^{\geqslant} =4$),
    \item If $g_{1}(a)\geqslant71.587$, and $g_{2}(a)\geqslant16.7268$, then $a$ is assigned to at least Class 4 ($d_{17}^{\geqslant} =4$).
\end{enumerate}
\textbf{\enquote{at-most} rules}
\begin{enumerate}
    \item If $g_{4}(a)\leqslant14778.3463$, then $a$ is assigned to at most Class 3 ($d_{1}^{\leqslant} =3$),
    \item If $g_{2}(a)\leqslant15.2179$, and $g_{4}(a)\leqslant18024.8875$, then $a$ is assigned to at most Class 3 ($d_{2}^{\leqslant} =3$),
    \item If $g_{2}(a)\leqslant14.6101$, and $g_{3}(a)\leqslant9.992$, and $g_{4}(a)\leqslant23251.6207$, then $a$ is assigned to at most Class 3 ($d_{3}^{\leqslant} =3$),
    \item If $g_{1}(a)\leqslant69.41$, then $a$ is assigned to at most Class 3 ($d_{4}^{\leqslant} =3$),
    \item If $g_{1}(a)\leqslant71.674$, and $g_{3}(a)\leqslant11.9111$, then $a$ is assigned to at most Class 3 ($d_{5}^{\leqslant} =3$),
    \item If $g_{3}(a)\leqslant9.8278$, and $g_{4}(a)\leqslant7987.8422$, then $a$ is assigned to at most Class 2 ($d_{6}^{\leqslant} =2$),
    \item If $g_{2}(a)\leqslant11.3845$, then $a$ is assigned to at most Class 2 ($d_{7}^{\leqslant} =2$),
    \item If $g_{2}(a)\leqslant12.1676$, and $g_{4}(a)\leqslant5327.7883$, then $a$ is assigned to at most Class 2 ($d_{8}^{\leqslant} =2$),
    \item If $g_{1}(a)\leqslant65.694$, and $g_{3}(a)\leqslant9.55$, then $a$ is assigned to at most Class 2 ($d_{9}^{\leqslant} =2$),
    \item If $g_{1}(a)\leqslant74.615$, and $g_{3}(a)\leqslant7.26$, then $a$ is assigned to at most Class 2 ($d_{10}^{\leqslant} =2$),
    \item If $g_{3}(a)\leqslant5.64$, and $g_{4}(a)\leqslant2578.156$, then $a$ is assigned to at most Class 1 ($d_{11}^{\leqslant} =1$),
    \item  If $g_{2}(a)\leqslant11.3835$, and $g_{3}(a)\leqslant4.8783$, then $a$ is assigned to at most Class 1 ($d_{12}^{\leqslant} =1$),
    \item If $g_{1}(a)\leqslant58.916$, and $g_{4}(a)\leqslant5376.3961$, then $a$ is assigned to at most Class 1 ($d_{13}^{\leqslant} =1$),
    \item If $g_{1}(a)\leqslant59.766$, and $g_{4}(a)\leqslant2036.9972$, then $a$ is assigned to at most Class 1 ($d_{14}^{\leqslant} =1$).
\end{enumerate}
Let's see how these rules can explain the classifications of four exemplary countries:
\begin{itemize}
    \item Senegal (ID 169) is assigned to Class 1 because, on the one hand, it does not match any \enquote{at-least} rule, and, on the other hand, it matches the \nth{1}, \nth{2}, \nth{3}, \nth{4}, \nth{5}, \nth{6}, \nth{7}, \nth{8}, \nth{10} and \nth{12} \enquote{at-most} rules. The conclusion comes from the fact that the worst class of \enquote{at-most} satisfied rules is $1$. In particular, {due to rule \nth{12}, one can say that the classification to \enquote{at-most} Class 1 is justified by the fact that Senegal's \textit{Expected Years of schooling} is not greater than 11.3835 years and its \textit{Mean years of schooling} is not greater than 4.8783 years};
    \item Comoros (ID 152) is assigned to Class 2 because, on the one hand, it matches the \nth{1} and \nth{3} \enquote{at-least} rules and, on the other hand, it matches the \nth{1}, \nth{2}, \nth{3}, \nth{4}, \nth{5}, \nth{6}, \nth{9} and \nth{10} \enquote{at-most} rules. The conclusion comes from the fact that the best class of \enquote{at-least} satisfied rules is $2$, and the worst class of \enquote{at-most} satisfied rules is also $2$. In particular, the classification to \enquote{at-least} Class 2 can be justified by the two matching \enquote{at-least} rules. From the \nth{1} rule, one can say that its assignment to at least Class $2$ is explained by the fact that its \textit{Expected years of schooling} is {not smaller than} 11.018 years, and its \textit{GNI per capita} is {not smaller than} 3157.3586\$. Analogously, using the \nth{3} rule, the same classification is justified by the fact that its \textit{Life expectancy at birth} is {not smaller than} 61.929 years, and its \textit{Mean years of schooling} is {not smaller than} 5.8443 years. Regarding the classification to \enquote{at-most} Class 2, instead, it can explained by the \nth{6}, \nth{9} or \nth{10} rules. For example, from the \nth{9} rule one can say that its assignment to at most Class $2$ is explained by the fact that Comoros' \textit{Life expectancy at birth} is {not greater than} 65.694 years and its \textit{Mean years of schooling} is {not greater than} 9.55 years;
    \item Brazil (ID 90) is assigned to Class 3 because, on the one hand, it matches the \nth{1}, \nth{2}, \nth{3}, \nth{4}, \nth{5} and \nth{6} \enquote{at-least} rules and, on the other hand, it matches the \nth{1} \enquote{at-most} rule. The conclusion comes from the fact that the best class of \enquote{at-least} satisfied rules is $3$, and the worst class of the \enquote{at-most} satisfied rules is also $3$. In particular, the classification to \enquote{at-least} Class 3 can be justified by the \nth{5} and \nth{6} rules. For example, from the \nth{5} rule, one can say that its assignment to at least Class 3 is explained by the fact that Brazil's \textit{Expected years of schooling} is not smaller than 14.6195 years, and its \textit{GNI per capita} is not smaller than 10296.6496\$. Regarding the classification to \enquote{at-most} Class 3, instead, it is explained by the \nth{1} rule because Brazil's \textit{GNI per capita} is {not greater than} 14778.3463\$;
    \item The Netherlands (ID 11) is assigned to Class 4 because, on the one hand, it matches all \enquote{at-least} rules, and, on the other hand, it does not match any \enquote{at-most} rule. The conclusion comes from the fact that the best class of the \enquote{at-least} satisfied rules is $4$. In particular, the classification to \enquote{at-least} Class 4 can be justified by one of the \nth{12}, \nth{13}, \nth{14}, \nth{15}, \nth{16}, \nth{17} rules. For example, considering the \nth{14} rule, one can say that it is explained by the fact that the Netherlands' \textit{Life expectancy at birth} is not smaller than 73.246 years, and its \textit{Mean years of schooling} is not smaller than 12.2412 years.
\end{itemize}
Of course, similar argumentation can be used to explain the classifications of all other countries in terms of their evaluations on the four criteria at hand.

\subsection{Scenario (iii): Constructing a composite indicator in terms of decision rules from the Decision Maker’s classification of reference units}
\label{sec:scenario3_intro}
With this example, we illustrate how to construct a CI starting from the DM’s classification of reference units. Let us consider the stock portfolio selection case study presented in \cite{Emamat2022} where $50$ stocks, evaluated according to $8$ criteria (all of them with increasing direction of preference) as shown in Appendix A  in Table \ref{tab:ptf_evaluation_criteria}, have to be assigned to 3 classes. To this aim, two different MCDA methods, namely ELECTRE Tri-B \citep{Yu1992} and FlowSort \citep{NemeryLamboray2008} are used. Let us suppose that the DM provides classifications for a set of reference units. Their evaluations on the criteria at hand, together with the corresponding classifications, are presented in Table \ref{tab:scenario3_refunit}. In particular, the classifications were obtained using the ELECTRE Tri-B pessimistic procedure without applying any veto thresholds, as reported in Table 17 of \cite{Emamat2022}.  The units in Table \ref{tab:scenario3_newunit}, instead, are non-reference units, that is, units without a given classification.
\begin{table}[h!]
\centering
\caption{Reference and non-reference units performances evaluated on $8$ criteria and the corresponding direction of preference}
\label{tab:scenario3}
\begin{subtable}{\textwidth}
\subcaption{Reference units}
\label{tab:scenario3_refunit}
    \begin{tabular}
    {c|>{\centering}m{0.08\textwidth}>{\centering}m{0.08\textwidth}>{\centering}m{0.1\textwidth}>{\centering}m{0.08\textwidth}>{\centering}m{0.08\textwidth}>{\centering}m{0.08\textwidth}>{\centering}m{0.08\textwidth}>{\centering}m{0.08\textwidth}|c}
    \toprule
    Unit & Return ($\uparrow$)& Beta ($\uparrow$)& Net Profit Margin ($\uparrow$)& ROA ($\uparrow$)& ROE ($\uparrow$)& EPS ($\uparrow$)& P/E ($\uparrow$)& P/BV ($\uparrow$) & Class\\
    \midrule 
A1&86.692&1.146&10.327&4.94&22.855&402&7.169&2.183&2\\
A4&24.145&1.489&30.193&6.101&21.223&469.667&11.951&1.742&3\\
A12&97.489&0.593&45.938&14.533&45.104&1171.667&8.333&3.108&1\\
A17&82.361&1.023&28.614&14.808&34.276&664&11.682&2.669&3\\
A18&61.004&0.471&144.01&29.242&50.169&3060.333&9.156&7.513&2\\
A26&129.896&1.148&17.002&11.337&29.042&569.333&11.715&1.78&3\\
A33&139.535&0.739&26.339&31.301&62.479&3465.667&2.71&4.585&1\\
A36&32.199&1.017&49.163&32.073&41.901&1034.333&4.15&2.015&2\\
A45&150.822&0.788&37.259&31.046&36.09&2201.667&12.932&4.325&3\\
\bottomrule 
\end{tabular}
\end{subtable}\\[2mm]
\begin{subtable}{\textwidth}
\subcaption{Non-reference/new units}
\label{tab:scenario3_newunit}
    \begin{tabular}
    {c|>{\centering}m{0.08\textwidth}>{\centering}m{0.08\textwidth}>{\centering}m{0.1\textwidth}>{\centering}m{0.08\textwidth}>{\centering}m{0.08\textwidth}>{\centering}m{0.08\textwidth}>{\centering}m{0.08\textwidth}>{\centering}m{0.08\textwidth}|c}
    \toprule
    Unit & Return ($\uparrow$)& Beta ($\uparrow$)& Net Profit Margin ($\uparrow$)& ROA ($\uparrow$)& ROE ($\uparrow$)& EPS ($\uparrow$)& P/E ($\uparrow$)& P/BV ($\uparrow$) & Class\\
    \midrule 
A7&35.747&1.005&52.491&34.533&44.535&1565.333&7.145&3.172&?\\
A19&49.54&1.967&29.94&18.873&36.125&1005&10.14&2.547&?\\
A25&62.824&0.749&39.808&7.556&27.533&653.333&8.608&1.929&?\\
A34&50.781&1.223&27.082&16.003&34.194&669.667&6.455&1.936&?\\
\bottomrule 
\end{tabular}
\end{subtable}
\end{table}
Applying DRSA, $21$ \enquote{at-least} and $17$ \enquote{at-most} rules are generated. These rules can explain the classification assigned by the DM to reference units from Table \ref{tab:scenario3_refunit}. In particular, the minimal subsets of these rules covering all reference units are the following:\\[1mm]
\textbf{\enquote{at-least} rules}
\begin{enumerate}
\item If $g_{3}(a)\geqslant28.614$, and $g_{4}(a)\geqslant14.808$, then $a$ is assigned to at least Class 2; supported by classification examples: $\left\{\mbox{A17},\mbox{A18},\mbox{A36},\mbox{A45}\right\}$,
\item If $g_{2}(a)\geqslant0.788$, then $a$ is assigned to at least Class 2; supported by classification examples: \linebreak $\left\{\mbox{A1},\mbox{A4},\mbox{A17},\mbox{A26},\mbox{A36},\mbox{A45}\right\}$,
\item If $g_{7}(a)\geqslant11.682$, then $a$ is assigned to at least Class 3; supported by classification examples: \linebreak $\left\{\mbox{A4},\mbox{A17},\mbox{A26},\mbox{A45}\right\}$.
\end{enumerate}
\textbf{\enquote{at-most} rules}
\begin{enumerate}
\item If $g_{7}(a)\leqslant9.156$, then $a$ is assigned to at most Class 2; supported by classification examples: \linebreak $\left\{\mbox{A1},\mbox{A12},\mbox{A18},\mbox{A33},\mbox{A36}\right\}$,
\item If $g_{2}(a)\leqslant0.739$, and $g_{8}(a)\leqslant4.585$, then $a$ is assigned to at most Class 1; supported by classification examples: $\left\{\mbox{A12},\mbox{A33}\right\}$.
\end{enumerate}
The above rules can furthermore provide and explain the classification of the non-reference units from Table \ref{tab:scenario3_newunit}. In fact:
\begin{itemize}
    \item A7 is assigned to Class 2 because it matches the first 2 \enquote{at-least} rules (giving it the minimum classification of 2) and it matches the first \enquote{at-most} rule (giving it the maximum classification of 2);
    \item A14 is assigned to Class 3 because it matches the second and the third \enquote{at-least} rules (giving it the minimum classification of 3) and it doesn't match any \enquote{at-most} rule (giving it the maximum classification of 3);
    \item A25 is assigned to at least Class 1 and to at most Class 2. In fact, it doesn't match any \enquote{at-least} rule (giving it the minimum classification of 1) and it matches the first \enquote{at-most} rule (giving it the maximum classification of 2). In other words, it means the unit belongs to Class 1 or 2;
    \item A35 is assigned to at least Class 3 and at most Class 1 because it matches the third \enquote{at-least} rule (giving it the minimum classification of 3) and it matches the second \enquote{at-most} rule (giving it the maximum classification of 1).
\end{itemize}
From the above example, one can see that the first 3 units are classified in a non-contradictory way, since their minimum class assignment (at least) is lower or equal than their maximum class assignment (at most), while the last unit is classified in a \enquote{contradictory way} because its minimum class assignment (at least) is greater than its maximum class assignment (at most). Our proposed procedure explained in Section \ref{sec:algorithm} is able to manage such contradictions.
\subsection{Scenario (iv): Explaining the result of an MCDA scoring method}
With this example, we aim to show that rule-based explanations can also be applied to a composite indicator constructed through an MCDA aggregation of criteria. The rules inferred from the scores assigned to selected reference units by the composite indicator can then be used to evaluate and score other, non-reference units. To this end, we consider a hypothetical case study presented in \cite{FigueiraGrecoRoy2022} in which the ELECTRE-Score method was used to assess several potential locations for a new hotel based on five evaluation criteria: Investment costs ($g_1$), Annual costs ($g_2$), Recruitment ($g_3$), Image ($g_4$), and Access ($g_5$). The first two criteria have a decreasing direction of preference $\left(\downarrow\right)$, while the others have an increasing direction of preference $\left(\uparrow\right)$.

The application of ELECTRE-Score begins with assigning scores to a set of reference units, whose performances on the five considered criteria are shown in Table \ref{tab:score_reference_alternatives}. These units are called reference units because they will serve to explain the scoring procedure using decision rules. The procedure for assigning scores (presented in the last column of the same table) is, in this case, the Deck of Cards method \citep{FigueiraRoy2002}. However, it should be noted that the scores could also be obtained using other MCDA methods, such as Direct Rating \citep{BeltonStewart2002}, Deck-of-cards-based Ordinal Regression \citep{BarbatiGrecoLami2024,CorrenteGrecoZappalà2025}, AHP \citep{Saaty1977}, BWM \citep{Rezaei2015}, or MACBETH \citep{BanaCostaVansnick1994}, to mention some of the best-known approaches in the literature. The units without a given score, referred to as non-reference units, are shown in Table \ref{tab:score_noreference_alternatives}. They will receive a score using a procedure based on the decision rules derived from the scoring of the reference units. Note that the units in Table \ref{tab:score_reference_alternatives} are the reference alternatives from \cite{FigueiraGrecoRoy2022}, to which the Deck of Cards method was applied to obtain scores. In Table \ref{tab:score_noreference_alternatives}, units 
$x_1$ and $x_4$ correspond to alternatives 
$a_1$ and $a_2$ in Table 1 of \cite{FigueiraGrecoRoy2022}, and their final scores were computed using the ELECTRE-Score method. Units 
$x_2$ and $x_3$ were constructed by the authors specifically for didactic purposes.

\begin{table}[!h]
\caption{Reference and non-reference units performances evaluated on $5$ criteria and the corresponding direction of preference}
\label{tab:score_alternatives}
\centering
\begin{subtable}[t]{0.45\textwidth}
\subcaption{Reference units}
\label{tab:score_reference_alternatives}
\vspace{0pt}
\centering
\setlength{\tabcolsep}{5pt}
    \begin{tabular}{lcccccc}
    \toprule
    Unit&$g_{1}\left(\downarrow\right)$&$g_{2}\left(\downarrow\right)$&$g_{3}\left(\uparrow\right)$&$g_{4}\left(\uparrow\right)$&$g_{5}\left(\uparrow\right)$&Score\\
    \midrule
    $a_{1}$&18000&4000&1&1&1&0\\
    $a_{2}$&17000&3500&2&2&1&16.67\\
    $a_{3}$&16500&3700&1&2&1&16.67\\
    $a_{4}$&15350&3200&3&1&2&41.67\\
    $a_{5}$&14250&2850&3&4&3&50\\
    $a_{6}$&13750&3150&4&3&3&50\\
    $a_{7}$&12650&2650&4&4&5&66.67\\
    $a_{8}$&11500&2100&5&6&5&75\\
    $a_{9}$&11000&2500&6&5&7&75\\
    $a_{10}$&10000&2000&7&7&7&100\\
    \bottomrule
    \end{tabular}
\end{subtable}%
\hfill
\begin{subtable}[t]{0.45\textwidth}
\subcaption{Non-reference/new units}
\label{tab:score_noreference_alternatives}
\vspace{0pt}
\centering
\setlength{\tabcolsep}{5pt}
    \begin{tabular}{lcccccc}
    \toprule    Unit&$g_{1}\left(\downarrow\right)$&$g_{2}\left(\downarrow\right)$&$g_{3}\left(\uparrow\right)$&$g_{4}\left(\uparrow\right)$&$g_{5}\left(\uparrow\right)$&Score\\
    \midrule
    $x_{1}$&13000&3000&4&4&4&?\\
    $x_{2}$&10500&2800&5&5&5&?\\
    $x_{3}$&10700&2650&7&8&8&?\\
    $x_{4}$&15000&2500&6&2&7&?\\
    \bottomrule
    \end{tabular}
\end{subtable}
\end{table}  
Applying DRSA, $26$ \enquote{at-least} and $16$ \enquote{at-most} rules are built. Each of them is able to explain the score assigned to units from Table \ref{tab:score_reference_alternatives} in terms of their performances on the five criteria at hand and, more importantly, independently of the way they are aggregated to obtain the score of no-reference units from Table \ref{tab:score_noreference_alternatives}. The minimal subsets of these rules covering all the units in Table \ref{tab:score_reference_alternatives} are the following:

\newpage
\noindent \textbf{\enquote{at-least} rules}
\begin{enumerate}
\item If $g_{1}(a)\leqslant17000$, then the score of $a$ is at least $16.67$; supported by examples: \newline $\left\{a_{2},a_{3},a_{4},a_{5},a_{6},a_{7},a_{8},a_{9},a_{10}\right\}$,
\item If $g_{1}(a)\leqslant15350$, then the score of $a$ is at least $41.67$; supported by examples: $\left\{a_{4},a_{5},a_{6},a_{7},a_{8},a_{9},a_{10}\right\}$,
\item If $g_{1}(a)\leqslant14250$, then the score of $a$ is at least $50$; supported by examples: $\left\{a_{5},a_{6},a_{7},a_{8},a_{9},a_{10}\right\}$,
\item If $g_{3}(a)\geqslant4$, and $g_{4}(a)\geqslant4$, then the score of $a$ is at least $66.67$; supported by examples: $\left\{a_{7},a_{8},a_{9},a_{10}\right\}$,
\item If $g_{3}(a)\geqslant5$, then the score of $a$ is at least $75$; supported by examples: $\left\{a_{8},a_{9},a_{10}\right\}$,
\item If $g_{3}(a)\geqslant7$, then the score of $a$ is at least $100$; supported by examples: $\left\{a_{10}\right\}$.
\end{enumerate}
\textbf{\enquote{at-most} rules}
\begin{enumerate}
\item If $g_{3}(a)\leqslant6$, then the score of $a$ is at most $75$; supported by examples: \newline $\left\{a_{1},a_{2},a_{3},a_{4},a_{5},a_{6},a_{7},a_{8},a_{9}\right\}$,
\item If $g_{3}(a)\leqslant4$, then the score of $a$ is at most $66.67$; supported by examples: $\left\{a_{1},a_{2},a_{3},a_{4},a_{5},a_{6},a_{7}\right\}$,
\item If $g_{5}(a)\leqslant3$, then the score of $a$ is at most $50$; supported by examples: $\left\{a_{1},a_{2},a_{3},a_{4},a_{5},a_{6}\right\}$,
\item If $g_{4}(a)\leqslant2$, then the score of $a$ is at most $41.67$; supported by examples: $\left\{a_{1},a_{2},a_{3},a_{4}\right\}$,
\item If $g_{3}(a)\leqslant2$, then the score of $a$ is at most $16.67$; supported by examples: $\left\{a_{1},a_{2},a_{3}\right\}$,
\item If $g_{1}(a)\geqslant18000$, then the score of $a$ is at most $0$; supported by examples: $\left\{a_{1}\right\}$.
\end{enumerate}

Let us consider now four non-reference units shown in Table \ref{tab:score_noreference_alternatives}. The above \enquote{\enquote{at-least}} and \enquote{\enquote{at-most}} rules can then be used to assign a score to these units. In result of this procedure:
\begin{itemize}
    \item $x_1$ gets a score of $66.67$ because it matches the first $4$ \enquote{at-least} rules (giving it the minimum score of $66.67$), and the first $2$ \enquote{at-most} rules (giving it the maximum score of $66.67$);
    \item $x_2$ gets a score of $75$ because it matches the first $5$ \enquote{at-least} rules (giving it the minimum score of $75$), and the first \enquote{at-most} rule (giving it the maximum score of $75$);
    \item $x_3$ gets a score of $100$ because it matches all the \enquote{at-least} rules (giving it the minimum score of $100$), and none of the \enquote{at-most} rules (giving it the maximum score of $100$);
    \item $x_4$ gets a score of at least $75$ and at most $41.67$. In fact, it matches the first, the second and the fifth \enquote{at-least} rules (giving it the minimum score of $75$), and the first and the fourth \enquote{at-most} rules (giving it the maximum score of $41.67$).
\end{itemize}
Assigning a score to non-reference units can generate different scenarios. As shown in the above example, on the one hand, the first 3 units are scored in a non-contradictory way, since their minimum (at least) score is lower or equal than their maximal (at most) score (and in these specific cases the two scores are exactly equal). On the other hand, the last unit would be scored in a \enquote{contradictory way} since its minimum score (at least) is greater than its maximum score (at most). To manage such contradictions, we propose a new procedure, explained in Section \ref{sec:algorithm}, that applies DRSA to assign appropriate scores or classifications to the units without contradictions.
\section{The background of the methodology}\label{sec:Background}
In this section, keeping the notation introduced at the beginning of Section \ref{sec:examples_intro}, we shall explain the basic concepts necessary to understand our proposal. 
\subsection{Basic concepts}

In the following, without loss of generality, we will assume that all criteria have an increasing direction of preference. We aim to assign each unit from $A$ to one or more scores or Classes $C_1,\ldots, C_p$ preferentially ordered by the DM so that $C_1$ is the worst score or class and $C_p$ is the best one. For each unit $a\in A$, $s^{-}(a)$ and $s^{+}(a)$ will represent the lowest and the highest score or class which can be assigned to $a$, respectively, so that $s^{-}(a)\leqslant s^{+}(a)$. To this aim, we want to use the Dominance-based Rough Set Approach (DRSA) \citep{GrecoMatarazzoSlowinski2001} that yields \textit{\enquote{at-least}} $\left(r_i^{\geqslant}\right)$ and \textit{\enquote{at-most}} $\left(r_i^{\leqslant}\right)$ decision rules explaining scores or classifications provided by the DM on a set of reference units $A^{R}\subseteq A$. To simplify, let's use classifications only, understanding they may also be score assignments. The rules have the following syntax:
\begin{description}
    \item[$r_i^{\geqslant}$:] If the evaluation of $a$ on $g_{j_1}$ is not worse than $\overline{q}^{i}_{j_1}$, and $\ldots$, and the evaluation of $a$ on $g_{j_s}$ is not worse than $\overline{q}^{i}_{j_s}$, then $a$ is classified to at least Class $t$, $t = 2,\ldots,p$,
    \item[$r_i^{\leqslant}$:] If the evaluation of $a$ on $g_{j_1}$ is not better than $\underline{q}^{i}_{j_1}$, and $\ldots$, and the evaluation of $a$ on $g_{j_s}$ is not better than $\underline{q}^{i}_{j_s}$, then $a$ is classified to at most Class $t$, $t = 1,\ldots,p-1$,
\end{description}
where $\overline{q}^{i}_{j}$ and $\underline{q}^{i}_{j}$ are evaluations taken by reference units on $g_{j}\in G$.\\
In the following:
\begin{itemize}
\item $Cl_{t}$ will denote the set of reference units assigned to Class $t$, $t=1,\ldots,p$;
\item $Cl_t^{\geqslant}=Cl_{t}\cup\cdots\cup Cl_{p}$ is the set of reference units assigned to at least Class $t$, with $t=2,\ldots,p$;
\item $Cl_t^{\leqslant}=Cl_{1}\cup\cdots\cup Cl_{t}$ is the set of reference units assigned to at most Class $t$, with $t=1,\ldots,p-1$;
\item ${\cal R}^{\geqslant}$ and ${\cal R}^{\leqslant}$ will denote the set of \enquote{at-least} and \enquote{at-most} rules, respectively;
\item $r^{\geqslant}_i$ and $r^{\leqslant}_i$ denote generic rules from the previous sets.
\end{itemize}
Each rule can be represented in a  compact way as $E\rightarrow H$, where $E$ is the condition part, and $H$ is the decision part. For example, considering $r_i^{\geqslant}$ above, $E$ is \enquote{\textit{the evaluation of $a$ on $g_{j_1}$ is not worse than $\overline{q}^{i}_{j_1}$, and $\ldots$, and the evaluation of $a$ on $g_{j_s}$ is not worse than $\overline{q}^{i}_{j_s}$}}, while $H$ is \enquote{\textit{$a$ is classified to at least Class $t$, $t=2,\ldots,p$}}. The following sets can therefore be defined: 
\begin{itemize}
\item ${\cal R}^{\geqslant t}\subseteq {\cal R}^{\geqslant}$: is the set of \enquote{at-least} rules recommending the decision \enquote{\textit{is assigned to at least Class $t$}}, with $t=2,\ldots,p$;
\item ${\cal R}^{\leqslant t}\subseteq {\cal R}^{\leqslant}$: is the set of \enquote{at-most} rules recommending the decision \enquote{\textit{is assigned to at most Class $t$}}, with $t=1,\ldots,p-1$.
\end{itemize}
In the following, we define some additional notation:
\begin{itemize}
    \item For each $a\in A$, $r_i^{\geqslant}\in {\cal R}^{\geqslant t}$ and $r_i^{\leqslant}\in {\cal R}^{\leqslant t}$, the following indicator functions $\gamma_i^{\geqslant}(a)$ and $\gamma_i^{\leqslant}(a)$ are defined:
    \begin{equation*}
    \gamma_i^{\geqslant}(a) = \left\{
    \begin{array}{ll}
         1&\mbox{if $a$ matches the}\\
         & \mbox{condition part of $r_i^{\geqslant}$} \\[2mm]
         0&\mbox{otherwise},
    \end{array}\right.,\qquad\qquad\gamma_i^{\leqslant}(a) = \left\{
    \begin{array}{ll}
         1&\mbox{if $a$ matches the}\\
         & \mbox{condition part of $r_i^{\leqslant}$}  \\[2mm]
         0&\mbox{otherwise}.
    \end{array}
    \right.
\end{equation*}
On the basis of the previous indicator functions, we can define the following sets: 
\begin{itemize}
\item for each $r_i^{\geqslant}\in{\cal R}^{\geqslant t}$, $E^{\geqslant}_{i}=\left\{a\in A^R: \gamma_{i}^{\geqslant}(a)=1\right\}$. $E^{\geqslant}_{i}$ is the set of reference units matching the condition part of $r_i^{\geqslant}$;
\item for each $r_i^{\leqslant}\in{\cal R}^{\leqslant t}$, $E^{\leqslant}_{i}=\left\{a\in A^R: \gamma_{i}^{\leqslant}(a)=1\right\}$. $E^{\leqslant}_{i}$ is the set of reference units matching the condition part of $r_i^{\leqslant}$;
\end{itemize}
    \item For each $r_i^{\geqslant}\in {\cal R}^{\geqslant t}$ and $r_i^{\leqslant}\in {\cal R}^{\leqslant t}$ their \textit{relative support} is defined as:
    \begin{equation}
        \label{eq:support_all}
        supp^{\geqslant}_{i} = \frac{\displaystyle\sum_{a\in A^R}^{} \gamma_{i}^{\geqslant}(a)}{\left|Cl_{t}^{\geqslant}\right|}=\frac{\left|E_{i}^{\geqslant}\cap Cl_{t}^{\geqslant}\right|}{\left|Cl_{t}^{\geqslant}\right|}, \qquad\qquad supp^{\leqslant}_{i} = \frac{\displaystyle\sum_{a\in A^R}^{} \gamma_{i}^{\leqslant}(a)}{\left|Cl_{t}^{\leqslant}\right|}=\frac{\left|E_{i}^{\leqslant}\cap Cl_{t}^{\leqslant}\right|}{\left|Cl_{t}^{\leqslant}\right|}.
    \end{equation}
    On the one hand, $supp^{\geqslant}_{i}$ is the number of units in $A^R$ matching $r_i^{\geqslant}$ (with decision \enquote{then $a$ is classified to {at} least Class $t$}) divided by the number of reference units assigned to at least Class $t$, while, on the other hand, $supp^{\leqslant}_{i}$ is the number of units in $A^R$ matching $r_i^{\leqslant}$ (with decision \enquote{then $a$ is classified to {at} most Class $t$}) divided by the number of reference units assigned to at most Class $t$;
    \item Considering a set of classifications provided by the DM on units from $A^{R}$ and the sets of rules $\mathcal{R}^{\geqslant}$ and $\mathcal{R}^{\leqslant}$ explaining such classifications, a contingency table can be built for each rule $E\rightarrow H$, as shown in Table \ref{tab:ContingencyTable}.
    \begin{table}[!h]
    \caption{Contingency table of $E$ and $H$ for rule $E\rightarrow H$}
    \label{tab:ContingencyTable}
    \centering
    \begin{tabular}{cccc}
    \toprule
    & $H$ & $\neg H$ & $\sum$  \\
    \midrule
    $E$ & $b$ & $d$ & $b+d$ \\
    $\neg E$ & $c$ & $e$ & $c+e$ \\
    $\sum$ & $b+c$ & $d+e$ & $\left|A^{R}\right|$  \\
    \bottomrule
    \end{tabular}
    \end{table} 
    There, $b$ is the number of reference units matching the condition and the decision part of the rule, $c$ is the number of reference units matching only the decision part of the rule, $d$ is the number of reference units matching only the condition part of the rule and, finally, $e$ is the number of reference units matching neither the condition nor the decision part of the rule. On the basis of this table, the following \textit{Bayesian confirmation measures} can be defined \citep{greco2016measures}:
\begin{equation}
\label{eq:bayesian}
S(H,E)=\frac{b}{b+d}-\frac{c}{c+e} \qquad \mbox{and}\qquad N(H,E)=\frac{b}{b+c}-\frac{d}{d+e}.
\end{equation}
\end{itemize}

Let us also denote by $d_{i}^{\geqslant}$ and $d_{i}^{\leqslant}$ the minimum and the maximum class to which units satisfying the condition part of $r_{i}^{\geqslant}$ and $r_i^{\leqslant}$, respectively, are assigned. This means that, on the one hand, if $r_{i}^{\geqslant}\in{\cal R}^{\geqslant t}$, then, $d_{i}^{\geqslant}=t$, while, on the other hand, if $r_{i}^{\leqslant}\in{\cal R}^{\leqslant t}$, then, $d_{i}^{\leqslant}=t$. 

\subsection{Decision rule induction}\label{Rule_induction}
In this section, we present the procedure used to induce the sets of \enquote{at-least} decision rules $\mathcal{R}^{\geqslant}$ and \enquote{at-most} decision rules $\mathcal{R}^{\leqslant}$. We note that the proposed algorithm allows for the induction of decision rules without requiring a predefined number of classes. Consequently, it can also be applied to composite indicators that yield comprehensive evaluations on a continuous scale, as is common in practice. This is achieved by treating each distinct score assigned by the composite indicator as an ordered class. We begin with the \enquote{at-least} decision rules by introducing some basic concepts. 

Given a set of units $A^R$ evaluated on a set of criteria $G=\{g_1,\ldots,g_m\}$, and an \enquote{at-least} rule $r_i^{\geqslant}$, its \textit{confidence} $conf(r_i^{\geqslant})$ is equal to the ratio between the number of units in $A^R$ that satisfy both its condition and decision part, and the number of units in $A^R$ that satisfy its condition part, regardless of whether the decision part is satisfied. For example, considering the following at-least rule
\begin{center}
$r_i^{\geqslant}$: If the evaluation of $a$ on $g_{j_1}$ is not worse than $\overline{q}^{i}_{j_1}$, and $,\ldots,$ and the evaluation of $a$ on $g_{j_s}$ is not worse than $\overline{q}^{i}_{j_s}$, then $a\in Cl_{t}^{\geqslant}$, $\ t \in \{2,\ldots,p \}$,
\end{center}
one has 
\begin{equation}
\textit{conf}\left(r^\geqslant_i\right)=\displaystyle\frac{\left|\left\{a\in A^R:g_{j}(a)\geqslant\overline{q}^i_j \text { for all } g_j \in P^\geqslant_i: a \in Cl^\geqslant_t\right\}\right|}{\left|\left\{a\in A^R:g_{j}(a)\geqslant\overline{q}^i_j \text { for all } g_j \in P^\geqslant_i\right\}\right|}
\end{equation}
with $P_i^{\geqslant}=\left\{g_{j_1},\ldots,g_{j_s}\right\}\subseteq G$.

A decision rule $r_i^{\geqslant}$ is called \textbf{exact} if $\textit{conf}\left(r_i^{\geqslant}\right) = 1$, meaning that all units satisfy both the condition part and the decision part. If $\textit{conf}\left(r_i^{\geqslant}\right) < 1$, the rule is called \textbf{approximate} (or \textbf{indeterminate}), indicating that some units satisfy the condition part but not the decision part. In this paper, we consider exact rules. Here, however, for the sake of completeness, we present an algorithm to induce both exact and approximate rules, with the latter required to have a confidence not smaller than $c$.

Consider the following rules:
\begin{center}
$r_i^{\geqslant}$: If 
$g_j(a) \geqslant \overline{q}^{i}_j$ for all $g_j \in P^\geqslant_{i}$, then $a \in Cl^\geqslant_t$,
\end{center}
\begin{center}
$r_{i^\prime}^{\geqslant}$: If 
$g_j(a) \geqslant \overline{q}^{i^\prime}_j$ for all $g_j \in P^\geqslant_{i^\prime}$, then $a \in Cl^\geqslant_{t^\prime}$.
\end{center}
We say that $r_i^\geqslant$ is \emph{not weaker than} $r_{i^\prime}^\geqslant$, denoted $r^\geqslant_i \succsim r_{i^\prime}^\geqslant$, if the following conditions hold:
\begin{itemize}
    \item $P_i^\geqslant \subseteq P_{i^\prime}^\geqslant$ (the set of conditions of $r_i^\geqslant$ is contained in that of $r_{i^\prime}^\geqslant$),
    \item $\overline{q}^i_j \leqslant \overline{q}^{i^\prime}_j$ for all $g_j \in P^\geqslant_i \cap P^\geqslant_{i^\prime}$ (thresholds of $r_i^\geqslant$ are not higher than those of $r_{i^\prime}^\geqslant$),
    \item $\textit{conf}\left(r_i^{\geqslant}\right) \geqslant \textit{conf}\left(r_{i^\prime}^{\geqslant}\right)$ (confidence of $r_i^\geqslant$ is at least that of $r_{i^\prime}^\geqslant$),
    \item $t \geqslant t^\prime$ (the class assigned by $r_i^\geqslant$ is not worse than that of $r_{i^\prime}^\geqslant$).
\end{itemize}
We define two related relations:
\begin{itemize}
    \item $\sim$, the symmetric part of $\succsim$, where
    $$r^\geqslant_i \sim r_{i^\prime}^\geqslant \quad \text{if and only if} \ \ r^\geqslant_i \succsim r_{i^\prime}^\geqslant \text{ and } r^\geqslant_{i^\prime} \succsim r_i^\geqslant;$$
    \item $\succ$, the asymmetric part of $\succsim$, where
    $$r^\geqslant_i \succ r_{i^\prime}^\geqslant \quad \text{if and only if} \ \ r^\geqslant_i \succsim r_{i^\prime}^\geqslant \;\;\text{and}\;not \left(r^\geqslant_{i^\prime} \succsim r_i^\geqslant\right).$$
\end{itemize}
From these definitions, the following properties hold:
\begin{itemize}
    \item $r^\geqslant_i \sim r_{i^\prime}^\geqslant$ if and only if $r^\geqslant_i = r_{i^\prime}^\geqslant$.
    \item If $r^\geqslant_i \succ r_{i^\prime}^\geqslant$ and a unit $a \in A$ satisfies the condition part of $r^\geqslant_{i^\prime}$, then it also satisfies the condition part of $r_{i}^\geqslant$. In this case:
    \begin{itemize}
    \item According to $r_i^\geqslant$, $a$ is assigned to Class $Cl_t$ or better, with confidence $\textit{conf}\left(r^\geqslant_i\right)$. 
    \item According to $r_{i^\prime}^\geqslant$, $a$ would instead be assigned to Class $Cl_{t^\prime}$ or better, where $Cl_{t^\prime}$ is not better than $Cl_t$ and $\textit{conf}\left(r_{i^\prime}^\geqslant\right) \leqslant \textit{conf}\left(r_i^\geqslant\right)$.\\ 
    This implies that if $r^\geqslant_i \succ r_{i^\prime}^\geqslant$, then $r_{i^\prime}^\geqslant$ can be safely removed without changing the final classification.
    \end{itemize}
\end{itemize}

We provide Algorithm \ref{algo:rulesgeneration} for decision rule induction that is based on the following principle: to each $a \in A^R$, with $a \in Cl_t$ and $P \subseteq G$, is associated the decision rule
\begin{center}
$r^{\geqslant}$: If 
$g_j(x) \geqslant \overline{q}_j$ for all $g_j \in P$, then $x \in Cl^\geqslant_t$,
\end{center}
with $\overline{q}_j=g_j(a)$ for all $g_j \in P$. The proposed algorithm finds all possible \enquote{at-least} rules.

\begin{algorithm}
\setlength{\baselineskip}{18pt}
\caption{Pseudo-code for \enquote{at-least} rules generation \label{algo:rulesgeneration}}
\begin{normalsize}

\begin{algorithmic}[1]
\Require Performances
$g_j\left(a\right),\;j=1,\ldots,m$, class assignments $Cl_1,\ldots,Cl_p$ for reference units $a\in A^{R}\subseteq A$, minimal rule confidence $c$.
\Ensure Set of decision rules $\mathcal{R}^{\geqslant} = \left\{r_i^{\geqslant}\right\}$, with 
\begin{center}
$r_i^\geqslant:$ If $g_{j_1} \left(a\right) \geqslant \overline{q}^{i}_{j_1},$ and $\ldots,$ and $g_{j_s}\left(a\right) \geqslant \overline{q}^{i}_{j_s}$, then $a \in Cl^\geqslant_{t}$, \;$t=2,\ldots,p$
\end{center}
and confidence $\textit{conf}\left(r^{\geqslant}_i\right)$ $\left(P^\geqslant_i=\left\{g_{j_1}, \ldots, g_{j_s}\right\} \subseteq G \right)$
\Steps
\State ${\cal R}^{\geqslant}=\emptyset$, ${\cal P}=sort\left(2^G\setminus\left\{\emptyset\right\}\right)$
\For{$P\in \mathcal{P}$}
\State $\mathcal{Q} = \emptyset$
\For{$a\in A^R \cap Cl_{2}^{\geqslant}$}
\For{$t = 2,\ldots,p$}
\State \makecell[l]{Compute $\displaystyle L_{\geqslant t}^P\left(a\right)=\frac{\left|\left\{a'\in D_P^{+}(a):a'\in Cl^{\geqslant}_t\right\}\right|}{\left|D_P^{+}(a)\right|}$\\
where $D_P^{+}(a)=\left\{a'\in A^R: g_j\left(a'\right)\geqslant g_j(a), \forall g_j\in P\right\}$}
\If{$L_{\geqslant t}^P\left(a\right)\geqslant c$}
\If{$\mathcal{R}^\geqslant=\emptyset$ or for all $r_i^{\geqslant}\in \mathcal{R}^{\geqslant}$ at least one among the following conditions holds:
\Statex \hspace{31mm} -  $C_1\left(r_i^\geqslant\right)$: $P_i^\geqslant\not\subseteq P$ 
\Statex \hspace{31mm} - $C_2\left(r_i^\geqslant\right)$: $P_i^\geqslant\subseteq P$ and there exists $g_j\in P_i^\geqslant$ such that $\overline{q}_{j}^{i}>g_j\left(a\right)$
\Statex \hspace{31mm} - $C_3\left(r_i^\geqslant\right)$: $P_i^\geqslant\subseteq P$, $\overline{q}_{j}^{i}\leqslant g_j\left(a\right)$ for all $g_j\in P_i^\geqslant$ and $d_i^\geqslant < t$ 
\Statex \hspace{31mm} - $C_4\left(r_i^\geqslant\right)$:  $P_i^\geqslant\subseteq P$, $\overline{q}_{j}^{i}\leqslant g_j\left(a\right)$ for all $g_j\in P_i^\geqslant$, $d_i^\geqslant \geqslant t$ and $L_{\geqslant t}^P\left(a\right)>\mbox{\textit{conf}}\left(r_i^\geqslant\right)$
\Statex \hspace{28mm}}
\State \makecell[l]{Generate rule $\widetilde{r}_{\geqslant t}^{P}\left(a\right)$:
\enquote{If $g_{j_1}\left(x\right)\geqslant g_{j_1}\left(a\right),\ldots,g_{j_s}\left(x\right)\geqslant g_{j_s}\left(a\right), \mbox{then } x\in Cl_{t}^{\geqslant}$}\\ with confidence $\mbox{\textit{conf}}\left(\widetilde{r}_{\geqslant t}^P\left(a\right)\right)=L_{\geqslant t}^P\left(a\right)$}
\State $\mathcal{Q} = \mathcal{Q}\cup \left\{\widetilde{r}_{\geqslant t}^P\left(a\right)\right\}$
\EndIf
\EndIf
\EndFor
\EndFor
\State Remove duplicate $\widetilde{r}_{\geqslant t}^P\left(a\right)$ from $\mathcal{Q}$, keeping only unique rows
\For{$\widetilde{r}_{\geqslant t}^P\left(a\right) \in \mathcal{Q}$}
\If{for all $\widetilde{r}_{\geqslant t^\prime}^P\left(a^\prime\right) \in \mathcal{Q}\setminus \widetilde{r}_{\geqslant t}^P\left(a\right)$ at least one among the following conditions holds:
\vspace{2mm}
\Statex \hspace{20mm} - $C_{1}^\prime\left(\widetilde{r}_{\geqslant t^\prime}^P\left(a^\prime\right)\right)$: there exists $g_j\in P$ such that $g_j\left(a^\prime\right)> g_j\left(a\right)$
\vspace{2mm}
\Statex \hspace{20mm} - 
$C_{2}^\prime\left(\widetilde{r}_{\geqslant t^\prime}^P\left(a^\prime\right)\right)$: $g_j\left(a^\prime\right)\leqslant g_j\left(a\right)$ for all $g_j\in P$ and $t^\prime<t$
\vspace{2mm}
\Statex \hspace{20mm} - $C_{3}^\prime\left(\widetilde{r}_{\geqslant t^\prime}^P\left(a^\prime\right)\right)$: $g_j\left(a^\prime\right)\leqslant g_j\left(a\right)$ for all $g_j\in P$, $t^\prime\geqslant t$ and $L_{\geqslant t}^P\left(a\right)>L_{\geqslant t^\prime}^{P}\left(a^\prime\right)$
\Statex \hspace{16mm}}
\State $\mathcal{R}^\geqslant=\mathcal{R}^\geqslant\cup \left\{\widetilde{r}_{\geqslant t}^{P}\left(a\right)\right\}$
\EndIf
\EndFor
\EndFor
\end{algorithmic}
\end{normalsize}
\end{algorithm}

\begin{description}    
\item[\textbf{Input}:] There are given performances $g_j\left(a\right),j=1,\ldots,m,$ of units from the set of reference units $A^R$, and their classifications $a\in Cl_t$ given by the DM. Moreover, the DM is fixing the minimal rule confidence value $c\in\left]0,1\right]$. The lower the value of $c$, the higher the tolerated proportion of counterexamples relative to the number of units in the dataset that satisfy the condition part of the decision rule.
\item[\textbf{1}:] The set ${\cal R}^{\geqslant}$ of \enquote{at-least} rules is initialized as the empty set. 
The procedure aims to generate all rules that can be derived for every subset of $G$. 
However, rule construction must involve the minimum number of criteria. 
To ensure this, the subsets of $G$ (that is, $2^{G} \setminus \{\emptyset\}$) must be ordered so that any subset $P \in 2^{G}\setminus \left\{\emptyset\right\}$ is analyzed only after all its proper subsets have already been considered\footnote{Given two sets $A$ and $B$, $A$ is a proper subset of $B$ iff $A\subset B$.}. 
This ordering can be achieved by enumerating all non-empty subsets of $G$ using their binary representation. 
Specifically, one can consider the integers from $1$ to $2^{|G|}-1$ in the binary form, 
and include in the subset the criteria corresponding to the positions of the 1's in each binary number (assuming bit 1 corresponds to $g_1$, bit 2 to $g_2$, and so on). 
For example, if $G = \{g_1, g_2, g_3\}$, the integers from 1 to 7 ($001$, $010$, $011$, $100$, $101$, $110$, $111$ in binary) correspond to the subsets: 
$\{g_3\}, \{g_2\}, \{g_2,g_3\}, \{g_1\}, \{g_1,g_3\}, \{g_1,g_2\}, \{g_1,g_2,g_3\}$. 
Equivalently, the subsets can also be processed incrementally by size; for $G = \{g_1, g_2, g_3\}$, this gives the order: 
$\{g_1\}, \{g_2\}, \{g_3\}, \{g_1,g_2\}, \{g_1,g_3\}, \{g_2,g_3\}, \{g_1,g_2,g_3\}$. 
Thus, ${\cal P}$ denotes the collection $2^{G} \setminus \{\emptyset\}$, sorted according to this rule.
\item[\textbf{2}:] Select $P\in{\cal P}$ following the sorting performed in step 1. This procedure will end when all subsets of criteria in ${\cal P}$ have been analyzed (step 21:).
\item[\textbf{3}:] The local variable $\mathcal{Q}$, denoting the set of rules created in correspondence of the set of criteria $P$, is initialized as the empty set.
\item[\textbf{4}:] Select a reference unit $a$ that the user has classified as belonging to at least $Cl_2$. This procedure ends when all reference units classified to at least Class 2 have been analyzed (step 14:). 
\item[\textbf{5}:] Select Class $Cl_t$, with $t\in\{2,\ldots,p\}$. This procedure ends when all classes have been considered (step 13:). 
\item[\textbf{6}:] Calculate the credibility $L^{P}_{\geqslant t}\left(a\right)$ of $a$ to belong to $Cl_t^{\geqslant}$ considering the subset of criteria $P$ (selected in step 2:). \\
Denote by $D^+_P\left(a\right)$ the set of reference units weakly dominating $a$ with respect to $P$ ($a^\prime$ weakly dominates $a$ with respect to $P$ iff $g_j\left(a^\prime\right)\geqslant g_{j}\left(a\right)$ for all $g_j\in P$). The credibility $L^{P}_{\geqslant t}\left(a\right)$ is defined as the ratio between the number of reference units that weakly dominate $a$ with respect to $P$ and are assigned by the user to at least Class $Cl_t$ $\left(\left|\left\{a' \in D^{+}_{P}(a): a' \in Cl_t^{\geqslant}\right\}\right|\right)$, and the number of reference units that weakly dominate $a$ with respect to $P$ 
$\left(\left|D^{+}_{P}(a)\right|\right)$. Intuitively, $L^{P}_{\geqslant t}\left(a\right)$ measures the proportion of units in $A^R$ that are at least as good as $a$ with respect to the criteria in $P$ and are assigned to at least Class $Cl_t$. 
\item[\textbf{7}:] Verify if the credibility of $a$ to belong to $Cl_t^{\geqslant}$ with respect to $P$ is not lower than the confidence value $c$. In such a case, proceed to step 8:, otherwise, pass to the next $t$ (step 5:).
\item[\textbf{8}:] Check if $\mathcal{R}^{\geqslant}=\emptyset$ or if for each rule $r_i^{\geqslant}\in \mathcal{R}^\geqslant$ (all at-least rules built up to now) one of the following conditions holds:
\begin{itemize}
\item $P_i^\geqslant\not\subseteq P$: this means that the set of criteria on which $r_i^{\geqslant}$ is based $\left(P_i^{\geqslant}\right)$ is not a subset of $P$,
\item $P_i^\geqslant\subseteq P$ \text{and there exists} $g_j\in P_i^\geqslant$ such that $\overline{q}^{i}_{j}> g_j\left(a\right)$: this means that $r_i^{\geqslant}$ is based on a subset of $P$ $\left(P_i^{\geqslant}\subseteq P\right)$ and the evaluation of $a$ on at least one criterion $g_j\in P_{i}^{\geqslant}$ is lower than the threshold on $g_j$ considered in $r_i^{\geqslant}$ $\left(\overline{q}^{i}_{j}> g_j\left(a\right)\right)$,
\item $P_i^\geqslant\subseteq P$, $\overline{q}^{i}_{j}\leqslant g_j\left(a\right)$ for all $g_j\in P_i^\geqslant$ and $d_i^{\geqslant}<t$ : this means that $r_i^{\geqslant}$ is based on a subset of $P$ $\left(P_i^{\geqslant}\subseteq P\right)$, the evaluation of $a$ on each criterion $g_j$ from $P_i^{\geqslant}$ is not lower than the threshold on $g_j$ considered in the rule $\left(\overline{q}^{i}_{j}\leqslant g_j\left(a\right)\;\mbox{for all}\;g_j\in P_i^\geqslant\right)$ and the decision part of $r_i^{\geqslant}$ is lower than $t$ $\left(d_{i}^{\geqslant}<t\right)$,
\item $P_i^\geqslant\subseteq P$, $\overline{q}^{i}_{j}\leqslant g_j\left(a\right)$ for all $g_j\in P_i^\geqslant$, $d_i^{\geqslant}\geqslant t$ and $L_{\geqslant t}^{P}(a)>conf\left(r_i^{\geqslant}\right)$: this means that $r_i^{\geqslant}$ is based on a subset of $P$ $\left(P_i^\geqslant\subseteq P\right)$, the evaluation of $a$ on each $g_j\in P_i^{\geqslant}$ is not lower than the threshold on $g_j$ considered in the rule $\left(\overline{q}^{i}_{j}\leqslant g_j\left(a\right)\;\mbox{for all}\;g_j\in P_i^\geqslant\right)$, the decision part of $r_i^{\geqslant}$ is not lower than $t$ $\left(d_i^{\geqslant}\geqslant t\right)$ and the credibility of $a$ to be assigned to at least Class $Cl_t$ is greater than $r_i^{\geqslant}$'s confidence $\left(L_{\geqslant t}^P\left(a\right)>\mbox{\textit{conf}}(r_i^{\geqslant}\right))$.
\end{itemize} 
If this is the case, go to step 9:, otherwise, increase $t$ (go to step 5:). 
\item[\textbf{9}:] A new rule $\widetilde{r}_{\geqslant t}^{P}\left(a\right)$ is created based on reference unit $a$, with thresholds taken from $a$'s evaluations on criteria in $P=\left\{g_{j_1},\ldots,g_{j_s}\right\}$, and its confidence equals the credibility of $a$ being assigned to at least Class $Cl_t^{\geqslant}$ $\left(conf\left(\widetilde{r}_{\geqslant t}^{P}(a)\right)=L^{P}_{\geqslant t}(a)\right)$. 
\item[\textbf{10}:] Rule $\widetilde{r}_{\geqslant t}^{P}(a)$ is added to set ${\cal Q}$ $\left(\mathcal{Q}=\mathcal{Q} \cup \left\{\widetilde{r}_{\geqslant t}^{P}\left(a\right)\right\}\right)$. At the end of steps 4:-14: ${\cal Q}$ is composed of all rules generated based on reference units classified to at least $Cl_{2}$ with respect to the subset of criteria $P$ and for each Class $Cl_t$, $t=2,\ldots,p$. In particular, these rules are not weaker than previous rules stored in ${\cal R}^{\geqslant}$.  
\item[\textbf{15}:] Since different reference units can generate the same rule (because they have the same evaluations on criteria in $P$), the duplicate rules should be removed.
\item[\textbf{16}-\textbf{20}:] While in steps 8:-11: we check whether a new rule to be created - based on reference unit $a$, the set of criteria $P$ and Class $Cl_t$ - is weaker than any rule already stored in ${\cal R}^{\geqslant}$, in steps 16:-20: we remove from ${\cal Q}$ all rules that are weaker than others in the same set and we store in ${\cal R}^{\geqslant}$ the remaining ones. \\
In particular, $\widetilde{r}_{\geqslant t}^{P}\left(a\right)\in{\cal Q}$ is included in ${\cal R}^{\geqslant}$ if for all $\widetilde{r}_{\geqslant t^{\prime}}^{P}\left(a^{\prime}\right)\in{\cal Q}$ at least one of the following conditions holds:
\begin{itemize}
\item there exists $g_j\in P$ such that $g_j\left(a'\right)> g_j\left(a\right)$: this means that there is at least one criterion $g_j \in P$ on which $a^{\prime}$ has a greater evaluation than $a$, implying that the condition imposed by $a$ on $g_j$ is less restrictive than the one imposed by $a^{\prime}$ on the same criterion,   
\item  $g_j\left(a'\right)\leqslant g_j\left(a\right)$ for all $g_j\in P$ and $t^{\prime}<t$: this means that the evaluations of $a$ on criteria from $P$ are not lower than those of $a^{\prime}$ but the rule built on $a$ indicates a decision class greater than the decision class of the rule based on $a^{\prime}$, 
\item  $g_j\left(a'\right)\leqslant g_j\left(a\right)$ for all $g_j\in P$, $t^{\prime}\geqslant t$ and $L^{P}_{\geqslant t}\left(a\right)>L^{P}_{\geqslant t^{\prime}}\left(a^\prime\right)$: this means that $a$ has evaluations not lower than $a^{\prime}$ on all criteria from $P$, the decision class of the rule built on $a^{\prime}$ is not lower than the decision class of the rule built on $a$ but the credibility of $a$ to be assigned to at least Class $Cl_t$ is greater than the credibility of $a^{\prime}$ to be assigned to at least Class $Cl_{t^{\prime}}$. 
\end{itemize}
\end{description}

The algorithm for the induction of \enquote{at-most} decision rules is presented below as Algorithm \ref{algo:rulesgeneration_atmost}. Its steps are analogous to Algorithm \ref{algo:rulesgeneration}.

\begin{algorithm}
\setlength{\baselineskip}{18pt}
\caption{Pseudo-code for \enquote{at-most} rules generation  \label{algo:rulesgeneration_atmost}}
\begin{normalsize}

\begin{algorithmic}[1]
\Require Performances
$g_j\left(a\right),\;j=1,\ldots,m$, class assignments $Cl_1,\ldots,Cl_p$ for reference units $a\in A^{R}\subseteq A$, minimal rule confidence $c$.
\Ensure Set of decision rules $\mathcal{R}^{\leqslant} = \left\{r_i^{\leqslant}\right\}$, with 
\begin{center}
$r_i^\leqslant:$ If $g_{j_1} \left(a\right) \leqslant \underline{q}^{i}_{j_1},$ and $\ldots,$ and $g_{j_s}\left(a\right) \leqslant \underline{q}^{i}_{j_s}$, then $a \in Cl^\leqslant_{t}\;t = 1,\ldots,p-1$
\end{center}
and confidence $\textit{conf}\left(r^{\leqslant}_i\right)$ $\left(P^\leqslant_i=\left\{g_{j_1}, \ldots, g_{j_s}\right\} \subseteq G \right)$
\Steps
\State ${\cal R}^{\leqslant}=\emptyset$, ${\cal P}=sort\left(2^G\setminus\left\{\emptyset\right\}\right)$
\For{$P\in \mathcal{P}$}
\State $\mathcal{Q} = \emptyset$
\For{$a\in A^R \cap Cl_{p-1}^{\leqslant}$}
\For{$t = 1,\ldots,p-1$}
\State \makecell[l]{Compute $\displaystyle L_{\leqslant t}^{P}\left(a\right)=\frac{\left|\left\{a^\prime\in D_P^{-}(a):a^\prime\in Cl^{\leqslant}_t\right\}\right|}{\left|D_P^{-}(a)\right|}$\\
where $D_P^{-}(a)=\left\{a^\prime\in A^R: g_j\left(a^\prime\right)\leqslant g_j(a), \forall g_j\in P\right\}$}
\If{$L_{\leqslant t}^{P}\left(a\right)\geqslant c$}
\If{$\mathcal{R}^\leqslant=\emptyset$ or for all $r_i^{\leqslant}\in \mathcal{R}^{\leqslant}$ at least one among the following conditions holds:
\Statex \hspace{31mm} -  $C_1\left(r_i^\leqslant\right)$: $P_i^\leqslant\not\subseteq P$ 
\Statex \hspace{31mm} - $C_2\left(r_i^\leqslant\right)$: $P_i^\leqslant\subseteq P$ and there exists $g_j\in P_i^\leqslant$ such that $\underline{q}_{j}^{i}<g_j\left(a\right)$
\Statex \hspace{31mm} - $C_3\left(r_i^\leqslant\right)$: $P_i^\leqslant\subseteq P$, $\underline{q}_{j}^{i}\geqslant g_j\left(a\right)$ for all $g_j\in P_i^\leqslant$ and $d_i^\leqslant > t$ 
\Statex \hspace{31mm} - $C_4\left(r_i^\leqslant\right)$:  $P_i^\leqslant\subseteq P$, $\underline{q}_{j}^{i}\geqslant g_j\left(a\right)$ for all $g_j\in P_i^\leqslant$, $d_i^\leqslant \leqslant t$ and $L_{\leqslant t}^{P}\left(a\right)>\mbox{\textit{conf}}\left(r_i^\leqslant\right)$
\Statex \hspace{28mm}}
\State \makecell[l]{Generate rule $\widetilde{r}_{\leqslant t}^{P}\left(a\right)$:
\enquote{If $g_{j_1}\left(x\right)\leqslant g_{j_1}\left(a\right),\ldots,g_{j_s}\left(x\right)\leqslant g_{j_s}\left(a\right), \mbox{then } x\in Cl_{t}^{\leqslant}$}\\ with confidence $\mbox{\textit{conf}}\left(\widetilde{r}_{\leqslant t}^P\left(a\right)\right)=L_{\leqslant t}^{P}\left(a\right)$}
\State $\mathcal{Q} = \mathcal{Q}\cup \left\{\widetilde{r}_{\leqslant t}^P\left(a\right)\right\}$
\EndIf
\EndIf
\EndFor
\EndFor
\State Remove duplicate $\widetilde{r}_{\leqslant t}^P\left(a\right)$ from $\mathcal{Q}$, keeping only unique rows
\For{$\widetilde{r}_{\leqslant t}^P\left(a\right) \in \mathcal{Q}$}
\If{for all $\widetilde{r}_{\leqslant t^\prime}^P\left(a^\prime\right) \in \mathcal{Q}\setminus\widetilde{r}_{\leqslant t}^P\left(a\right)$ at least one among the following conditions holds:
\vspace{2mm}
\Statex \hspace{20mm} - $C_{1}^\prime\left(\widetilde{r}_{\leqslant t^\prime}^P\left(a^\prime\right)\right)$: there exists $g_j\in P$ such that $g_j\left(a^\prime\right)< g_j\left(a\right)$
\vspace{2mm}
\Statex \hspace{20mm} - $C_{2}^\prime\left(\widetilde{r}_{\leqslant t^\prime}^P\left(a^{\prime}\right)\right)$: $g_j\left(a^\prime\right)\geqslant g_j\left(a\right)$ for all $g_j\in P$ and $t^\prime>t$
\vspace{2mm}
\Statex \hspace{20mm} - $C_{3}^\prime\left(\widetilde{r}_{\leqslant t^\prime}^P\left(a^\prime\right)\right)$: $g_j\left(a^\prime\right)\geqslant g_j\left(a\right)$ for all $g_j\in P$, $t^\prime\leqslant t$ and $L_{\leqslant t}^{P}\left(a\right)>L_{\leqslant t^\prime}^{P}\left(a^\prime\right)$
\Statex \hspace{16mm}}
\State $\mathcal{R}^\leqslant=\mathcal{R}^\leqslant\cup \left\{\widetilde{r}_{\leqslant t}^{P}\left(a\right)\right\}$
\EndIf
\EndFor
\EndFor
\end{algorithmic}
\end{normalsize}
\end{algorithm}

\vspace{12pt}
\textbf{Example.}
We will illustrate Algorithms \ref{algo:rulesgeneration} and \ref{algo:rulesgeneration_atmost} using the reference units reported in Appendix A in Table \ref{tab:hdi_index}.  
In what follows, we present how some of the \enquote{at-least} and \enquote{at-most} rules are generated, referring to the corresponding steps of the pseudo-codes.  
Let $A^R = \{a_1,\ldots, a_{193}\}$ denote the set of reference units, and $G=\{g_1,g_2,g_3,g_4\}$ the set of criteria, all of which are of the gain-type (to be maximized).

Algorithm \ref{algo:rulesgeneration} iteratively explores all non-empty subsets $P\subseteq G$ (step 2). For a given subset $P$, it first computes the credibility $L_{\geqslant t}^{P}(a)$ of each reference unit $a\in A^R$ that has been classified by the user to at least $Cl_2$, to be classified into $Cl^{\geqslant}_{t}$ for $t = 2,3,4$. It then retains, as potential rule anchors, only those reference units whose credibility is not lower than the minimum threshold $c$ (here, $c=1$), and that are not weaker than the rules generated in previous iterations (steps 4-10). Finally, it identifies the non-dominated anchors, i.e., the reference units generating the strongest rules for the same $P$ (steps 16-20).
\paragraph{Rules induced by the subset $\bm{P=\left\{g_1\right\}}$}
Let us show how the first \enquote{at-least} rules based on $P =\left\{g_1\right\}$ are built. These are:
\begin{align*}
r_1^{\geqslant}:\;&\mbox{if $g_1\left(a\right)\geqslant80.9890$, then $a$ is assigned to at least Class $4$, whose base unit is $a_{7}$},\\
r_2^{\geqslant}:\;&\mbox{if $g_1\left(a\right)\geqslant75.2930$, then $a$ is assigned to at least Class $3$, whose base unit is $a_{80}$},\\
r_3^{\geqslant}:\;&\mbox{if $g_1\left(a\right)\geqslant68.25$, then $a$ is assigned to at least Class $2$, whose base unit is $a_{112}$}.
\end{align*}
\begin{description}
	\item[Step 1:] $\mathcal{R}^\geqslant$ is initialized equal to $\emptyset$, and $\mathcal{P}$ is the ordered set of $2^G\setminus\emptyset$, that is
	\begin{align*}
	\mathcal{P} = &\left\{\left\{g_1\right\},\left\{g_2\right\},\left\{g_3\right\},\left\{g_4\right\},\left\{g_1,g_2\right\},\left\{g_1,g_3\right\},\left\{g_1,g_4\right\},\left\{g_2,g_3\right\},\left\{g_2,g_4\right\},\left\{g_3,g_4\right\},\left\{g_1,g_2,g_3\right\},\right.\\
	&\left.\left\{g_1,g_2,g_4\right\},\left\{g_1,g_3,g_4\right\},\left\{g_2,g_3,g_4\right\},\left\{g_1,g_2,g_3,g_4\right\}\right\}.
	\end{align*}
	\item[Steps 2-3:] The subset of criteria $P = \left\{g_1\right\}$, being the first element of $\mathcal{P}$, is selected. Moreover, the temporary set $\mathcal{Q}$ of generated rules for $P$ is initialized as an empty set.
	\item[Steps 4-14:] For each unit $a\in A^R$ being assigned by the user to $Cl^\geqslant_2$ and for each $t = 2, 3,4$, the algorithm computes the dominance set $D^+_P\left(a\right)$ and, consequently, the credibility $L_{\geqslant t}^P\left(a\right)$ for each unit to be classified to at least $Cl_t$ with respect to $P$. For instance, for $a_7$ ($g_1(a_7) = 80.9890$) and $t=4$ we have 
	\begin{align*}
    D^+_P\left(a_7\right) =& \left\{a_{1},a_{2},a_{3},a_{4},a_{5},a_{6},a_{7},a_{8},a_{9},a_{10},a_{11},a_{12},a_{13},a_{14},a_{15},a_{16},a_{18},a_{19},a_{20},a_{22},a_{23},a_{24},a_{25},\right.\\
    &\left.a_{26},a_{27},a_{28},a_{29},a_{30},a_{35},a_{40},a_{42},a_{43}\right\},
    \end{align*}
	all belonging to $Cl_4$. Consequently, the credibility $L_{\geqslant4}^P\left(a_7\right)=1$. For this reason, in addition to the fact that $\mathcal{R}^\geqslant=\emptyset$, the candidate rule $\widetilde{r}_{\geqslant4}^P\left(a_7\right)$ is added to the temporary set $\mathcal{Q}$.
	\item[Step 15:] After exploring all reference units for $P=\{g_1\}$, $\mathcal{Q}$ contains duplicated rules induced by $a_{39}$ and $a_{54}$, with $t = 2,3$ (both of them have the same evaluation - $79.236$ - on $g_1$) as well as rules induced by $a_{115}$ and $a_{147}$, with $t=2$ (both of them have the same evaluation - $70.484$ - on $g_1$). One of the duplicated rules for each of these pairs is therefore deleted.
	\item[Steps 16-18:] These lines iterate over the candidate rules in $\mathcal{Q}$ that were generated for the same subset $P$. For each candidate $\widetilde{r}_{t}^{P}\left(a\right)\in\mathcal{Q}$, the algorithm checks whether there exists another one for the same $P$ producing a stronger (i.e., more specific or dominating) rule. Only candidates not dominated by others are finally retained. In this case, the candidates $\widetilde{r}^P_{\geqslant4}\left(a_{7}\right)$, $\widetilde{r}^P_{\geqslant3}\left(a_{80}\right)$ and $\widetilde{r}^P_{\geqslant2}\left(a_{112}\right)$ are the only ones not dominated by any other rule in $\mathcal{Q}$ and, therefore, the rules are added to $\mathcal{R}^\geqslant$ as $r_1^\geqslant$, $r_2^\geqslant$ and $r_3^\geqslant$ with $\mbox{\textit{conf}}\left(r_1^\geqslant\right)=\mbox{\textit{conf}}\left(r_2^\geqslant\right)=\mbox{\textit{conf}}\left(r_3^\geqslant\right)=1$.
\end{description}
\paragraph{Rules induced by the subset $\bm{P=\left\{g_2\right\}}$}
The rules based on $P=\left\{g_2\right\}$ are built analogously. These are:
\begin{align*}
    r_4^{\geqslant}:\;&\mbox{if $g_2\left(a\right)\geqslant17.3433$, then $a$ is assigned to at least Class $4$, whose base unit is $a_{7}$},\\
	r_5^{\geqslant}:\;&\mbox{if $g_2\left(a\right)\geqslant15.02980$, then $a$ is assigned to at least Class $3$, whose base unit is $a_{25}$},\\
    r_6^{\geqslant}:\;&\mbox{if $g_2\left(a\right)\geqslant12.9640$, then $a$ is assigned to at least Class $2$, whose base unit is $a_{59}$}.
\end{align*}
\begin{description}
	\item[Steps 2-3:] The subset of criteria $P = \left\{g_2\right\}$, being the second element of $\mathcal{P}$, is selected. Moreover, the temporary set $\mathcal{Q}$ of generated rules from $P$ is initialized as an empty set.
	\item[Steps 4-14:] For each unit $a\in A^R$ being assigned by the user to $Cl^\geqslant_2$ and for each $t = 2, 3,4$, the algorithm computes the dominance set $D^+_P\left(a\right)$ and, consequently, the credibility $L_{\geqslant t}^P\left(a\right)$ for each unit to be classified to at least $Cl_t$ with respect to $P$. For instance, for $a_7$ ($g_2(a_7) = 17.3433$) and $t=4$ we have 
    \begin{align*}
    D^+_P\left(a_7\right) = \left\{a_{2},a_{3},a_{4},a_{5},a_{6},a_{7},a_{8},a_{10},a_{11},a_{12},a_{13},a_{15},a_{16},a_{23},a_{27},a_{33},a_{46},a_{48},a_{51},a_{52}\right\}
    \end{align*}
	all belonging to $Cl_4$. Consequently, the credibility $L_{\geqslant4}^P\left(a_7\right)=1$. Once it is checked that it is not dominated by any of $\mathcal{R}^{\geqslant} = \left\{r_1^\geqslant,r_2^\geqslant,r_3^\geqslant\right\}$, the candidate rule $\widetilde{r}_{\geqslant4}^P\left(a_7\right)$ is added to set $\mathcal{Q}$.
	\item[Step 15:] In set $\mathcal{Q}$, there are no duplicated rules, so all of them are stored.
	\item[Steps 16-18:] These lines iterate over the candidate rules in $\mathcal{Q}$ that were generated for the same subset $P$. For each candidate $\widetilde{r}_{t}^{P}\left(a\right)\in\mathcal{Q}$, the algorithm checks whether there exists another one for the same $P$ producing a stronger (i.e., more specific or dominating) rule. Only candidates not dominated by others are finally retained. In this case, the candidates $\widetilde{r}^P_{\geqslant4}\left(a_{7}\right)$, $\widetilde{r}^P_{\geqslant3}\left(a_{25}\right)$ and $\widetilde{r}^P_{\geqslant2}\left(a_{59}\right)$ are the only ones not dominated by any other rule in $\mathcal{Q}$ and, therefore, the rules are inserted in $\mathcal{R}^\geqslant$ as $r_4^\geqslant$, $r_5^\geqslant$ and $r_6^\geqslant$ with $\mbox{\textit{conf}}\left(r_4^\geqslant\right)=\mbox{\textit{conf}}\left(r_5^\geqslant\right)=\mbox{\textit{conf}}\left(r_6^\geqslant\right)=1$.
\end{description}
By repeating the above steps for all $P\in \mathcal{P}$, all rules $\mathcal{R}^{\geqslant}$ are induced.

\vspace{12pt}
Let us conclude this section with a comparison to DOMLEM \citep{greco2000algorithm}, which is the algorithm most commonly used to induce decision rules within DRSA. Similar to DOMLEM, the proposed algorithm relies on dominance-based rough approximations to induce decision rules. However, the two approaches differ substantially in their objectives and computational properties. DOMLEM is a heuristic, greedy algorithm designed to cover all decision examples in the dataset with a set of rules that is as small as possible. It does not guarantee that the resulting rule set is minimal in terms of cardinality, nor that all minimal rules consistent with the data are identified.
In contrast, the algorithm proposed in this paper is exhaustive: it systematically generates all minimal, non-contradictory rules in a single run, ensuring the completeness of the induced rule set. The computational complexity of our algorithm is exponential in the number of criteria, since it explores all possible combinations of condition attributes. Nevertheless, this is not a practical limitation for typical applications of composite indicators, which rarely involve more than 15–20 elementary indicators. In such cases, the algorithm runs efficiently even on standard personal computers. When the number of indicators exceeds this range, heuristic algorithms such as DOMLEM can be applied to obtain an approximate, yet computationally efficient, set of rules.

\subsection{Rules based classification}
Given sets of rules ${\cal R}^{\geqslant}$ and ${\cal R}^{\leqslant}$ explaining the classifications provided by the DM, for each $a\in A$, $s^{-}(a)$ and $s^{+}(a)$ are defined as follows:
\begin{equation}
\label{eq:class_rule}
\begin{array}{l}
s^{-}(a)=\left\{
\begin{array}{ll}
    1&\mbox{if $\gamma_i^{\geqslant}(a) = 0$, for all $r_i^{\geqslant}\in \mathcal{R}^{\geqslant}$},\\[2mm]
\displaystyle\max_{i:\; r_i^{\geqslant}\in {\cal R}^{\geqslant}} \left\{d_{i}^\geqslant\cdot \gamma^{\geqslant}_{i}(a)\right\}&\mbox{otherwise},
\end{array}
\right.\\[8mm]
s^{+}(a)=\left\{
\begin{array}{ll}
p&\mbox{if $\gamma_i^{\leqslant}(a) = 0$, for all $r_i^{\leqslant}\in \mathcal{R}^{\leqslant}$,}\\[2mm]
\displaystyle\min_{i:\; r_i^{\leqslant}\in {\cal R}^{\leqslant}} \left\{d_{i}^\leqslant\cdot \gamma^{\leqslant}_{i}(a) \right\}&\mbox{otherwise.}
\end{array}
\right.
\end{array}
\end{equation}
\begin{figure}[!h]
    \centering
    \begin{tikzpicture}[>=Latex,thick,x=0.5\textwidth]
\draw[->,ultra thick] (-0.6,0) -- (-0.6,6);
\draw (-0.63,1) -- (-0.57,1);
\draw (-0.63,2) -- (-0.57,2);
\draw (-0.63,3) -- (-0.57,3);
\draw (-0.63,4) -- (-0.57,4);
\draw (-0.63,5) -- (-0.57,5);
\draw[->] (-0.6,3.05) -- (-0.66,3.05) -- (-0.66,6);
\draw[->] (-0.6,3.05) -- (-0.54,3.05) -- (-0.54,6);
\node[right] at (-0.54,2.9) {$r_{i_3}^{\geqslant}$};
\draw[->] (-0.6,4.05) -- (-0.73,4.05) -- (-0.73,6);
\draw[->] (-0.6,4.05) -- (-0.47,4.05) -- (-0.47,6);
\node[right] at (-0.47,3.9) {$r_{i_4}^{\geqslant}$};
\node[align=center] at (-0.6,-0.6) {$s^{-}\left(a\right)=4$\\
	$s^{+}\left(a\right)=5$};
\node[align=center] at (-0.6,-1.4) {Instance 1};
\draw[->,ultra thick] (-0.2,0) -- (-0.2,6);
\draw (-0.23,1) -- (-0.17,1);
\draw (-0.23,2) -- (-0.17,2);
\draw (-0.23,3) -- (-0.17,3);
\draw (-0.23,4) -- (-0.17,4);
\draw (-0.23,5) -- (-0.17,5);
\draw[->] (-0.2,2.95) -- (-0.32,2.95) -- (-0.32,0);
\draw[->] (-0.2,2.95) -- (-0.08,2.95) -- (-0.08,0);
\node[right] at (-0.08,3.1) {$r_{i_3}^{\leqslant}$};
\draw[->] (-0.2,3.95) -- (-0.26,3.95) -- (-0.26,0);
\draw[->] (-0.2,3.95) -- (-0.14,3.95) -- (-0.14,0);
\node[right] at (-0.14,4.1) {$r_{i_4}^{\leqslant}$};
\node[align=center] at (-0.2,-0.6) {$s^{-}\left(a\right)=1$\\
	$s^{+}\left(a\right)=3$};
\node[align=center] at (-0.2,-1.4) {Instance 2};
\draw[->,ultra thick] (0.2,0) -- (0.2,6);
\draw (0.17,1) -- (0.23,1);
\draw (0.17,2) -- (0.23,2);
\draw (0.17,3) -- (0.23,3);
\draw (0.17,4) -- (0.23,4);
\draw (0.17,5) -- (0.23,5);
\draw[->] (0.2,3.05) -- (0.26,3.05) -- (0.26,6);
\draw[->] (0.2,3.05) -- (0.14,3.05) -- (0.14,6);
\node[right] at (0.26,2.95) {$r_{i_3}^{\geqslant}$};
\draw[->] (0.2,3.95) -- (0.33,3.95) -- (0.33,0);
\draw[->] (0.2,3.95) -- (0.07,3.95) -- (0.07,0);
\node[right] at (0.33,4.1) {$r_{i_4}^{\leqslant}$};
\node[align=center] at (0.2,-0.6) {$s^{-}\left(a\right)=3$\\
	$s^{+}\left(a\right)=4$};
\node[align=center] at (0.2,-1.4) {Instance 3};
\draw[->, ultra thick] (0.6,0) -- (0.6,6);
\draw (0.57,1) -- (0.63,1);
\draw (0.57,2) -- (0.63,2);
\draw (0.57,3) -- (0.63,3);
\draw (0.57,4) -- (0.63,4);
\draw (0.57,5) -- (0.63,5);
\draw[->] (0.6,4.05) -- (0.68,4.05) -- (0.68,6);
\draw[->] (0.6,4.05) -- (0.52,4.05) -- (0.52,6);
\node[right] at (0.68,3.95) {$r_{i_4}^{\geqslant}$};
\draw[->] (0.6,2.95) -- (0.68,2.95) -- (0.68,0);
\draw[->] (0.6,2.95) -- (0.52,2.95) -- (0.52,0);
\node[right] at (0.68,2.95) {$r_{i_3}^{\leqslant}$};
\node[align=center] at (0.6,-0.6) {$s^{-}\left(a\right)=4$\\
	$s^{+}\left(a\right)=3$};
\node[align=center] at (0.6,-1.4) {Instance 4};
\node[right] at (-1,1) {};
\node[right] at (-1,2) {$d_{i_2}^{\geqslant}=2$};
\node[right] at (-1,3) {$d_{i_3}^{\geqslant}=3$};
\node[right] at (-1,4) {$d_{i_4}^{\geqslant}=4$};
\node[right] at (-1,5) {$d_{i_5}^{\geqslant}=5$};
\node[left] at (1,1) {$d_{i_1}^{\leqslant}=1$};
\node[left] at (1,2) {$d_{i_2}^{\leqslant}=2$};
\node[left] at (1,3) {$d_{i_3}^{\leqslant}=3$};
\node[left] at (1,4) {$d_{i_4}^{\leqslant}=4$};
\node[left] at (1,5) {};
\draw[dashed,opacity=0.2] (-0.8,1) -- (0.8,1);
\draw[dashed,opacity=0.2] (-0.8,2) -- (0.8,2);
\draw[dashed,opacity=0.2] (-0.8,3) -- (0.8,3);
\draw[dashed,opacity=0.2] (-0.8,4) -- (0.8,4);
\draw[dashed,opacity=0.2] (-0.8,5) -- (0.8,5);
\draw[dashed,opacity=0.8] (-0.4,-1.2) -- (-0.4,6);
\draw[dashed,opacity=0.8] (0,-1.2) -- (0,6);
\draw[dashed,opacity=0.8] (0.4,-1.2) -- (0.4,6);
\end{tikzpicture}
\caption{Graphical representation of $s^-$ and $s^+$ computation}
\label{fig:classification_rules}
\end{figure}
Let us observe that:
\begin{itemize}
\item The classification of units given by (\ref{eq:class_rule}) is strictly dependent on the sets of \enquote{at-least} and \enquote{at-most} rules used, that is, ${\cal R}^{\geqslant}$ and ${\cal R}^{\leqslant}$;
\item Applying the classification rule according to (\ref{eq:class_rule}), it is possible that $s^{-}(a)>s^{+}(a)$ for some $a\in A\setminus A^R$, which means a contradictory assignment.
\end{itemize}

\begin{defn}\label{def:NonContradictory}
The classification of units from $A$ produced by (\ref{eq:class_rule}) considering ${\cal R}^{\geqslant}$ and ${\cal R}^{\leqslant}$ is \textit{non-contradictory} if $s^{-}(a)\leqslant s^{+}(a)$, for all $a\in A$. Moreover, 
$$
\mathcal{C}_{{\cal R}^{\geqslant},{\cal R}^{\leqslant}}: A \rightarrow \left\{1,\ldots,p\right\}\times\left\{1,\ldots,p\right\},
$$
such that $\mathcal{C}_{{\cal R}^{\geqslant},{\cal R}^{\leqslant}}(a)=\left[s^-(a),s^+(a)\right]$, denotes the classification of units from $A$ produced by (\ref{eq:class_rule}) considering ${\cal R}^{\geqslant}$ and ${\cal R}^{\leqslant}$.
\end{defn}

In Figure \ref{fig:classification_rules}, we illustrate how $s^-(a)$ and $s^{+}(a)$ are derived from eq. (\ref{eq:class_rule}). We assume a classification into five classes where: 
\begin{itemize}
\item An \enquote{at-least} rule $r_i^{\geqslant}\in{\cal R}^{\geqslant}$ can assign a unit to at least Class $d^{\geqslant}_{i}$, so that $d^{\geqslant}_{i}\in\{2,3,4,5\},$
\item An \enquote{at-most} rule $r_i^{\leqslant}\in{\cal R}^{\leqslant}$ can assign a unit to at most Class $d^{\leqslant}_{i}$, so that $d^{\leqslant}_{i}\in\{1,2,3,4\}.$
\end{itemize}
We now consider four instances: 
\begin{description}
\item[Instance 1:] Unit $a$ matches only rules $r_{i_3}^{\geqslant}$ and $r_{i_4}^{\geqslant}$, with decisions $d_{i_3}^{\geqslant}=3$ and $d_{i_4}^{\geqslant}=4$. Consequently, $s^{-}(a)=4$, $s^{+}(a)=5$,
\item[Instance 2:] Unit $a$ matches only rules $r_{i_3}^{\leqslant}$ and $r_{i_4}^{\leqslant}$, with decisions $d_{i_3}^{\leqslant}=3$ and $d_{i_4}^{\leqslant}=4$. Consequently, $s^{-}(a)=1$, $s^{+}(a)=3$,
\item[Instance 3:] Unit $a$ matches rules $r_{i_3}^{\geqslant}$ and $r_{i_4}^{\leqslant}$, with decisions $d_{i_3}^{\geqslant}=3$ and $d_{i_4}^{\leqslant}=4$. Consequently, $s^{-}(a)=3$, $s^{+}(a)=4$,
\item[Instance 4:] Unit $a$ matches rules $r_{i_4}^{\geqslant}$ and $r_{i_3}^{\leqslant}$, with decisions $d_{i_4}^{\geqslant}=4$ and $d_{i_3}^{\leqslant}=3$. Consequently, $s^{-}(a)=4$, $s^{+}(a)=3$. This leads to a contradiction, since the lower bound exceeds the upper bound.
\end{description}

In the next Section, we shall present an algorithm for building the \enquote{maximal} and \enquote{minimal} sets of \enquote{at-least} and \enquote{at-most} rules providing a non-contradictory classification of units from $A$ on the basis of classifications of reference units given by the DM.

\section{Building the rule explanation of the composite indicator}\label{sec:algorithm}
\begin{algorithm*}
\setlength{\baselineskip}{18pt}
\caption{Pseudo-code representation \label{DRSAScoreAlgorithm}}
\begin{algorithmic}
\Require Performances of all units from set $A$ and DM's class assignment $\left[s^-_R(a),s^+_R(a)\right]$ for reference units $a\in A^{R}\subseteq A$.
\Ensure Maximal and minimal sets of rules classifying all units $a \in A$ in a non-contradictory way.
\Steps
\State 1: Apply DRSA to get the set of \enquote{at-least} $\left({\cal R}^{\geqslant}\right)$ and the set of \enquote{at-most} $\left({\cal R}^{\leqslant}\right)$ rules explaining the classifications provided in the input.
\State 2: Sort rules from ${\cal R}^{\geqslant} \cup {\cal R}^{\leqslant}$ lexicographically according to their \textit{relative support} and \textit{Bayesian confirmation measures}.
\State 3: Select the maximum number of rules from ${\cal R}^{\geqslant}\cup{\cal R}^{\leqslant}$, that is ${\cal R}_{selected}^{\geqslant}\cup{\cal R}_{selected}^{\leqslant}$, such that $\mathcal{C}_{{\cal R}_{selected}^{\geqslant},{\cal R}_{selected}^{\leqslant}}$ provides a non-contradictory classification of units from $A$.
\State 4: Check if for all $a\in A^{R}$, $\mathcal{C}_{{\cal R}_{selected}^{\geqslant},{\cal R}_{selected}^{\leqslant}}(a) = \left[s_{R}^{-}(a),s_{R}^{+}(a)\right]$. If this is not the case, update the last classification of reference units so that they are the same as in the input. For non-reference units, ensure that their class assignment respects the dominance principle. Denote by $\left[s_{final}^{-}(a),s_{final}^{+}(a)\right]$ the corrected final classification of all units $a\in A$.
\State 5: Apply DRSA to the final classification obtained in step 4 to get the maximal sets of rules (i.e., all possible rules) ${\cal R}_{maximal}^{\geqslant}$ and ${\cal R}_{maximal}^{\leqslant}$ explaining it.
\State 6: Find the minimal sets of rules (i.e., minimal cover sets) ${\cal R}_{minimal}^{\geqslant}$ and ${\cal R}_{minimal}^{\leqslant}$ explaining the final classification provided in step 4.
\end{algorithmic}
\end{algorithm*}
The methodology for building the rule explanation of the composite indicator is presented as Algorithm \ref{DRSAScoreAlgorithm}, both as a pseudo-code and as a block-scheme in Figure \ref{fig:flowchart_knonw_units}.
\begin{figure}[h!]
\centering
{
\begin{tikzpicture}[node distance=0.5cm and 1cm]
\hyphenpenalty=10000 \exhyphenpenalty=10000
\node (s1) [process, text width=11cm, align=center] {\textbf{Input:}\\
Performances of all units on all criteria are given. The DM provides classifications of a number of reference units};
\node (s2) [process, below=of s1, text width=11cm, align=center] {\textbf{Step 1:}\\
DRSA is applied to induce all \enquote{at-least} and \enquote{at-most} rules from the classification of reference units provided in the input};
\node (s3) [process, below=of s2, text width=11cm, align=center] {\textbf{Step 2:}\\
Rules are ordered lexicographically according to relative support and Bayesian confirmation measures};
\node (s4) [process, below=of s3, text width=11cm, align=center] {\textbf{Step 3:}\\ 
A maximum number of previously generated rules making non-contradictory classification of all considered units is selected};
\node (d1) [decision, below=of s4, text width=3cm, align=center] {Are all reference units classified as wished by DM?};
\node (s5) [process, left=of d1, text width=6cm, align=center] {\textbf{Step 4:}\\ 
Fix classification of reference units as specified by DM};
\node (s6) [process, below=of d1, text width=11cm, align=center] {\textbf{Step 5:}\\
DRSA is applied to induce all possible \enquote{at-least} and \enquote{at-most} rules from the classification of all considered units};
\node (s7) [process, below=of s6, text width=11cm, align=center] {\textbf{Step 6:}\\
Find minimal sets of \enquote{at-least} and \enquote{at-most} rules explaining the classification of all units by solving MILP problem (\ref{eq:MinimalRules}), and show them to DM};

\draw [arrow] (s1) -- (s2);
\draw [arrow] (s2) -- (s3);
\draw [arrow] (s3) -- (s4);
\draw [arrow] (s4) -- (d1);
\draw [arrow] (d1) -- node[above] {No} (s5);
\draw [arrow] (s5.south) |- (s6.west);
\draw [arrow] (d1) -- node[right] {Yes} (s6);
\draw [arrow] (s6) -- (s7);
\end{tikzpicture}
}
    \caption{Block-scheme representation of Algorithm \ref{DRSAScoreAlgorithm}}
    \label{fig:flowchart_knonw_units}
\end{figure}
A more detailed explanation of the algorithm steps follows:
\begin{description}
\item[Input:] There are given performances of all units from set $A$ on the considered criteria. The DM classifies reference units from $A^{R}\subseteq A$ according to their preferences. For each $a\in A^R$, $\left[s^{-}_{R}(a),s^{+}_{R}(a)\right]$ is the classification given by the DM to $a$, so that $s^{-}_{R}(a)\leqslant s^{+}_{R}(a)$.\\
Let us underline that the classifications given by the DM must not contradict the dominance principle so that if $a,b\in A^R$ and $a$ dominates $b$ ($a$ is at least as good as $b$ on all criteria and better on at least one of them), then, $s_{R}^{-}(a)\geqslant s_{R}^{-}(b)$ and $s_{R}^{+}(a)\geqslant s_{R}^{+}(b)$. If this is not the case, the assignments of $a$ and $b$ should be revised together with the DM. 
\item[1:] DRSA is applied to generate the sets of \enquote{at-least} and \enquote{at-most} rules $\left({\cal R}^{\geqslant}\; \mbox{and}\; {\cal R}^{\leqslant}\right)$ explaining the classifications provided in the input. Let us observe that DRSA generates rules on the basis of \enquote{precise} classifications given by the DM. Since, in our case, the DM can specify imprecise classifications for reference units, that is, $s_{R}^{-}(a)<s_{R}^{+}(a)$ for some $a\in A^R$, we need to apply DRSA to explain such classifications in the following way. To obtain \enquote{at-least} rules, we consider that each unit $a\in A^R$ is classified in $s_{R}^{-}(a)$, while to obtain \enquote{at-most} rules, we consider that $a$ is classified in $s_R^{+}(a)$.\\
For example, let us assume that $A^R = \left\{a,b\right\}$, $\left[s_{R}^{-}(a),s_{R}^{+}(a)\right] = \left[2,3\right]$ and $\left[s_{R}^{-}(b),s_{R}^{+}(b)\right] = \left[3,5\right]$. Then, to get \enquote{at-least} rules explaining this classification, DRSA is applied considering $a$ assigned to Class $2$ and $b$ to Class $3$, while to get \enquote{at-most} rules DRSA is applied considering $a$ assigned to Class $3$ and $b$ to Class $5$.
\item[2:] The rules from ${\cal R}^{\geqslant}\cup {\cal R}^{\leqslant}$ are sorted lexicographically based on their \textit{relative support} (see (\ref{eq:support_all})) and \textit{Bayesian confirmation measures} $S$ and $N$ (see (\ref{eq:bayesian})). Specifically, the rules are first arranged in non-increasing order according to their relative support. If two rules share the same relative support, they are then sorted by the confirmation measure $S$ in a non-increasing manner. Finally, if both the relative support and the confirmation measure $S$ are identical, the rules are ordered based on the confirmation measure $N$, also in a non-increasing order. In the unlikely event that two rules have the same relative support and confirmation measures, their order is determined randomly. In the following, we assume that rules from ${\cal R}^{\geqslant}\cup{\cal R}^{\leqslant}$ are ordered from $r_1$ to $r_{\left|{\cal R}^{\geqslant}\cup{\cal R}^{\leqslant}\right|}$ as described above. Moreover, let us observe that each $r_{i}$, with $i=1,\ldots,\left|{\cal R}^{\geqslant}\cup{\cal R}^{\leqslant}\right|$, belongs to ${\cal R}^{\geqslant}$ or ${\cal R}^{\leqslant}$.
\item[3:] Set ${\cal R}_{selected}^{\geqslant}={\cal R}_{selected}^{\leqslant} = \emptyset$. These will be the maximal subsets of rules from ${\cal R}^{\geqslant}$ and $ {\cal R}^{\leqslant}$ explaining the non-contradictory classification of units from $A$ (see Definition \ref{def:NonContradictory}).\\
For each $r_{i}$, $i=1,\ldots,\left|{\cal R}^{\geqslant}\cup {\cal R}^{\leqslant}\right|$, first, we add it to $\mathcal{R}_{selected}^{\geqslant}$ or $\mathcal{R}_{selected}^{\leqslant}$ (if $r_{i}\in \mathcal{R}^{\geqslant}$, $r_{i}$ is added to $\mathcal{R}_{selected}^{\geqslant}$, while if $r_{i}\in \mathcal{R}^{\leqslant}$, $r_{i}$ is added to $\mathcal{R}_{selected}^{\leqslant}$). If the classification of units from $A$ obtained by (\ref{eq:class_rule}) using rules from $\mathcal{R}_{selected}^{\geqslant}$ and $\mathcal{R}_{selected}^{\leqslant}$ is non-contradictory, then $r_{i}$ is maintained in the corresponding set to which it has been added, otherwise, it is removed. For each $a\in A$, let us denote by {$\mathcal{C}_{{\cal R}_{selected}^{\geqslant},{\cal R}_{selected}^{\leqslant}}(a)=\left[s^{-}(a),s^{+}(a)\right]$} the classification of $a\in A$ obtained by (\ref{eq:class_rule}) using ${\cal R}_{selected}^{\geqslant}$ and ${\cal R}_{selected}^{\leqslant}$.
\item[4:] The non-contradictory classification obtained using rules from $\mathcal{R}_{selected}^{\geqslant}$ and $\mathcal{R}_{selected}^{\leqslant}$ is based on the classification of reference units from $A^R$ provided in the input. Two cases can arise: 
\begin{itemize}
\item \textit{All reference units are assigned as specified by the DM}: for all $a\in A^R$, $\mathcal{C}_{{\cal R}_{selected}^{\geqslant},{\cal R}_{selected}^{\leqslant}}(a)=\left[s_{R}^{-}(a),s_{R}^{+}(a)\right]$;
\item \textit{Some reference unit is not assigned as specified by the DM}: there exists at least one $a\in A^R$ such that $\mathcal{C}_{{\cal R}_{selected}^{\geqslant},{\cal R}_{selected}^{\leqslant}}(a)\neq\left[s_{R}^{-}(a),s_{R}^{+}(a)\right]$.
\end{itemize}
Since we want to build a set of rules explaining the non-contradictory classification obtained in step 3 but respecting the exact classification of reference units given by the DM in the input, we apply DRSA to the classification $\left[s^{-}_{final}(a),s^{+}_{final}(a)\right]$ of all units $a\in A$ so that
{
$$
\left[s^{-}_{final}(a),s^{+}_{final}(a)\right]=
\left\{
\begin{array}{lll}
\left[s^-_R(a),s^+_R(a)\right] & \mbox{if} & a \in A^R, \\[3mm]
\left[s^-(a),s^+(a)\right] & \mbox{if} & a \in A\setminus A^R.  
\end{array}
\right.
$$}

We pay attention that this classification respects the dominance principle as explained in the input. Indeed, if $a\in A^R$ and $b\in A\setminus A^R$, it is possible that $a$ dominates $b$ but $s_{R}^{-}(a)<s^{-}(b)$ or $s_{R}^{+}(a)<s^{+}(b)$. In this case, $s^{-}(b)$ or $s^{+}(b)$ should be revised, in accordance with the DM, so that $s^{-}(b)\leqslant s_{R}^{-}(a)$ and $s^{+}(b)\leqslant s_{R}^{+}(a)$. Conversely, if $b$ dominates $a$ but $s_{R}^{-}(a)>s^{-}(b)$ or $s_{R}^{+}(a)>s^{+}(b)$, then, similarly to what was said before, the classification of $b$ should be adjusted so that $s^{-}(b)\geqslant s_{R}^{-}(a)$ and $s^{+}(b)\geqslant s_{R}^{+}(a)$. \\
Let us observe that this phenomenon can appear if and only if one of the units is a reference and the other is not. Indeed, on one hand, they cannot be both reference units since their original classification has been checked in the input, while, on other hand, they cannot be both non-reference units since generated rules respect the dominance relation.
\item[5:] Apply DRSA to get ${\cal R}_{maximal}^{\geqslant}$ and ${\cal R}_{maximal}^{\leqslant}$ explaining the \enquote{corrected} classification of units obtained in step 4, that is, $\mathcal{C}_{{\cal R}_{maximal}^{\geqslant},{\cal R}_{maximal}^{\leqslant}}(a)=\left[s_{final}^{-}(a),s_{final}^{+}(a)\right]$ for all $a\in A$. Also in this case, since the classification of some units could be imprecise, DRSA is applied as described in step 1.
\item[6:] In the previous step, we built the set of all possible rules explaining the final classification. For each $a\in A$, on one hand, the classification $s_{final}^{-}(a)>1$ is explained by at least one rule in ${\cal R}_{maximal}^{\geqslant}$, while, on other hand, the classification $s_{final}^{-}(a)<p$ is explained by at least one rule in ${\cal R}_{maximal}^{\leqslant}$. Since multiple rules can be used to justify the given classifications, our goal is to establish a minimal set of rules for the explanation. To this aim, the following MILP problem has to be solved: \begin{equation}\label{eq:MinimalRules}
\begin{array}{l}
\min\left[\displaystyle\sum_{r_{i}^{\geqslant} \in \mathcal{R}_{maximal}^{\geqslant}}\rho_{i}^{\geqslant}+\sum_{r_{i}^{\leqslant} \in \mathcal{R}_{maximal}^{\leqslant}}\rho_{i}^{\leqslant}\right], \;\;\mbox{subject to} \\[1cm]
\left.
\begin{array}{l}
\displaystyle\sum_{\substack{r_{i}^{\geqslant} \in \mathcal{R}_{maximal}^{\geqslant}: \\ d_{i}^{\geqslant}=s_{final}^{-}(a)}} \rho_{i}^{\geqslant}\cdot\gamma_{i}^{\geqslant}(a)\geqslant 1, \;\;\mbox{for all}\;\;a\in A: s_{final}^{-}(a)>1,\\[1,2cm] 
\displaystyle\sum_{\substack{r_{i}^{\leqslant} \in \mathcal{R}_{maximal}^{\leqslant}: \\ d_{i}^{\leqslant}=s_{final}^{+}(a)}} \rho_{i}^{\leqslant}\cdot\gamma_{i}^{\leqslant}(a)\geqslant 1, \;\;\mbox{for all}\;\;a\in A: s_{final}^{+}(a)<p,\\[1,2cm]
\rho_{i}^{\geqslant}\in\{0,1\},\;\;\mbox{for all}\;\;r_{i}^{\geqslant}\in{\cal R}_{maximal}^{\geqslant},\\[2mm]
\rho_{i}^{\leqslant}\in\{0,1\},\;\;\mbox{for all}\;\;r_{i}^{\leqslant}\in{\cal R}_{maximal}^{\leqslant},
\end{array}
\right\}E_{min}
\end{array}
\end{equation}
where:
\begin{itemize}
\item $p$ is the number of classes to which units could be assigned;
\item $\displaystyle\sum_{\substack{r_{i}^{\geqslant} \in \mathcal{R}_{maximal}^{\geqslant}: \\ d_{i}^{\geqslant}=s_{final}^{-}(a)}} \rho_{i}^{\geqslant}\cdot\gamma_{i}^{\geqslant}(a)\geqslant 1$ imposes that, for each unit $a$ such that $s^{-}_{final}(a)>1$, at least one rule $r_{i}^{\geqslant} \in \mathcal{R}_{maximal}^{\geqslant}$ satisfied by $a$ and assigning it to at least Class $s_{final}^{-}(a)$ is selected;
\item $\displaystyle\sum_{\substack{r_{i}^{\leqslant} \in \mathcal{R}_{maximal}^{\leqslant}: \\ d_{i}^{\leqslant}=s_{final}^{+}(a)}} \rho_{i}^{\leqslant}\cdot\gamma_{i}^{\leqslant}(a)\geqslant 1$ imposes that, for each unit $a$ such that $s^{+}_{final}(a)<p$, at least one rule $r_{i}^{\leqslant} \in \mathcal{R}_{maximal}^{\leqslant}$ satisfied by $a$ and assigning it to at most Class $s_{final}^{+}(a)$ is selected.
\end{itemize}
From a computational point of view, the solution of MILP problem (\ref{eq:MinimalRules}) is a vector of binary variables $\left[\left[\rho_{i}^{\geqslant*}\right]_{ r_i^{\geqslant}\in{\cal R}_{maximal}^{\geqslant}},\left[\rho_{i}^{\leqslant*}\right]_{ r_i^{\leqslant}\in{\cal R}_{maximal}^{\leqslant}}\right]$, such that $r_i^{\geqslant}\in{\cal R}_{maximal}^{\geqslant}$ $\left(r_i^{\leqslant}\in{\cal R}_{maximal}^{\leqslant}\right)$ is selected if $\rho_{i}^{\geqslant*}=1$ $\left(\rho_{i}^{\leqslant*}=1\right)$, while it is not selected if $\rho_{i}^{\geqslant*}=0$ $\left(\rho_{i}^{\leqslant*}=0\right)$. The solution, therefore, is a minimal set of rules taken from ${\cal R}_{maximal}^{\geqslant}$ and ${\cal R}_{maximal}^{\leqslant}$ providing the final classification. Denoting by ${\cal R}_{minimal}^{\geqslant}$ and ${\cal R}^{\leqslant}_{minimal}$ the sets composed of rules selected through (\ref{eq:MinimalRules}), that is 
$$
{\cal R}_{minimal}^{\geqslant}=\left\{r_{i}^{\geqslant}\in{\cal R}_{maximal}^{\geqslant}: \rho_{i}^{\geqslant*}=1\right\}\;\;\mbox{and}\;\;{\cal R}_{minimal}^{\leqslant}=\left\{r_{i}^{\leqslant}\in{\cal R}_{maximal}^{\leqslant}: \rho_{i}^{\leqslant*}=1\right\}
$$
the classification of units from $A$ obtained through (\ref{eq:class_rule}) using ${\cal R}_{minimal}^{\geqslant}$ and ${\cal R}_{minimal}^{\leqslant}$ is the same as the one obtained using ${\cal R}_{maximal}^{\geqslant}$ and ${\cal R}_{maximal}^{\leqslant}$, that is, 
$$
\mathcal{C}_{{\cal R}_{maximal}^{\geqslant},{\cal R}_{maximal}^{\leqslant}}(a)=\mathcal{C}_{{\cal R}_{minimal}^{\geqslant},{\cal R}_{minimal}^{\leqslant}}(a)=\left[s^{-}_{final}(a),s^{+}_{final}(a)\right],\;\mbox{for all}\; a\in A.
$$
\end{description}

Observe moreover, that together with minimization of the number of rules, other requirements can be imposed, such as maximization of the minimal support or the minimal confirmation measure. This can be achieved by adding constraints representing the required conditions to optimization problem (\ref{eq:MinimalRules}). For example, if some minimal support is required for each selected rule, one can add the following constraints:

$$
\rho^\geqslant_i \cdot supp^\geqslant_i + (1 - \rho^\geqslant_i) \cdot M \geqslant minsupp
$$

\noindent and 

$$
\rho^\leqslant_i \cdot supp^\leqslant_i + (1 - \rho^\leqslant_i) \cdot M \geqslant minsupp
$$

\noindent where $minsupp$ denotes the minimal required threshold for the decision rule support. Analogous constraints can be added to ensure minimal requirements for a confirmation measure.

\subsection{Illustrative example}
\label{sec:example_procedure_first}
{To show the applicability of our methodology, let us consider the case study of stock portfolio selection previously introduced in Section \ref{sec:scenario3_intro}.} Our aim is to classify the units into $3$ classes ordered form the worst ($C_1$) to the best ($C_3$) using the algorithm presented in Section \ref{sec:algorithm}.

Let us assume the DM provides the classification of some reference units $\left(\left[s_{R}^{-}(a),s_{R}^{+}(a)\right]\right)$ shown in Table \ref{tab:ExamplesAndClassifications_2} (input of Algorithm \ref{DRSAScoreAlgorithm}).
\begin{table}[h!]
\renewcommand{\arraystretch}{1.2}
\centering
\caption{Classification of reference units provided by the DM (in columns labeled $\left[s^{-}_{R}(a),s^{+}_{R}(a)\right]$) and classification of all units obtained by (\ref{eq:class_rule}). On the one hand, $\left[s^{-}(a),s^{+}(a)\right]$, for all $a\in A$,  is obtained using ${\cal R}_{selected}^{\geqslant}$ and ${\cal R}_{selected}^{\leqslant}$, while, on the other hand, $\left[s^{-}_{final}(a),s^{+}_{final}(a)\right]$, for all $a\in A$, is obtained using ${\cal R}_{minimal}^{\geqslant}$ and ${\cal R}_{minimal}^{\leqslant}$.}
\label{tab:ExamplesAndClassifications_2}
\setlength{\tabcolsep}{1pt}
\begin{tabular}{c|c|c|c||c|c|c|c}
\toprule
Unit & $\left[s^{-}_{R}(a),s^{+}_{R}(a)\right]$ & $\left[s^{-}(a),s^{+}(a)\right]$ & $\left[s_{final}^{-}(a),s_{final}^{+}(a)\right]$ & Unit & $\left[s^{-}_{R}(a),s^{+}_{R}(a)\right]$ & $\left[s^{-}(a),s^{+}(a)\right]$ & $\left[s_{final}^{-}(a),s_{final}^{+}(a)\right]$\\
\midrule
A1 & $\left[\bm{2},\bm{2}\right]$ & $\left[2,2\right]$ &  $\left[2,2\right]$ & A26 & $\left[\bm{3},\bm{3}\right]$ & $\left[2,3\right]$ & $\left[\bm{3},\bm{3}\right]$ \\ 
A2 &  & $\left[1,1\right]$ &  $\left[1,1\right]$ & A27 & & $\left[2,2\right]$ & $\left[2,2\right]$ \\ 
A3  & & $\left[2, 3\right]$ & $\left[2, 3\right]$ & A28 & & $\left[1,1\right]$ & $\left[1,1\right]$ \\ 
A4 & $\left[\bm{3},\bm{3}\right]$ & $\left[2,3\right]$ & $\left[\bm{3}, \bm{3}\right]$ & A29 & & $\left[2,2\right]$ & $\left[2,2\right]$ \\ 
A5 & & $\left[2,3\right]$ & $\left[2,3\right]$ & A30 & & $\left[1,1\right]$ & $\left[1,1\right]$ \\ 
A6 & & $\left[2,2\right]$ & $\left[2,2\right]$ & A31 & & $\left[2,3\right]$ & $\left[2,3\right]$ \\ 
A7 & & $\left[2,2\right]$ & $\left[2,2\right]$ & A32 & & $\left[1,1\right]$ & $\left[1,1\right]$ \\ 
A8 & & $\left[3,3\right]$ & $\left[3,3\right]$ & A33 & $\left[\bm{1},\bm{1}\right]$ & $\left[1,1\right]$ & $\left[1,1\right]$ \\ 
A9 & & $\left[2,2\right]$ & $\left[2,2\right]$ & A34 & & $\left[2,2\right]$ & $\left[2,2\right]$ \\ 
A10 & & $\left[2,2\right]$ & $\left[2,2\right]$ & A35 & & $\left[1,1\right]$ & $\left[1,1\right]$ \\ 
A11 & & $\left[2,2\right]$ & $\left[2,2\right]$ & A36 & $\left[\bm{2},\bm{2}\right]$ & $\left[2,2\right]$ & $\left[2,2\right]$ \\ 
A12 & $\left[\bm{1},\bm{1}\right]$ & $\left[1,1\right]$ & $\left[1,1\right]$ & A37 & & $\left[2,2\right]$ & $\left[2,2\right]$ \\ 
A13 & & $\left[2,2\right]$ & $\left[2,2\right]$ & A38 & & $\left[2,3\right]$ & $\left[2,3\right]$ \\ 
A14 & & $\left[2,2\right]$ & $\left[2,2\right]$ & A39 & & $\left[2,2\right]$ & $\left[2,2\right]$ \\ 
A15 & & $\left[2,3\right]$ & $\left[2,3\right]$ & A40 & & $\left[2,3\right]$ & $\left[2,3\right]$ \\ 
A16 & & $\left[2,2\right]$ & $\left[2,2\right]$ & A41 & & $\left[2,2\right]$ & $\left[2,2\right]$ \\ 
A17 & $\left[\bm{3},\bm{3}\right]$ & $\left[2,3\right]$ & $\left[\bm{3},\bm{3}\right]$ & A42 & & $\left[2,2\right]$ & $\left[2,2\right]$ \\ 
A18 & $\left[\bm{2},\bm{2}\right]$ & $\left[2,2\right]$ & $\left[2,2\right]$ & A43 & & $\left[1,2\right]$ & $\left[1,2\right]$ \\ 
A19 & & $\left[2,3\right]$ & $\left[2,3\right]$ & A44 & & $\left[2,3\right]$ & $\left[2,3\right]$ \\ 
A20 & & $\left[2,2\right]$ & $\left[2,2\right]$ & A45 & $\left[\bm{3},\bm{3}\right]$ & $\left[3,3\right]$ & $\left[3,3\right]$ \\ 
A21 & & $\left[2,2\right]$ & $\left[2,2\right]$ & A46 & & $\left[2,2\right]$ & $\left[2,2\right]$ \\ 
A22 & & $\left[2,3\right]$ & $\left[2,3\right]$ & A47 & & $\left[2,2\right]$ & $\left[2,2\right]$ \\ 
A23 & & $\left[2,2\right]$ & $\left[2,2\right]$ & A48 & & $\left[1,1\right]$ & $\left[1,1\right]$ \\ 
A24 & & $\left[2,3\right]$ & $\left[2,3\right]$ & A49 & & $\left[2,3\right]$ & $\left[2,3\right]$ \\ 
A25 & & $\left[1,2\right]$ & $\left[1,2\right]$ & A50 & & $\left[3,3\right]$ & $\left[3,3\right]$ \\ 
\bottomrule 
\end{tabular}
\end{table}
Using DRSA, we get $21$ \enquote{at-least} rules $\left(\left|{\cal R}^{\geqslant}\right|=21\right)$ and $17$ \enquote{at-most} rules $\left(\left|{\cal R}^{\leqslant}\right|=17\right)$ (step 1 of Algorithm \ref{DRSAScoreAlgorithm}).

After ordering the rules as described in step 2 of Algorithm (\ref{DRSAScoreAlgorithm}), we select 7 \enquote{at-least} rules from $\mathcal{R}^{\geqslant}$ $\left(\left|{\cal R}_{selected}^{\geqslant}\right|=7\right)$ and 13 \enquote{at-most} rules from $\mathcal{R}^{\leqslant}$ $\left(\left|{\cal R}_{selected}^{\leqslant}\right|=13\right)$ (step 3 of Algorithm \ref{DRSAScoreAlgorithm}) that once applied in (\ref{eq:class_rule}) give the non-contradictory classification of units $\left(\left[s^{-}(a),s^{+}(a)\right]\right)$ shown in Table \ref{tab:ExamplesAndClassifications_2}.

Looking at these classifications, the DM observes that few reference units are not classified as they had wished. In particular, A4, A17 and A26 are classified into $\left[2,3\right]$ while the DM stated that they should be assigned to Class $3$. We then adjust the classification obtained before so that the reference units are classified as wished by the DM $\left(\mbox{the changed classifications are put in bold in columns}\;\left[s^{-}_{final}(a),s^{+}_{final}(a)\right]\right)$.\\
DRSA is then applied to explain the \enquote{corrected} classifications obtaining $\left|\mathcal{R}_{maximal}^{\geqslant}\right|=74$, and $\left|\mathcal{R}_{maximal}^{\leqslant}\right|=73$. 

Even if the final classification is justifiable by the rules in $\mathcal{R}_{maximal}^{\geqslant}$ and $\mathcal{R}_{maximal}^{\leqslant}$, we solve the MILP problem (\ref{eq:MinimalRules}) to select the minimum number of \enquote{at-least} and \enquote{at-most} rules covering all the units, obtaining:
$$
{\cal R}_{minimal}^{\geqslant}=\left\{r_{35}^{\geqslant},r_{49}^{\geqslant},r_{69}^{\geqslant},r_{72}^{\geqslant}\right\}\;\;\mbox{and}\;\;{\cal R}_{minimal}^{\leqslant}=\left\{r_{1}^{\leqslant},r_{46}^{\leqslant},r_{56}^{\leqslant}\right\}.
$$

This means that $\left[s^{-}_{final}(a),s^{+}_{final}(a)\right]$ for all $a\in A$ shown in Table \ref{tab:ExamplesAndClassifications_2} can be explained considering only $4$ of the $74$ \enquote{at-least} rules and $3$ of the $73$ \enquote{at-most} rules. In particular, the minimal sets of rules are the following:\\[1mm]
    \textbf{\enquote{at-least} rules}
    \begin{itemize}
        \item $r^{\geqslant}_{35}$: if $g_{2}(a)\geqslant0.788$, then $a$ is assigned to at least Class 2 ($d_{35}^{\geqslant} =2$),
        \item $r^{\geqslant}_{49}$: if $g_{1}(a)\geqslant35.747$, and $g_{3}(a)\geqslant52.491$, and $g_{7}(a)\geqslant7.145$, then $a$ is assigned to at least Class 2 ($d_{49}^{\geqslant} =2$),
        \item $r^{\geqslant}_{69}$: if $g_{2}(a)\geqslant1.489$, and $g_{3}(a)\geqslant30.193$, then $a$ is assigned to at least Class 3 ($d_{69}^{\geqslant} =3$),
        \item $r^{\geqslant}_{72}$: if $g_{1}(a)\geqslant82.361$, and $g_{7}(a)\geqslant11.682$, then $a$ is assigned to at least Class 3 ($d_{72}^{\geqslant} = 3$).
    \end{itemize}
    \textbf{\enquote{at-most} rules}    
    \begin{itemize}
        \item $r^{\leqslant}_{1}$: if $g_{7}(a)\leqslant9.156$, then $a$ is assigned to at most Class 2 ($d_{1}^{\leqslant} =2$),
        \item $r^{\leqslant}_{46}$: if $g_{1}(a)\leqslant103.495$, and $g_{2}(a)\leqslant1.379$, and $g_{4}(a)\leqslant14.722$, then $a$ is assigned to at most Class 2 ($d_{46}^{\leqslant} =2$),
        \item $r^{\leqslant}_{56}$: if $g_{2}(a)\leqslant0.739$, and $g_{8}(a)\leqslant4.585$, then $a$ is assigned to at most Class 1 ($d_{56}^{\leqslant} =1$).
    \end{itemize}
Now, let us show how these rules can be used to explain the classification reported in Table \ref{tab:ExamplesAndClassifications_2}. To this aim let us consider units A2, A5 and A20 that are classified in $\left[1,1\right]$, $\left[2,3\right]$ and $\left[2,2\right]$, respectively. In particular:  
\begin{itemize}
\item A2:
\begin{itemize}
\item does not match any \enquote{at-least} rule;
\item matches $r_{1}^{\leqslant}$ and $r_{56}^{\leqslant}$ \enquote{at-most} rules.
\end{itemize}
Applying (\ref{eq:class_rule}), and considering ${\cal R}_{minimal}^{\geqslant}$ and ${\cal R}_{minimal}^{\leqslant}$, we can therefore obtain 
$$
\left[s_{final}^{-}\left(\mbox{A2}\right),s_{final}^{+}\left(\mbox{A2}\right)\right]=\left[1,1\right];
$$ 
\item A5:
\begin{itemize}
\item matches $r_{35}^{\geqslant}$ \enquote{at-least} rule;
\item does not match any \enquote{at-most} rule.
\end{itemize}
Applying (\ref{eq:class_rule}), and considering ${\cal R}_{minimal}^{\geqslant}$ and ${\cal R}_{minimal}^{\leqslant}$, we can therefore obtain  
$$
\left[s_{final}^{-}\left(\mbox{A5}\right),s_{final}^{+}\left(\mbox{A5}\right)\right]=\left[2,3\right];
$$ 
\item A20:
\begin{itemize}
\item matches $r_{35}^{\geqslant}$ \enquote{at-least} rule;
\item matches $r_{1}^{\leqslant}$ \enquote{at-most} rule.
\end{itemize}
Applying (\ref{eq:class_rule}), and considering ${\cal R}_{minimal}^{\geqslant}$ and ${\cal R}_{minimal}^{\leqslant}$, we can therefore obtain  
$$
\left[s_{final}^{-}\left(\mbox{A20}\right),s_{final}^{+}\left(\mbox{A20}\right)\right]=\left[2,2\right].
$$ 
\end{itemize}
Analogous justifications could be given for the classification of all other units.

\section{Classifying new units and explaining such classifications}\label{sec:Newalternative}
In this section, let us show how a new set of units $A_{new}$ that was not present at the beginning of the analysis (therefore $A\cap A_{new}=\emptyset$) could be classified using all rules explaining {$\left[s^{-}_{final}(a),s^{+}_{final}(a)\right]$ for all $a\in A$}. 

We will apply all rules from $\mathcal{R}^{\geqslant}_{maximal}$, $\mathcal{R}^{\leqslant}_{maximal}$ to classify all units from $A\cup A_{new}$, paying attention to reclassification of each unit $a\in A$ into $\left[s^{-}_{final}\left(a\right), s^{+}_{final}\left(a\right)\right]$ and non-contradictory classification of new units $A_{new}$. To this aim, we have to solve the following MILP problem: 

\begin{footnotesize}
\begin{subequations}
\label{eq:NewAssignment}
\renewcommand{\theequation}{\theparentequation.\arabic {equation}}
\begin{tikzpicture}[overlay, remember picture,thick]
  \draw [decorate,decoration={brace,amplitude=8pt,mirror}] 
    (17.5, -15) -- (17.5,-1.5) node[right, midway] {\hspace{1.5mm} $E_{new}$};
    \draw [decorate,decoration={brace,amplitude=8pt,mirror}] 
    (15, -12) -- (15,-5) node[right, midway] {$\;\;\;\forall x\in A_{new}$};
\end{tikzpicture}
\begin{align}
&\hspace{-20mm}\eta^*=\min \displaystyle\sum_{a\in A}\left[\eta^{\geqslant}(a)+\eta^{\leqslant}(a)\right], \;\;\mbox{subject to} \tag{\ref{eq:NewAssignment}}\\[5mm]
&\hspace{-20mm}\displaystyle\sum_{\substack{r_{i}^{\geqslant} \in \mathcal{R}_{maximal}^{\geqslant}: \\d_{i}^{\geqslant}=s_{final}^{-}(a)}} \rho_{i}^{\geqslant}\cdot\gamma_{i}^{\geqslant}(a)\geqslant 1-\eta^{\geqslant}(a), \;\;\mbox{for all}\;\;a\in A: s_{final}^{-}(a)>1, \label{eq:1}\\[2mm]
&\hspace{-20mm}\displaystyle\sum_{\substack{r_{i}^{\leqslant} \in \mathcal{R}_{maximal}^{\leqslant}: \\d_{i}^{\leqslant}=s_{final}^{+}(a)}} \rho_{i}^{\leqslant}\cdot\gamma_{i}^{\leqslant}(a)\geqslant 1-\eta^{\leqslant}(a), \;\;\mbox{for all}\;\;a\in A: s_{final}^{+}(a)<p,\label{eq:2}\\[2mm]
&\hspace{-20mm}s^{-}(x)\leqslant s^{+}(x),\label{eq:3}\\[2mm]
&\hspace{-20mm}s^{-}(x)\geqslant 1 \;\;\mbox{if}\;\;\displaystyle\sum_{r_{i}^{\geqslant} \in \mathcal{R}_{maximal}^{\geqslant}}\gamma_{i}^{\geqslant}(x)=0,\label{eq:4}\\[2mm]
&\hspace{-20mm}\left.
\begin{array}{l}
\displaystyle\sum_{r_{i}^{\geqslant} \in \mathcal{R}_{maximal}^{\geqslant}} \rho_{i}^{\geqslant}\cdot\gamma_{i}^{\geqslant}(x)\geqslant 1,\\[2mm]
s^{-}(x)\geqslant \rho_{i}^{\geqslant}\cdot\gamma_{i}^{\geqslant}(x)\cdot d_{i}^{\geqslant}\;\;
\mbox{for all}\;r_i^{\geqslant}\in{\cal R}_{maximal}^{\geqslant}
\end{array}
\right\} \mbox{if}\;\;\displaystyle\sum_{r_{i}^{\geqslant} \in \mathcal{R}_{maximal}^{\geqslant}}\gamma_{i}^{\geqslant}(x)>0\label{eq:5} \\[2mm]
&\hspace{-20mm}s^{+}(x)\leqslant p \;\;\mbox{if}\;\;\displaystyle\sum_{r_{i}^{\leqslant} \in \mathcal{R}_{maximal}^{\leqslant}}\gamma_{i}^{\leqslant}(x)=0,\label{eq:6} \\[2mm] 
&\hspace{-20mm}\left.
\begin{array}{l}
\displaystyle\sum_{r_{i}^{\leqslant} \in \mathcal{R}_{maximal}^{\leqslant}} \rho_{i}^{\leqslant}\cdot\gamma_{i}^{\leqslant}(x)\geqslant 1,\\[2mm]
s^{+}(x)\leqslant \rho_{i}^{\leqslant}\cdot\gamma_{i}^{\leqslant}(x)\cdot d_{i}^{\leqslant}+\left(1-\rho_{i}^{\leqslant}\right)\cdot p\;\;
\mbox{for all}\;r_i^{\leqslant}\in{\cal R}_{maximal}^{\leqslant}:\gamma_{i}^{\leqslant}(x)=1\\[2mm]
\end{array}
\right\}\mbox{if}\;\;\displaystyle\sum_{r_{i}^{\leqslant} \in \mathcal{R}_{maximal}^{\leqslant}}\gamma_{i}^{\leqslant}(x)>0,\label{eq:7} \\[2mm]
&\hspace{-20mm} \rho_{i}^{\geqslant}\in\{0,1\},\;\;\mbox{for all}\;\;r_{i}^{\geqslant}\in{\cal R}_{maximal}^{\geqslant} \label{eq:8}\\[2mm]
&\hspace{-20mm}\rho_{i}^{\leqslant}\in\{0,1\},\;\;\mbox{for all}\;\;r_{i}^{\leqslant}\in{\cal R}_{maximal}^{\leqslant}\label{eq:9}\\[2mm]
&\hspace{-20mm}\eta^{\geqslant}(a)\in\{0,1\},\;\;\mbox{for all}\;\;a\in A,\label{eq:10}\\[2mm]
&\hspace{-20mm}\eta^{\leqslant}(a)\in\{0,1\},\;\;\mbox{for all}\;\;a\in A,\label{eq:11}
\end{align}
\end{subequations}
\end{footnotesize}

where the constraints have the following meaning: 
\begin{description}
\item[{$\left[\ref{eq:1}\right]$}] imposes that, for all $a\in A$ such that $s^{-}_{final}(a)>1$, at least one rule $r_i^{\geqslant}\in{\cal R}^{\geqslant}_{maximal}$ satisfied by $a$ and having decision part $d_i^{\geqslant}$ equal to $s^{-}_{final}(a)$ has to be selected. If $\eta^{\geqslant}(a)=0$, then this requirement is satisfied, while, in case of $\eta^{\geqslant}(a)=1$, it is not, because the constraint is always satisfied even if $a$ is not assigned to $s^{-}_{final}\left(a\right)$;
\item[{$\left[\ref{eq:2}\right]$}] imposes that, for all $a\in A$ such that $s^{+}_{final}(a)<p$, at least one rule $r_i^{\leqslant}\in{\cal R}^{\leqslant}_{maximal}$ satisfied by $a$ and having decision part $d_i^{\leqslant}$ equal to $s^{+}_{final}(a)$ has to be selected. If $\eta^{\leqslant}(a)=0$, then this requirement is satisfied, while, in case of $\eta^{\leqslant}(a)=1$, it is not, because the constraint is always satisfied even if $a$ is not assigned to $s^{+}_{final}\left(a\right)$;
\item[{$\left[\ref{eq:3}\right]$}] imposes that the lowest class ($s^{-}(x)$) and the highest class ($s^{+}(x)$) to which $x\in A_{new}$ is assigned are non-contradictory;
\item[{$\left[\ref{eq:4}\right]$}] imposes that if $x\in A_{new}$ does not match any $r_i^{\geqslant}\in{\cal R}_{maximal}^{\geqslant}$ $\left(\displaystyle\sum_{r_{i}^{\geqslant} \in \mathcal{R}_{maximal}^{\geqslant}}\gamma_{i}^{\geqslant}(x)=0\right)$, then $s^{-}(x)$ at least equal to 1;
\item[{$\left[\ref{eq:5}\right]$}] is composed of two different constraints. The first $\left(\displaystyle\sum_{r_{i}^{\geqslant} \in \mathcal{R}_{maximal}^{\geqslant}} \rho_{i}^{\geqslant}\cdot\gamma_{i}^{\geqslant}(x)\geqslant 1\right)$ imposes that at least one $r_i^{\geqslant}\in{\cal R}_{maximal}^{\geqslant}$ satisfied by $x$ should be selected. The second
$$\left(s^{-}(x)\geqslant \rho_{i}^{\geqslant}\cdot\gamma_{i}^{\geqslant}(x)\cdot d_{i}^{\geqslant}\;\;
\mbox{for all}\;r_i^{\geqslant}\in{\cal R}_{maximal}^{\geqslant}\right)$$
imposes that $s^{-}(x)$ is at least equal to the highest class $\left(d_i^{\geqslant}\right)$ of the rules selected $\left(\rho_{i}^{\geqslant}=1\right)$ and satisfied by $x$ $\left(\gamma_{i}^{\geqslant}(x)=1\right)$;
\item[{$\left[\ref{eq:6}\right]$}] imposes that if $x\in A_{new}$ does not match any $r_i^{\leqslant}\in{\cal R}_{maximal}^{\leqslant}$ $\left(\displaystyle\sum_{r_{i}^{\leqslant} \in \mathcal{R}_{maximal}^{\leqslant}}\gamma_{i}^{\leqslant}(x)=0\right)$, then $s^{+}(x)$ must be at most equal to $p$;
\item[{$\left[\ref{eq:7}\right]$}] is composed of two different constraints. The first $\left(\displaystyle\sum_{r_{i}^{\leqslant} \in \mathcal{R}_{maximal}^{\leqslant}} \rho_{i}^{\leqslant}\cdot\gamma_{i}^{\leqslant}(x)\geqslant 1\right)$ imposes that at least one $r_i^{\leqslant}\in{\cal R}_{maximal}^{\leqslant}$ satisfied by $x$ should be selected. The second 
$$
\left(s^{+}(x)\leqslant \rho_{i}^{\leqslant}\cdot\gamma_{i}^{\leqslant}(x)\cdot d_{i}^{\leqslant}+\left(1-\rho_{i}^{\leqslant}\right)\cdot p\;\;
\mbox{for all}\;r_i^{\leqslant}\in{\cal R}_{maximal}^{\leqslant}:\gamma_{i}^{\leqslant}(x)=1\right)
$$
imposes that $s^{+}(x)$ is at most equal to the lowest class $\left(d_i^{\leqslant}\right)$ of the rules selected $\left(\rho_{i}^{\leqslant}=1\right)$ and satisfied by $x$ $\left(\gamma_{i}^{\leqslant}(x)=1\right)$. In this way, if $\rho_i^{\leqslant}=\gamma_{i}^{\leqslant}(x)=1$ ($r_{i}^{\leqslant}\in{\cal R}_{maximal}^{\leqslant}$ is selected and satisfied by $x$), then $s^{+}(x)\leqslant d_i^{\leqslant}$, while if $\rho_i^{\leqslant}=0$ and $\gamma_{i}^{\leqslant}(x)=1$ ($r_{i}^{\leqslant}\in{\cal R}_{maximal}^{\leqslant}$ is not selected but it is satisfied by $x$), then, $s^{+}(x)\leqslant p$;
\item[{$\left[\ref{eq:8}\right]$}-{$\left[\ref{eq:11}\right]$}] impose that $\rho_{i}^{\geqslant}$, $\rho_{i}^{\leqslant}$, $\eta^{\geqslant}(a)$ and $\eta^{\leqslant}(a)$ are binary variables.
\end{description}
Solving (\ref{eq:NewAssignment}), the following vector of binary values is obtained:
$$
\left[\left[\rho_{i}^{\geqslant*}\right]_{ r_i^{\geqslant}\in{\cal R}_{maximal}^{\geqslant}},\left[\rho_{i}^{\leqslant*}\right]_{ r_i^{\leqslant}\in{\cal R}_{maximal}^{\leqslant}},\left[\eta^{\geqslant *}(a)\right]_{a\in A},\left[\eta^{\leqslant *}(a)\right]_{a\in A}\right].
$$ 
Two cases can be observed:
\begin{description}
\item[case 1)] $\eta^{*}=0$: new units from $A_{new}$ can be classified in a non-contradictory way using rules from ${\cal R}^{\geqslant}_{maximal}$ and ${\cal R}^{\leqslant}_{maximal}$, without modifying $\left[s_{final}^{-}(a),s_{final}^{+}(a)\right]$ with $a\in A$; let's observe, however, that some rules from $\mathcal{R}^{\geqslant}_{maximal}$ and $\mathcal{R}^{\leqslant}_{maximal}$ got value $\rho^{\geqslant*}_{i}=0$ or $\rho^{\leqslant*}_{i}=0$, which means that they were eliminated to make the classification of all units non-contradictory; this elimination is driven by the minimization of $\eta^{*}$ and might be excessive; to avoid this excessive elimination, we will further proceed with a new MILP problem (\ref{eq:milp_newalt_max});
\item[case 2)] $\eta^{*}>0$: to classify the new units from $A_{new}$ in a non-contradictory way using using rules from ${\cal R}^{\geqslant}_{maximal}$ and ${\cal R}^{\leqslant}_{maximal}$, some classifications {$\left[s_{final}^{-}(a),s_{final}^{+}(a)\right]$} have to be modified, in particular, those $s^{-}_{final}(a)$ for which $\eta^{\geqslant *}(a)=1$ and those $s^{+}_{final}(a)$ for which $\eta^{\leqslant *}(a)=1$.
\end{description}

In order to select the greatest number of \enquote{at-least} and \enquote{at-most} rules from ${\cal R}_{maximal}^{\geqslant}$ and ${\cal R}_{maximal}^{\leqslant}$ that, once applied according to (\ref{eq:class_rule}) provides non-contradictory classification of units from $A_{new}$, and keeps $s^{-}_{final}(a)$ for all $a\in A$ such that $\eta^{\geqslant*}(a)=0$ and $s^{+}_{final}(a)$ for all $a\in A$ such that $\eta^{\leqslant*}(a)=0$, one has to solve the following MILP problem in both cases considered above:
\begin{equation}
\label{eq:milp_newalt_max}
\begin{array}{l}
\;\;\rho^{**}=\max\left[ \displaystyle\sum_{r_{i}^{\geqslant} \in \mathcal{R}_{maximal}^{\geqslant}}\rho_{i}^{\geqslant}+\sum_{r_{i}^{\leqslant} \in \mathcal{R}_{maximal}^{\leqslant}}\rho_{i}^{\leqslant}\right], \;\;\mbox{subject to} \\[10mm]
\left.
\begin{array}{l}
\;E_{new}\\[2mm]
\left.
\begin{array}{l}
\eta^{\geqslant}(a)=\eta^{\geqslant*}(a), \\[2mm] 
\eta^{\leqslant}(a)=\eta^{\leqslant*}(a), 
\end{array}
\right\}\;\mbox{for all}\;a\in A\\
\end{array}
\right\}E_{new}^{*}.
\end{array}
\end{equation}
Observe, moreover, that the rules selected by (\ref{eq:milp_newalt_max}), once applied according to (\ref{eq:class_rule}), classify units from $a\in A$ to different classes than before if $\eta^{\geqslant*}(a)$ or $\eta^{\leqslant*}(a)$ get value $1$ (then, $s^{-}(a)\neq s_{final}^{-}(a)$ or $s^{+}(a)\neq s_{final}^{+}(a)$). Since we select the maximum number of rules that satisfy all other constraints, the resulting classifications ensure that: $i)$ the fewest possible number of classifications is modified, $ii)$ the variation of classifications is kept as small as possible. \\
The solution of (\ref{eq:milp_newalt_max}) is the following vector of binary values:
$$
\left[\left[\rho_{i}^{\geqslant**}\right]_{ r_i^{\geqslant}\in{\cal R}_{maximal}^{\geqslant}},\left[\rho_{i}^{\leqslant**}\right]_{ r_i^{\leqslant}\in{\cal R}_{maximal}^{\leqslant}}\right].
$$
It defines the maximal number of \enquote{at-least} and \enquote{at-most} rules satisfying constraints of (\ref{eq:NewAssignment}). In particular, the sets of rules selected by (\ref{eq:milp_newalt_max}) are
$$
{\cal R}_{maximal}^{\geqslant^{**}}=\left\{r_{i}^{\geqslant}\in{\cal R}_{maximal}^{\geqslant}: \rho_{i}^{\geqslant**}=1\right\}\;\;\mbox{and}\;\;{\cal R}_{maximal}^{\leqslant**}=\left\{r_{i}^{\leqslant}\in{\cal R}_{maximal}^{\leqslant}: \rho_{i}^{\leqslant**}=1\right\}.
$$

Once ${\cal R}_{maximal}^{\geqslant^{**}}$ and ${\cal R}_{maximal}^{\leqslant^{**}}$ are used according to (\ref{eq:class_rule}), the new classification of units from $A\cup A_{new}$ is obtained: 
{
$$
\mathcal{C}_{{\cal R}^{\geqslant**}_{maximal},{\cal R}^{\leqslant**}_{maximal}}(a)=\left[s^{-**}(a),s^{+**}(a)\right],\; \mbox{for all}\; a \in A \cup A_{new}.
$$}
Let us observe that, for all $a\in A$ if $\eta^{\geqslant*}(a)=0$, $s^{-**}(a)=s^{-}_{final}(a)$ and if $\eta^{\leqslant*}(a)=0$, $s^{+**}(a)=s^{+}_{final}(a)$.

Finally, in order to build a minimal set of rules explaining {$\left[s^{-**}(a),s^{+**}(a)\right]$ for all $a\in A\cup A_{new}$}, we have to solve the following MILP problem:
\begin{equation}\label{eq:milp_newalt_minimize}
\begin{array}{l}
\rho^{***}=\min\left[\displaystyle\sum_{r_{i}^{\geqslant} \in \mathcal{R}_{maximal}^{\geqslant**}}\rho_{i}^{\geqslant}+\sum_{r_{i}^{\leqslant} \in \mathcal{R}_{maximal}^{\leqslant**}}\rho_{i}^{\leqslant}\right], \;\;\mbox{subject to} \\[1cm]
\left.
\begin{array}{l}
\displaystyle\sum_{\substack{r_{i}^{\geqslant} \in \mathcal{R}_{maximal}^{\geqslant**}: \\ d_{i}^{\geqslant}=s^{-**}(a)}} \rho_{i}^{\geqslant}\cdot\gamma_{i}^{\geqslant}(a)\geqslant 1, \;\;\mbox{for all}\;\;a\in A\cup A_{new}: s^{-**}(a)>1,\\[1,2cm] 
\displaystyle\sum_{\substack{r_{i}^{\leqslant} \in \mathcal{R}_{maximal}^{\leqslant**}: \\ d_{i}^{\leqslant}=s^{+**}(a)}} \rho_{i}^{\leqslant}\cdot\gamma_{i}^{\leqslant}(a)\geqslant 1, \;\;\mbox{for all}\;\;a\in A\cup A_{new}: s^{+**}(a)<p,\\[1,2cm]
\rho_{i}^{\geqslant}\in\{0,1\},\;\;\mbox{for all}\;\;r_{i}^{\geqslant}\in{\cal R}_{maximal}^{\geqslant**},\\[2mm]
\rho_{i}^{\leqslant}\in\{0,1\},\;\;\mbox{for all}\;\;r_{i}^{\leqslant}\in{\cal R}_{maximal}^{\leqslant**}.
\end{array}
\right\}E_{min}^{***}
\end{array}
\end{equation}

The solution of (\ref{eq:milp_newalt_minimize}) is the following vector of binary values:
$$
\left[\left[\rho_{i}^{\geqslant***}\right]_{ r_i^{\geqslant}\in{\cal R}_{maximal}^{\geqslant}},\left[\rho_{i}^{\leqslant***}\right]_{ r_i^{\leqslant}\in{\cal R}_{maximal}^{\leqslant}}\right],
$$
such that $r_{i}^{\geqslant}\in{\cal R}^{\geqslant**}_{maximal}$ (respectively $r_{i}^{\leqslant}\in{\cal R}^{\leqslant**}_{maximal}$) is selected if $\rho_{i}^{\geqslant***}=1$ (respectively $\rho_{i}^{\leqslant***}=1$) while it is not selected if $\rho_{i}^{\geqslant***}=0$ (respectively $\rho_{i}^{\leqslant***}=0$). The new minimal sets 
$$
{\cal R}^{\geqslant***}_{minimal}=\left\{r_{i}^{\geqslant}\in \mathcal{R}_{maximal}^{\geqslant**}:\rho_{i}^{\geqslant***} = 1\right\}\;\;\mbox{and}\;\;{\cal R}^{\leqslant***}_{minimal}=\left\{r_{i}^{\leqslant}\in \mathcal{R}_{maximal}^{\leqslant**}:\rho_{i}^{\leqslant***} = 1\right\}
$$
are such that 
$$
\mathcal{C}_{{\cal R}^{\geqslant***}_{minimal},{\cal R}^{\leqslant***}_{minimal}}(a)=\mathcal{C}_{{\cal R}^{\geqslant**}_{maximal},{\cal R}^{\leqslant**}_{maximal}}(a)=\left[s^{-**}(a),s^{+**}(a)\right], \;\mbox{for all}\;a \in A \cup A_{new}.
$$

Before concluding this Section, let us observe that if the classification of new units from $A_{new}$ using rules from $\mathcal{R}^{\geqslant}_{maximal}$ and $\mathcal{R}^{\leqslant}_{maximal}$ according to (\ref{eq:class_rule}) is non-contradictory, then it is not necessary to solve MILP problems (\ref{eq:NewAssignment}) and (\ref{eq:milp_newalt_max}) because we would get the same sets of rules. Then, in order to find minimal sets of rules classifying $A\cup A_{new}$ in a non-contradictory way, it is sufficient to solve MILP problem (\ref{eq:milp_newalt_minimize}) only.

\subsection{Illustrative example (continuation)}
Let us continue the example presented in Section \ref{sec:example_procedure_first}. Assume that five new units, with performances on the eight criteria shown in Table \ref{tab:Newptf}, are to be classified. Note that these units do not come from \cite{Emamat2022}; the authors created them for didactic purposes by combining some performances of the fifty units from Table \ref{tab:ptf_evaluation_criteria} across the eight criteria considered. Therefore, $A_{new}=\{x_1,x_2,x_3,x_4,x_5\}$.
\begin{table}[h!]
\centering
\caption{Performances of new units ($x\in A_{new}$) to be classified on the eight criteria at hand}
\label{tab:Newptf}
\setlength{\tabcolsep}{4pt}
\begin{tabular}{c|cc>{\centering}m{0.15\textwidth}ccccc}
\toprule
Unit & Return ($\uparrow$)& Beta ($\uparrow$)& Net Profit Margin ($\uparrow$)& ROA ($\uparrow$)& ROE ($\uparrow$)& EPS ($\uparrow$)& P/E ($\uparrow$)& P/BV ($\uparrow$)\\
\midrule
$x_1$&-9.145 & 0.1 & -10.159 & -6.553 & -47.223 & -505.333 & 2.71 & 0.773 \\ 
$x_2$&66.644 & 1.1798 & 33.0816 & 17.0123 & 29.0545 & 1342.6334 & 28.2945 & 2.9225 \\ 
$x_3$&263.307 & 3.421 & 144.01 & 61.141 & 85.927 & 9331.333 & 250 & 20.786 \\ 
$x_4$&-9.145 & 0.1 & 144.01 & 61.141 & 85.927 & 9331.333 & 28.2945 & 2.9225 \\ 
$x_5$&11.6453 & 0.35342 & 1.6054 & -1.3874 & -37.0626 & 245.2862 & 21.5803 & 2.3002 \\
\bottomrule
\end{tabular}
\end{table}

Classifying the units from $A_{new}$ using ${\cal R}_{maximal}^{\geqslant}$ and ${\cal R}_{maximal}^{\leqslant}$ according to (\ref{eq:class_rule}), we get the classification shown in Table \ref{tab:new_Alt_Classification}. We can observe that, $x_2$, $x_4$ and $x_5$ are classified in a contradictory way since $\mathcal{C}_{{\cal R}^{\geqslant}_{maximal},{\cal R}^{\leqslant}_{maximal}}(x_2)=\left[3,2\right]$, $\mathcal{C}_{{\cal R}^{\geqslant}_{maximal},{\cal R}^{\leqslant}_{maximal}}(x_4)=\left[3,1\right]$ and $\mathcal{C}_{{\cal R}^{\geqslant}_{maximal},{\cal R}^{\leqslant}_{maximal}}(x_5)=\left[2,1\right]$. In order to correct this contradiction, it is necessary to delete some rules from ${\cal R}_{maximal}^{\geqslant}$ or ${\cal R}_{maximal}^{\leqslant}$. When deleting some rules from the two sets, it may happen that some units from $A$ are differently classified.
\begin{table}[h!]
\caption{Column 2 shows the classification of previously known unit A35 and new units $x\in A_{new}$ obtained using rules from ${\cal R}_{maximal}^{\geqslant}$ and ${\cal R}_{maximal}^{\leqslant}$, while column 3 shows their classifications obtained by rules from ${\cal R}_{maximal}^{\geqslant**}$ and ${\cal R}_{maximal}^{\leqslant**}$}
\label{tab:new_Alt_Classification}
\begin{center}
\begin{tabular}{c|c|c}
\toprule	Unit&$C_{\mathcal{R}^{\geqslant}_{maximal},\mathcal{R}^{\leqslant}_{maximal}}(\cdot)$&$C_{\mathcal{R}^{\geqslant**}_{maximal},\mathcal{R}^{\leqslant**}_{maximal}}(\cdot)$\\
\midrule	A35&$\left[1,1\right]$&$\left[1,2\right]$\\
\midrule	$x_1$&$\left[1,1\right]$&$\left[1,1\right]$\\
$x_2$&$\left[3,2\right]$&$\left[2,2\right]$\\
$x_3$&$\left[3,3\right]$&$\left[3,3\right]$\\
$x_4$&$\left[3,1\right]$&$\left[2,2\right]$\\
$x_5$&$\left[2,1\right]$&$\left[2,2\right]$\\
\bottomrule
\end{tabular}    
    \end{center}
\end{table}
Thus, starting from the final classifications shown in Table \ref{tab:ExamplesAndClassifications_2} and using the two sets of rules ${\cal R}_{maximal}^{\geqslant}$ and ${\cal R}_{maximal}^{\leqslant}$ explaining them, we solve the MILP problem (\ref{eq:NewAssignment}) to check if it is necessary to reclassify some units from $A$ to get a non-contradictory classification of units from $A_{new}$. Solving (\ref{eq:NewAssignment}), we get $\eta^*=1$ and, in particular, $\eta^{\leqslant*}(\mbox{A35})=1$. This means that to get a non-contradictory classification of units from $A_{new}$ it is necessary to modify $s^{+}_{final}\left(\mbox{A35}\right)$.

By solving (\ref{eq:milp_newalt_max}), the maximal sets of \enquote{at-least} and \enquote{at-most} rules applied according to (\ref{eq:class_rule}) provide a non-contradictory classification of units from $A_{new}$, keep the exact classification of units from $A$ for which $\eta^{\geqslant*}(a)=\eta^{\leqslant*}(a)=0$, and find a new classification of unit A35. In particular, we get $\rho^{**}=125$ rules, with  $\left|\mathcal{R}_{maximal}^{\geqslant**}\right| = 67$ \enquote{at-least} rules (over a total of $\left|\mathcal{R}_{maximal}^{\geqslant}\right| = 74$) and $\left|\mathcal{R}_{maximal}^{\leqslant**}\right| = 58$ \enquote{at-most} rules (over a total of $\left|\mathcal{R}_{maximal}^{\leqslant}\right| = 73$). Reclassifying all units from $A$ using these two sets of rules, one finds that
$$
\mathcal{C}_{{\cal R}^{\geqslant**}_{maximal},{\cal R}^{\leqslant**}_{maximal}}(a)=\left[s^{-**}(a),s^{+**}(a)\right]=
\left\{
\begin{array}{lll}
\left[s^{-}_{final}(a),s^{+}_{final}(a)\right] &\mbox{if} & a\in A\setminus\{\mbox{A35}\}\\[3mm]
[1,2] &\mbox{if} & a=\mbox{A35}
\end{array}
\right.
$$
while, new units from $A_{new}$ are classified in a non-contradictory way as shown in the last column of Table \ref{tab:new_Alt_Classification}.

Finally, to explain the last classification $\left[s^{-**}(a), s^{+**}(a)\right]$ of $A\cup A_{new}$ with a minimal set of \enquote{at-least} $\left({\cal R}_{minimal}^{\geqslant***}\right)$ and \enquote{at-most} $\left({\cal R}_{minimal}^{\leqslant***}\right)$ rules, that is
$$
\mathcal{C}_{{\cal R}^{\geqslant***}_{minimal},{\cal R}^{\leqslant***}_{minimal}}(a)=\left[s^{-**}(a),s^{+**}(a)\right]
$$
the MILP problem (\ref{eq:milp_newalt_minimize}) is solved indicating the following minimal sets of rules:\\[1mm]
\textbf{\enquote{at-least} rules}
    \begin{itemize}
        \item $r^{\geqslant}_{2}$: if $g_{7}(a)\geqslant13.216$, then $a$ is assigned to at least Class 2 ($d_{2}^{\geqslant} =2$),
        \item $r^{\geqslant}_{35}$: if $g_{2}(a)\geqslant0.788$, then $a$ is assigned to at least Class 2 ($d_{35}^{\geqslant} =2$),
        \item $r^{\geqslant}_{49}$: if $g_{1}(a)\geqslant35.747$, and $g_{3}(a)\geqslant52.491$, and $g_{7}(a)\geqslant7.145$, then $a$ is assigned to at least Class 2 ($d_{49}^{\geqslant} =2$),
        \item $r^{\geqslant}_{62}$: if $g_{2}(a)\geqslant1.489$, and $g_{3}(a)\geqslant30.193$, then $a$ is assigned to at least Class 3 ($d_{62}^{\geqslant} =3$),
        \item $r^{\geqslant}_{65}$: if $g_{1}(a)\geqslant82.361$, and $g_{7}(a)\geqslant11.682$, then $a$ is assigned to at least Class 3 ($d_{65}^{\geqslant} =3$).
    \end{itemize}
    \textbf{\enquote{at-most} rules}
    \begin{itemize}
        \item $r^{\leqslant}_{1}$: if $g_{7}(a)\leqslant9.156$, then $a$ is assigned to at most Class 2 ($d_{1}^{\leqslant} =2$),
        \item $r^{\leqslant}_{43}$: if $g_{1}(a)\leqslant79.686$, and $g_{2}(a)\leqslant1.124$, then $a$ is assigned to at most Class 2 ($d_{43}^{\leqslant} =2$),
        \item $r^{\leqslant}_{48}$: if $g_{1}(a)\leqslant67.379$, and $g_{2}(a)\leqslant1.276$, and $g_{3}(a)\leqslant72.002$, then $a$ is assigned to at most Class 2 ($d_{48}^{\leqslant} =2$),
        \item $r^{\leqslant}_{55}$: if $g_{2}(a)\leqslant0.593$, and $g_{7}(a)\leqslant8.333$, then $a$ is assigned to at most Class 1 ($d_{55}^{\leqslant} =1$),
        \item $r^{\leqslant}_{56}$: if $g_{2}(a)\leqslant0.617$, and $g_{7}(a)\leqslant7.491$, then $a$ is assigned to at most Class 1 ($d_{56}^{\leqslant} =1$),
        \item $r^{\leqslant}_{57}$: if $g_{2}(a)\leqslant0.739$, and $g_{7}(a)\leqslant5.797$, then $a$ is assigned to at most Class 1 ($d_{57}^{\leqslant} =1$).
    \end{itemize}
Reading these rules, one can understand, for example, that:
\begin{itemize}
\item A35 does not match any \enquote{at-least} rule (so that it is assigned to at least Class $1$), and it matches $r_{43}^{\leqslant}$ (so that it is assigned to at most Class $2$);
\item $x_1$ does not match any \enquote{at-least} rule (so that it is assigned to at least Class $1$), and it matches all \enquote{at-most} rules (so that it is assigned to at most Class $1$);
\item $x_2$ matches $r_{2}^{\geqslant}$ and $r_{35}^{\geqslant}$ \enquote{at-least} rules (so that it is assigned to at least Class $2$), and it matches $r_{48}^{\leqslant}$ \enquote{at-most} rule (so that it is assigned to at most Class $2$);
\item $x_3$ matches all \enquote{at-least} rules (so that it is assigned to at least Class $3$), and it does not match any \enquote{at-most} rules (so that it is assigned to at most Class $3$);
\item $x_4$ matches $r_{2}^{\geqslant}$ \enquote{at-least} rule (so that it is assigned to at least Class $2$), and it matches $r_{43}^{\leqslant}$ \enquote{at-most} rule (so that it is assigned to at most Class $2$);
\item $x_5$ matches $r_{2}^{\geqslant}$ \enquote{at-least} rule (so that it is assigned to at least Class $2$), and it matches $r_{43}^{\leqslant}$ and $r_{48}^{\leqslant}$ \enquote{at-most} rules (so that it is assigned to at most Class $2$).
\end{itemize}


\section{Handling missing values in decision-rule-based composite indicators}
\label{sec:Missing-value}
\subsection{General idea and methodological basis}
Missing values pose a major challenge in data analysis in general \citep{donders2006gentle} and in the construction of composite indicators in particular \citep{joint2008handbook}. Because missing data are so pervasive in empirical research, a wide range of strategies have been developed to address them. Traditional approaches include case deletion, mean substitution, and missing-indicator methods. More advanced techniques, such as multiple imputation, are now widely recommended because they reduce bias and improve efficiency. However, all imputation-based methods impose artificial assumptions on the dataset. These assumptions may distort the original information structure and make subsequent analyses harder to interpret.

An alternative is to use decision rules obtained through DRSA, which enable analysis to proceed without imputing missing values \citep{greco1999handling,GrecoMatarazzoSlowinski2001}. DRSA operates directly on incomplete data, thereby preserving the original informational content. It also enhances the transparency of composite indicator construction, since the resulting rules are both methodologically simple and easy to interpret. 
In DRSA, the basic principle of rule induction from classification datasets with missing values is straightforward: the condition part of each decision rule consists of elementary conditions defined on selected attributes that have no missing values in the \textit{base unit} supporting the rule, and it is assumed to hold for any possible value that could replace the missing ones. For example, suppose a unit has missing values on two elementary indicators, $g_{j_1}$ and $g_{j_2}$. In that case, it can still serve as a base unit for decision rules whose elementary conditions are built on all other indicators, excluding $g_{j_1}$ and $g_{j_2}$. This implies that the resulting rules remain valid for any possible values that $g_{j_1}$ and $g_{j_2}$ might take for the considered unit in place of the missing ones. Accordingly, we say that a decision rule derived in this way is supported by a unit from the training database (classification example) if the unit’s attribute values satisfy the rule’s elementary conditions or contain missing data on some attributes appearing in these conditions.

\subsection{Illustrative example}
To illustrate the decision rule approach to constructing composite indicators in the presence of missing data, consider an example based on data from the Programme for International Student Assessment (PISA) \citep{oecd_pisa2022}. Suppose the training database consists of twenty students described in Table 10, with missing values indicated by “?”, and the following variables: 
\begin{itemize}
\item \textbf{PV1MATH, PV1READ:} Plausible values (first draw) of student proficiency in mathematics and reading,
\item \textbf{ESCS:} Index of Economic, Social and Cultural Status,
\item \textbf{EAS:} Equity-Adjusted Score (removes the effect of ESCS, rescaled 0–100), 
\item \textbf{Class}: Classification derived from EAS, using the rule:
\begin{itemize}
\item if $EAS < 40$, then, Class L,
\item if $40 \leqslant EAS < 55$, then, Class M,
\item if $EAS \geqslant 55$, then, Class H.
\end{itemize}
\end{itemize}

\begin{table}[htbp]
\centering
\caption{Sample of student-level PISA data including plausible values in mathematics and reading, socio-economic index (ESCS), equity-adjusted scores (EAS), and classification (L, M, H).}
\label{tab:student_data}
\begin{tabular}{cccccc}
\hline
Student & PV1MATH & PV1READ & ESCS & EAS & Class \\
\hline
S1 & 412 & 435 & -0.25 & 46.1& M \\
S2 & 390 & ?   & -0.95 & 36.7 & L \\
S3 & 378 & 400 & -1.20 & 34.5 & L \\
S4 & 505 & 498 & 0.65  & 64.9 & H \\
S5 & 421 & 420 & ?     & 43.5& M \\
S6 & 360 & 355 & -1.10 & 25.1& L \\
S7 & 440 & ?   & 0.10  & 51.5 & M \\
S8 & 550 & 560 & 0.85  & 81.3 & H \\
S9 & 401 & 410 & ?     & 36.5& L \\
S10 & 430 & 420 & -0.30 & 47.9& M \\
S11 & 465 & 470 & 0.05  & 58.2& H \\
S12 & 382 & 395 & -0.80 &33.7& L \\
S13 & ?   & 445 & -0.10 &44.8& M \\
S14 & 520 & 480 & 0.90  &72.1& H \\
S15 & 410 & ?   & -0.40 &42.3& M \\
S16 & 470 & 465 & ?     & 57.5& H \\
S17 & 355 & 340 & -1.30 &20.9& L \\
S18 & 495 & 510 & 0.50  &66.7& H \\
S19 & 430 & 440 & -0.15 &49.6& M \\
S20 & 385 & 405 & -0.70 &35.8& L \\
\hline
\end{tabular}
\end{table}

In the following analysis, we substitute EAS with Class. Thus, Table \ref{tab:student_data} is composed of twenty classification examples described by condition attributes PV1MATH, PV1READ, and assigned by Class to three ordered decision classes: L, M, H. From this dataset, we induced all possible decision rules according to the principles and algorithms presented in \cite{greco1999handling,GrecoMatarazzoSlowinski2001} and mentioned above. Next to the rules presented below, we list the units supporting them; the base units from which the threshold values of the elementary conditions were derived are highlighted in \textbf{bold}.
\begin{itemize}
\item $r^{\geqslant}_{1}$: If $\mbox{PV1MATH}(a)\geqslant410$, then, $a$ is assigned to at least M ($d_{1}^{\geqslant} =\mbox{M}$); supported by classification examples: \{S1, S4, S5, S7 S8, S10, S11, \underline{S13}, S14, \textbf{S15}, S16, S18, S19\},
\item $r^{\geqslant}_{2}$: If $\mbox{PV1MATH}(a)\geqslant465$ and $\mbox{PV1READ}(a)\geqslant470$, then, $a$ is assigned to at least H ($d_{2}^{\geqslant} =\mbox{H}$); supported by classification examples: \{S4, S8, \textbf{S11}, S14, S18\},
\item $r^{\geqslant}_{3}$: If $\mbox{PV1MATH}(a)\geqslant470$ and $\mbox{PV1READ}(a)\geqslant465$, then, $a$ is assigned to at least H ($d_{3}^{\geqslant} =\mbox{H}$); supported by classification examples: \{S4, S8, S14, \textbf{S16}, S18\},
\item $r^{\geqslant}_{4}$: If $\mbox{PV1MATH}(a)\geqslant465$ and $\mbox{ESCS}(a)\geqslant0.05$, then, $a$ is assigned to at least H ($d_{4}^{\geqslant} =\mbox{H}$); supported by classification examples: \{S4, S8, \textbf{S11}, S14, \underline{S16}, S18\},
\item $r^{\geqslant}_{5}$: If $\mbox{PV1READ}(a)\geqslant420$ and $\mbox{ESCS}(a)\geqslant-0.3$, then, $a$ is assigned to at least M ($d_{5}^{\geqslant} =\mbox{M}$); supported by classification examples: \{S1, S4, \underline{S5}, \underline{S7}, S8, \textbf{S10}, S11, S13, S14, \underline{S16} S18, S19\},
\item $r^{\geqslant}_{6}$: If $\mbox{PV1READ}(a)\geqslant480$ and $\mbox{ESCS}(a)\geqslant0.9$, then, $a$ is assigned to at least H ($d_{6}^{\geqslant} =\mbox{H}$); supported by classification examples: \{\textbf{S14}\},
\item $r^{\geqslant}_{7}$: If $\mbox{PV1READ}(a)\geqslant498$ and $\mbox{ESCS}(a)\geqslant0.65$, then, $a$ is assigned to at least H ($d_{7}^{\geqslant} =\mbox{H}$; supported by classification examples: \{\textbf{S4}, S8\},
\item $r^{\geqslant}_{8}$: If $\mbox{PV1READ}(a)\geqslant510$ and $\mbox{ESCS}(a)\geqslant0.5$, then, $a$ is assigned to at least H ($d_{8}^{\geqslant} =\mbox{H}$); supported by classification examples: \{S8, \textbf{S18}\},
\item $r^{\leqslant}_{1}$: If $\mbox{PV1MATH}(a)\leqslant440$, then, $a$ is assigned to at most Class M ($d_{1}^{\leqslant} =\mbox{M}$); supported by classification examples: \{S1, S2, S3, S5, S6, \textbf{S7}, S9, S10, S12, \underline{S13}, S15, S17, S19, S20\},
\item $r^{\leqslant}_{2}$: If $\mbox{PV1READ}(a)\leqslant445$, then, $a$ is assigned to at most Class M ($d_{2}^{\leqslant} =\mbox{M}$); supported by classification examples: \{S1, \underline{S2}, S3, S5, S6, \underline{S7}, S9, S10, S12, \textbf{S13}, \underline{S15}, S17, S19, S20\},
\item $r^{\leqslant}_{3}$: If $\mbox{PV1MATH}(a)\leqslant401$ and $\mbox{PV1READ}(a)\leqslant410$, then, $a$ is assigned to at most Class L ($d_{3}^{\leqslant} =\mbox{L}$); supported by classification examples: \{\underline{S2}, S3, S6, \textbf{S9}, S12, S17, S20\},
\item $r^{\leqslant}_{4}$: If $\mbox{PV1MATH}(a)\leqslant390$ and $\mbox{ESCS}(a)\leqslant-0.95$, then, $a$ is assigned to at most Class L ($d_{4}^{\leqslant} =\mbox{L}$); supported by classification examples: \{\textbf{S2}, S3, S6, S17\},
\item $r^{\leqslant}_{5}$: If $\mbox{PV1MATH}(a)\leqslant385$ and $\mbox{ESCS}(a)\leqslant-0.7$, then, $a$ is assigned to at most Class L ($d_{5}^{\leqslant} =\mbox{L}$); supported by classification examples: \{S3, S6, S12, S17, \textbf{S20}\},
\item $r^{\leqslant}_{6}$: If $\mbox{PV1READ}(a)\leqslant405$ and $\mbox{ESCS}(a)\leqslant-0.7$, then, $a$ is assigned to at most Class L ($d_{6}^{\leqslant} =\mbox{L}$); supported by classification examples: \{\underline{S2}, S3, S6, S12, S17, \textbf{S20}\}.
\end{itemize}
To illustrate how the rules were induced, consider rule $r_{1}^{\geqslant}$ as an example. This rule is based on student S15. Rule $r_{1}^{\geqslant}$ is supported by S1, S4, S5, S7, S8, S10, S11, S13, S14, S15, S16, S18, S19. These students are supporting the rule because, for each of them, either the condition $\mbox{PV1MATH}\geqslant 410$ is satisfied or the value of $\mbox{PV1MATH}$ is missing. Conversely, other students do not satisfy the condition. In Section \ref{sec:induction_missing}, we present the general procedure for decision rule induction in the case of missing values, labeled as Algorithm \ref{algo:rulesgeneration_missing}.

The induced decision rules can be applied to classify units (students in this case) according to the principles presented in Section \ref{sec:Background}, and graphically represented in Figure \ref{fig:classification_rules}. The classification of a unit depends on the satisfied rules. In the case of units with missing values on some attributes, a rule is satisfied by a unit only if all of its conditions are satisfied by the unit’s non-missing values on the attributes appearing in the rule’s elementary conditions. Considering unit $a$, the lower and upper bounds of its class assignment, $s^-(a)$ and $s^+(a)$, are determined according to (\ref{eq:class_rule}). Following Definition \ref{def:NonContradictory}, the classification is said to be non-contradictory if $s^{-}(a) \leqslant s^{+}(a)$.

It should be noted that if the students from Table \ref{tab:student_data} were classified using the rules induced from this database, not all students supporting these rules would be matched by them during classification. For instance, rule $r_{1}^{\geqslant}$ is not satisfied by S13, although it is satisfied by all other students supporting it. Indeed, S13 has a missing value for PV1MATH; however, whatever value this attribute might take, the rule still holds (all students with $PV1MATH \geqslant 410$ are assigned to at least Class M). Therefore, S13 \textbf{supports} rule $r_{1}^{\geqslant}$.  Nevertheless, a student with a missing value on PV1MATH, like S13, cannot be classified as belonging to at least Class M according to rule $r_{1}^{\geqslant}$, since the missing value might be smaller than 410, in which case the minimal conditions required by $r_{1}^{\geqslant}$ would not be satisfied. Students who support a rule but are not matched by it during classification are indicated above by underlining.

Let us now illustrate how the induced rules can be applied to students who are not included in the training dataset and have no classification labels. These students are presented in Table \ref{tab: new students}. 
\begin{table}[ht]
\centering
\caption{Five additional PISA students with missing values in some attributes and unknown classification.}
\label{tab: new students}
\begin{tabular}{ccccc}
\hline
Student & PV1MATH & PV1READ & ESCS & Class \\
\hline
S21 & 420 & ?   & -0.60 & ? \\
S22 & ?  & 375  & -1.10 & ? \\
S23 & 510 & 480  &  0.30 & ? \\
S24 & 350 & ?   & -0.85 & ? \\
S25 & 460 & 440  & -0.20 & ? \\
\hline
\end{tabular}
\end{table}

Applying the rules generated from Table \ref{tab:student_data} on students presented in Table \ref{tab: new students}, we get the following classification results:
\begin{itemize}
    \item Student S21, matching rules $r_{1}^{\geqslant}$ and $r_{1}^{\leqslant}$, is assigned to at least Class M and at most Class M, that is, to Class M,
    \item Student S22, matching rules $r_{2}^{\leqslant}$ and $r_{6}^{\leqslant}$, is assigned by rule $r_{2}^{\leqslant}$ to at most Class M and by rule $r_{6}^{\leqslant}$ to at most Class L. Consequently, S22 is assigned to at least Class L and at most Class L, that is, to Class L,
    \item Student S23, matching rules $r_{1}^{\geqslant}$, $r_{2}^{\geqslant}$, $r_{3}^{\geqslant}$, $r_{4}^{\geqslant}$ and $r_{5}^{\geqslant}$, is assigned by rules $r_{1}^{\geqslant}$ and $r_{5}^{\geqslant}$ to at least Class M and by rules $r_{2}^{\geqslant}$, $r_{3}^{\geqslant}$ and $r_{4}^{\geqslant}$ to at least Class H. Consequently, S23 is assigned to at least Class H and at most Class H, that is, to Class H,
    \item Student S24, matching rules $r_{1}^{\leqslant}$ and $r_{5}^{\leqslant}$, is assigned by rule $r_{1}^{\leqslant}$ to at most Class M and by rule $r_{5}^{\leqslant}$ to at most Class L. Consequently, S24 is assigned to at least L and at most L, that is, to Class L,
    \item Student S25, matching rules $r_{1}^{\geqslant}$, $r_{5}^{\geqslant}$ and $r_{2}^{\leqslant}$ is assigned to at least Class M and at most Class M, that is, to Class M.
\end{itemize}

To find sets of rules $\mathcal{R}^{\geqslant}_{maximal}$, $\mathcal{R}^{\leqslant}_{maximal}$ ensuring the non-contradictory classification of all (training and new) students S1-S25, we solve the MILP problem (\ref{eq:NewAssignment}). The solution yields a value of $\eta^* = 0$, which means that the classification of all students from Table \ref{tab:student_data} remains unchanged and the classification of new students is consistent.
This result implies that no rule from the induced set has to be removed to consistently classify the new units. Thus, all induced rules enter the maximal sets of consistent rules. In this case, it is not necessary to solve the MILP problem (\ref{eq:milp_newalt_max}), as it would in any case yield a solution with $\rho^{**} = 14$ and:
$$
\rho_{1}^{\geqslant**}=\rho_{2}^{\geqslant**}=\rho_{3}^{\geqslant**}=\rho_{4}^{\geqslant**}=\rho_{5}^{\geqslant**}=\rho_{6}^{\geqslant**}=\rho_{7}^{\geqslant**}=\rho_{8}^{\geqslant**}=\rho_{1}^{\leqslant**}=\rho_{2}^{\leqslant**}=\rho_{3}^{\leqslant**}=\rho_{4}^{\leqslant**}=\rho_{5}^{\leqslant**}=\rho_{6}^{\leqslant**}=1,
$$
meaning that the maximal sets of selected rules are
$$
\mathcal{R}_{maximal}^{\geqslant**}=\left\{r_{1}^{\geqslant}, r_{2}^{\geqslant}, r_{3}^{\geqslant}, r_{4}^{\geqslant}, r_{5}^{\geqslant},r_{6}^{\geqslant},r_{7}^{\geqslant},r_{8}^{\geqslant}\right\},\quad
\mathcal{R}_{maximal}^{\leqslant**} = \left\{r_{1}^{\leqslant},r_{2}^{\leqslant},r_{3}^{\leqslant},r_{4}^{\leqslant},r_{5}^{\leqslant},r_{6}^{\leqslant}\right\}.
$$

Applying (\ref{eq:class_rule}) with $\mathcal{R}_{maximal}^{\geqslant**}$ and $\mathcal{R}_{maximal}^{\leqslant**}$ to classify students S1-S20 produces the same results as the original classification reported in Table \ref{tab:student_data}.
\begin{table}[h!]
\centering
\caption{Student's classification by the maximal and minimal sets of rules found by solving MILP problems (\ref{eq:milp_newalt_max}) and (\ref{eq:milp_newalt_minimize}).} \label{tab:missingdata_result}
\begin{tabular}{cc|cc} \toprule Student&&$s^{-**}\left(a\right)$&$s^{+**}\left(a\right)$\\
\midrule
S21&&M&M\\
S22&&L&L\\
S23&&H&H\\
S24&&L&L\\
S25&&M&M\\
\bottomrule
\end{tabular}
\end{table}
The new students are all classified without contradictions, as shown in Table \ref{tab:missingdata_result}.
To identify the minimal set of rules that yield the same classifications, we solve the MILP problem (\ref{eq:milp_newalt_minimize}), obtaining $\rho^{***} = 10$, with either
$$
\rho_{1}^{\geqslant***} = \rho_{3}^{\geqslant***} = \rho_{4}^{\geqslant***} = \rho_{5}^{\geqslant***} = \rho_{1}^{\leqslant***} = \rho_{2}^{\leqslant***} = \rho_{3}^{\leqslant***} = \rho_{4}^{\leqslant***} = 
\rho_{5}^{\leqslant***} = 
\rho_{6}^{\leqslant***} = 1,
$$
or
$$
\rho_{1}^{\geqslant***} = \rho_{2}^{\geqslant***} = \rho_{3}^{\geqslant***} = \rho_{5}^{\geqslant***} = \rho_{1}^{\leqslant***} = \rho_{2}^{\leqslant***} = \rho_{3}^{\leqslant***} = \rho_{4}^{\leqslant***} = 
\rho_{5}^{\leqslant***} = 
\rho_{6}^{\leqslant***} = 1,
$$
Thus, the minimal sets of selected rules are either
$$
\mathcal{R}_{minimal}^{\geqslant***}=\left\{r_{1}^{\geqslant}, r_{3}^{\geqslant}, r_{4}^{\geqslant},r_{5}^{\geqslant}\right\},\quad
\mathcal{R}_{minimal}^{\leqslant***} = \left\{r_{1}^{\leqslant}, r_{2}^{\leqslant}, r_{3}^{\leqslant}, r_{4}^{\leqslant},r_{5}^{\leqslant},r_{6}^{\leqslant}\right\},
$$
or
$$
\mathcal{R}_{minimal}^{\geqslant***}=\left\{r_{1}^{\geqslant}, r_{2}^{\geqslant}, r_{3}^{\geqslant},r_{5}^{\geqslant}\right\},\quad
\mathcal{R}_{minimal}^{\leqslant***} = \left\{r_{1}^{\leqslant}, r_{2}^{\leqslant}, r_{3}^{\leqslant}, r_{4}^{\leqslant},r_{5}^{\leqslant},r_{6}^{\leqslant}\right\}
.$$

\subsection{Decision rule induction in case of missing values}\label{sec:induction_missing} The induction of decision rules from datasets containing missing values requires specific adaptations of the algorithm presented in Subsection \ref{Rule_induction}. Below, we describe how \enquote{at-least} decision rules are generated in the presence of missing values. The corresponding algorithm is presented as Algorithm \ref{algo:rulesgeneration_missing}. Adaptations for \enquote{at-most} decision rules follow a similar rationale and are therefore omitted for conciseness.
\begin{algorithm}
\setlength{\baselineskip}{18pt}
\caption{Pseudo-code for \enquote{at-least} rules generation in case of missing values in the data set \label{algo:rulesgeneration_missing}}
\begin{normalsize}

\begin{algorithmic}[1]
\Require Performances
$g_j\left(a\right),\;j=1,\ldots,m$, class assignments $Cl_1,\ldots,Cl_p$ for reference units $a\in A^{R}\subseteq A$, minimal rule confidence $c$.
\Ensure Set of decision rules $\mathcal{R}^{\geqslant} = \left\{r_i^{\geqslant}\right\}$, with 
\begin{center}
$r_i^\geqslant:$ If $g_{j_1} \left(a\right) \geqslant \overline{q}^{i}_{j_1},$ and $\ldots,$ and $g_{j_s}\left(a\right) \geqslant \overline{q}^{i}_{j_s}$, then $a \in Cl^\geqslant_{t}$, \;$t=2,\ldots,p$
\end{center}
and confidence $\textit{conf}\left(r^{\geqslant}_i\right)$ $\left(P^\geqslant_i=\left\{g_{j_1}, \ldots, g_{j_s}\right\} \subseteq G \right)$
\Steps
\State ${\cal R}^{\geqslant}=\emptyset$, ${\cal P}=sort\left(2^G\setminus\left\{\emptyset\right\}\right)$
\For{$P\in \mathcal{P}$}
\State $\mathcal{Q} = \emptyset$
\vspace{1.5mm}
\Statex \hspace{-5mm}{\small3bis:} $\;\;\overline{A}^R=\left\{a\in A^R:g_j\left(a\right)\neq \; ?\;\mbox{for all}\;g_j\in P \right\}$
\For{$a\in \overline{A}^R \cap Cl_{2}^{\geqslant}$}
\For{$t = 2,\ldots,p$}
\State \makecell[l]{Compute $\displaystyle L_{\geqslant t}^P\left(a\right)=\frac{\left|\left\{a'\in D_P^{+}(a):a'\in Cl^{\geqslant}_t\right\}\right|}{\left|D_P^{+}(a)\right|}$\\
where $D_P^{+}(a)=\left\{a'\in A^R: g_j\left(a^{\prime}\right)\geqslant g_j(a)\;\mbox{or}\;g_j\left(a^{\prime}\right)= \; ?, \forall g_j\in P\right\}$}
\If{$L_{\geqslant t}^P\left(a\right)\geqslant c$}
\If{$\mathcal{R}^\geqslant=\emptyset$ or for all $r^{\geqslant}_i\in \mathcal{R}^{\geqslant}$ at least one among the following conditions holds:
\Statex \hspace{31mm} -  $C_1\left(r_i^\geqslant\right)$: $P_i^\geqslant\not\subseteq P$ 
\Statex \hspace{31mm} - $C_2\left(r_i^\geqslant\right)$: $P_i^\geqslant\subseteq P$ and there exists $g_j\in P_i^\geqslant$ such that $\overline{q}_{j}^{i}>g_j\left(a\right)$
\Statex \hspace{31mm} - $C_3\left(r_i^\geqslant\right)$: $P_i^\geqslant\subseteq P$, $\overline{q}_{j}^{i}\leqslant g_j\left(a\right)$ for all $g_j\in P_i^\geqslant$ and $d_i^\geqslant < t$ 
\Statex \hspace{31mm} - $C_4\left(r_i^\geqslant\right)$:  $P_i^\geqslant\subseteq P$, $\overline{q}_{j}^{i}\leqslant g_j\left(a\right)$ for all $g_j\in P_i^\geqslant$, $d_i^\geqslant \geqslant t$ and $L_{\geqslant t}^P\left(a\right)>\mbox{\textit{conf}}\left(r_i^\geqslant\right)$
\Statex \hspace{28mm}}
\State \makecell[l]{Generate rule $\widetilde{r}_{\geqslant t}^{P}\left(a\right)$:
\enquote{If $g_{j_1}\left(x\right)\geqslant g_{j_1}\left(a\right),\ldots,g_{j_s}\left(x\right)\geqslant g_{j_s}\left(a\right), \mbox{then } x\in Cl_{t}^{\geqslant}$}\\ with confidence $\mbox{\textit{conf}}\left(\widetilde{r}_{\geqslant t}^P\left(a\right)\right)=L_{\geqslant t}^P\left(a\right)$}
\State $\mathcal{Q} = \mathcal{Q}\cup \left\{\widetilde{r}_{\geqslant t}^P\left(a\right)\right\}$
\EndIf
\EndIf
\EndFor
\EndFor
\State Remove duplicate $\widetilde{r}_{\geqslant t}^P\left(a\right)$ from $\mathcal{Q}$, keeping only unique rows
\For{$\widetilde{r}_{\geqslant t}^P\left(a\right) \in \mathcal{Q}$}
\If{for all $\widetilde{r}_{\geqslant t^\prime}^P\left(a^\prime\right) \in \mathcal{Q}\setminus \widetilde{r}_{\geqslant t}^P\left(a\right)$ at least one among the following conditions holds:
\vspace{2mm}
\Statex \hspace{20mm} - $C_{1}^\prime\left(\widetilde{r}_{\geqslant t^\prime}^P\left(a^\prime\right)\right)$: there exists $g_j\in P$ such that $g_j\left(a^\prime\right)> g_j\left(a\right)$
\vspace{2mm}
\Statex \hspace{20mm} - 
$C_{2}^\prime\left(\widetilde{r}_{\geqslant t^\prime}^P\left(a^\prime\right)\right)$: $g_j\left(a^\prime\right)\leqslant g_j\left(a\right)$ for all $g_j\in P$ and $t^\prime<t$
\vspace{2mm}
\Statex \hspace{20mm} - $C_{3}^\prime\left(\widetilde{r}_{\geqslant t^\prime}^P\left(a^\prime\right)\right)$: $g_j\left(a^\prime\right)\leqslant g_j\left(a\right)$ for all $g_j\in P$, $t^\prime\geqslant t$ and $L_{\geqslant t}^P\left(a\right)>L_{\geqslant t^\prime}^{P}\left(a^\prime\right)$
\Statex \hspace{16mm}}
\State $\mathcal{R}^\geqslant=\mathcal{R}^\geqslant\cup \left\{\widetilde{r}_{\geqslant t}^{P}\left(a\right)\right\}$
\EndIf
\EndFor
\EndFor
\end{algorithmic}
\end{normalsize}
\end{algorithm}

Algorithms \ref{algo:rulesgeneration} and \ref{algo:rulesgeneration_missing} share the same fundamental structure for generating \enquote{at-least} decision rules. The main distinction is in the handling of missing performance values: Algorithm \ref{algo:rulesgeneration} assumes complete information, whereas Algorithm \ref{algo:rulesgeneration_missing} explicitly accommodates missing data.
The modifications in Algorithm \ref{algo:rulesgeneration_missing} are highlighted as follows:
\begin{itemize}
    \item \textbf{Step 3bis:} A new subset $\overline{A}^R = \{a \in A^R : g_j(a) \neq \; ? \text{ for all } g_j \in P\}$ is introduced. This step filters reference units to include only those with known performances for the current attribute subset $P$, ensuring that incomplete data do not affect rule generation.
    \item \textbf{Step 4:} Iteration and selection of reference units now operate over $\overline{A}^R$ instead of $A^R$. This ensures that only units with complete information for the current attribute subset are considered as base units generating rules.
    \item \textbf{Step $6$:} The dominance relation $D_P^+(a)$ is redefined to handle missing values. Specifically, $a^\prime \in D_P^+(a)$ if $g_j(a^\prime) \geqslant g_j(a)$ \emph{or} $g_j(a') = \; ?$ for all $g_j \in P$, in contrast to Step 4 of Algorithm \ref{algo:rulesgeneration} which assumes all values are known. 
\end{itemize}
All subsequent steps, including rule evaluation (Step 8) and rule addition (Steps 16-18), remain unchanged. 
Overall, these modifications allow Algorithm \ref{algo:rulesgeneration_missing} to maintain the original logic of \enquote{at-least} rule generation of Algorithm \ref{algo:rulesgeneration} while producing reliable decision rules in the presence of missing values, improving the robustness and applicability of the method in real-world scenarios.

\vspace{12pt}
\textbf{Example.} To illustrate the operation of Algorithm \ref{algo:rulesgeneration_missing} on the dataset presented in Table \ref{tab:student_data}, we describe how some rules with confidence 1 are obtained, following the main steps of the procedure.
Let us generate the subsets of criteria $P\subseteq G$, starting with subsets \(P\) of size 1, then of size 2, and so on (step 2), and we generate and test candidate rules at each stage.
\paragraph{Rules induced for subset $\bm{P=\left\{\mbox{PV1MATH}\right\}}$}
The induced rule based on $P=\left\{\mbox{PV1MATH}\right\}$:
\begin{center}
    $r_{1}^{\geqslant}:$ if $\mbox{PV1MATH}(a)\geqslant410$, then, $a$ is assigned to at least Class M, whose base unit is S15.
\end{center}
\begin{description}
\item[\textbf{Step 1}] $\mathcal{R}^\geqslant$ is initialized at $\emptyset$, and $\mathcal{P}$ is the ordered set of $2^G\setminus\emptyset$, that is 
\begin{align*}
\mathcal{P} = &\left\{\left\{\mbox{PV1MATH}\right\},\left\{\mbox{PV1READ}\right\},\left\{\mbox{ESCS}\right\},\left\{\mbox{PV1MATH},\mbox{PV1READ}\right\},\left\{\mbox{PV1MATH},\mbox{ESCS}\right\},\right.\\
&\left.\left\{\mbox{PV1READ},\mbox{ESCS}\right\},\left\{\mbox{PV1MATH},\mbox{PV1READ},\mbox{ESCS}\right\}\right\}
\end{align*}
  \item[\textbf{Steps 2-3bis.:}] The subset of criteria $P=\left\{\mbox{PV1MATH}\right\}$, being the first element of $\mathcal{P}$, is selected, and the reduced learning set $\overline{A}^R$ is built including only students without missing data for this attribute, that is, all units except S13. Moreover, the temporary set $\mathcal{Q}$ of generated rules for $P$ is initialized as an empty set.
  \item[\textbf{Steps 4-14:}] For each unit $a\in \overline{A}^R$ being assigned by the user to at least $Cl_2$ and for each $t = 2,\ldots,p$, the algorithm computes the dominance set $D_P^+(a)$ using the modified relation that handles \mbox{missing} values and, consequently, the credibility $L^{P}_{\geqslant t}(a)$ for each student. For instance, for $a=\mbox{S15}$ ($\mbox{PV1MATH}\left(S15\right) = 410$) and $t=2$, we have:
  $$
  D_P^+(\mbox{S15}) = \{\mbox{S1, S4, S5, S7, S8, S10, S11, S13, S14, S15, S16, S18, S19}\},
  $$
  all of them belong to $Cl^\geqslant_2$. Consequently, the credibility $L^P_{\geqslant 2}(\mbox{S15}) = 1$. For this reason, together with the fact that up to now $\mathcal{R}^{\geqslant}=\emptyset$, the candidate rule $\widetilde{r}_{\geqslant 2}^P\left(\mbox{S15}\right)$ is added to the temporary set $\mathcal{Q}$.
  \item[\textbf{Steps 15:}] In set $\mathcal{Q}$, there are duplicated rules induced by S10 and S19 since both of them have the same evaluation (430) on PV1MATH. One of them is deleted, so that
  \begin{align*}
  \mathcal{Q} = & \left\{ \widetilde{r}^P_{\geqslant 2}\left(S1\right), \widetilde{r}^P_{\geqslant 2}\left(S4\right), \widetilde{r}^P_{\geqslant 2}\left(S5\right), \widetilde{r}^P_{\geqslant 2}\left(S7\right), \widetilde{r}^P_{\geqslant 2}\left(S8\right), \widetilde{r}^P_{\geqslant 2}\left(S10\right),\right.\\
  &\left.\widetilde{r}^P_{\geqslant 2}\left(S11\right), \widetilde{r}^P_{\geqslant 2}\left(S14\right), \widetilde{r}^P_{\geqslant 2}\left(S15\right), \widetilde{r}^P_{\geqslant 2}\left(S16\right), \widetilde{r}^P_{\geqslant 2}\left(S18\right)\right\}.
  \end{align*}
  \item[\textbf{Steps 16-18:}] These lines iterate over the candidate rules in $\mathcal{Q}$ that were generated for the same subset $P$. For each candidate $\widetilde{r}_{\geqslant t}^P\left(a\right)\in\mathcal{Q}$, the algorithm checks whether there exists another one for the same $P$ producing a stronger (i.e., more specific or dominating) rule. Only candidates not dominated by others are finally retained. In this case, the candidate $\widetilde{r}_{\geqslant 2}^P\left(S15\right)$ is the only one not dominated by any other rule in ${\cal Q}$ and, therefore, the rule is inserted in $\mathcal{R}^{\geqslant}$ as $r_{1}^{\geqslant}$, with $\mbox{\textit{conf}}\left(r_{1}^{\geqslant}\right)=1$.
\end{description}
\paragraph{Rules induced for subset $\bm{P=\left\{\mbox{PV1MATH, PV1READ}\right\}}$}
The induced rules based on $P=\left\{\mbox{PV1MATH, PV1READ}\right\}$ are:
\begin{description}
\item[$r_{2}^{\geqslant}$:] if $\mbox{PV1MATH}(a)\geqslant465$, and $\mbox{PV1READ}(a)\geqslant470$, then, $a$ is assigned to at least Class H, whose base unit is S11
\item[$r_{3}^{\geqslant}$:] if $\mbox{PV1MATH}(a)\geqslant470$, and $\mbox{PV1READ}(a)\geqslant465$, then, $a$ is assigned to at least Class H, whose base unit is S16.
\end{description}
In both cases, considering $P = \left\{\mbox{PV1READ}\right\}$ and $P = \left\{\mbox{ESCS}\right\}$, no units have a credibility greater than or equal to $c = 1$; therefore, no rules have been generated. Hence, up to now, the set of induced rules is $\mathcal{R}^{\geqslant} = \left\{r_1^\geqslant\right\}$.
\begin{description}
\item[\textbf{Steps 2-3bis.:}] The subset of criteria $P=\left\{\mbox{PV1MATH, PV1READ}\right\}$, being the fourth element of $\mathcal{P}$, is selected, and the reduced learning set $\overline{A}^R$ is built including only students without missing data for these attributes, that is all units unless S2, S7, S13 and S15. Moreover, the temporary set $\mathcal{Q}$ of generated rules from $P$ is initialized as an empty set.
  \item[\textbf{Steps 4-14:}] For each unit $a\in \overline{A}^R$ being assigned by the user to at least $Cl_2$ and for each $t = 2,\ldots,p$, the algorithm computes the dominance set $D_P^+(a)$ using the modified relation that handles missing values and, consequently, the credibility $L^{P}_{\geqslant}\left(a\right)$ of each student. For instance, for $a=\mbox{S11}$ ($\mbox{PV1MATH}\left(S11\right) = 465$ and $\mbox{PV1READ}\left(S11\right) = 470$) and $t=2,3$, we have:
  $
  D_P^+(\mbox{S11}) = \{\mbox{S4, S8, S11, S14, S18}\}.
  $
  All of them belong to $Cl^\geqslant_3$. Consequently, the credibility $L^{P}_{\geqslant 2}(\mbox{S11}) = L^{P}_{\geqslant 3}(\mbox{S11}) = 1$. Once it is checked that it is not dominated by any rule from $R^\geqslant = \left\{r_1^\geqslant\right\}$, the candidate rule $\widetilde{r}_{\geqslant 3}^P\left(\mbox{S11}\right)$ is added to the temporary set $\mathcal{Q}$.
  \item[\textbf{Steps 15:}] In set $\mathcal{Q}$ there are no duplicated rules, so all of them are stored, thus $$\mathcal{Q} = \left\{\widetilde{r}^P_{\geqslant 3}\left(S4\right), \widetilde{r}^P_{\geqslant 3}\left(S8\right), \widetilde{r}^P_{\geqslant 3}\left(S11\right), \widetilde{r}^P_{\geqslant 3}\left(S14\right), \widetilde{r}^P_{\geqslant 3}\left(S16\right), \widetilde{r}^P_{\geqslant 3}\left(S18\right)\right\}.$$
  \item[\textbf{Steps 16-18:}] These lines iterate over the candidate rules in $\mathcal{Q}$ that were generated for the same subset $P$. For each candidate $\widetilde{r}_{\geqslant t}^P\left(a\right)\in\mathcal{Q}$, the algorithm checks whether there exists another one for the same $P$ producing a stronger (i.e., more specific or dominating) rule. Only candidates not dominated by others are finally retained.  
  In this case, the candidates $\widetilde{r}_{\geqslant 2}^P\left(S11\right)$ and $\widetilde{r}_{\geqslant 3}^P\left(S16\right)$ are not dominated by any other rule in ${\cal Q}$ and, therefore, both of them are inserted in $\mathcal{R}^{\geqslant}$ as $r_{2}^{\geqslant}$ and $r_{3}^{\geqslant}$, with $\mbox{\textit{conf}}\left(r_{2}^{\geqslant}\right)=\mbox{\textit{conf}}\left(r_{3}^{\geqslant}\right)=1$, respectively.
\end{description}
By repeating the above steps for all $P\in \mathcal{P}$, all rules $\mathcal{R}^{\geqslant}$ are induced.

\section{Conclusions}
\label{sec:Conclusions}
Taking into account recent discussions on composite indicators and drawing on central concepts from the current debate in AI, this paper proposes a methodology for post-hoc explanation and ante-hoc interpretability of composite indicators.

Our approach is based on decision rules expressed in natural language, which enhance understandability and transparency. These rules address several critical issues in the construction of composite indicators - most notably, they avoid the use of weights and complex aggregation procedures, which are often sources of controversy. Our contribution goes beyond a straightforward application of DRSA by introducing an original procedure for selecting a minimal subset of decision rules that explain the multicriteria assessment without contradictions. Although the DRSA literature has extensively studied decision rule induction, it has not tackled the issue of identifying a consistent and minimal subset of rules from the full set of possible rules derived from classification data. By consistency, we refer to avoiding situations where the same unit is simultaneously assigned to at least a higher class (e.g., good) and at most a lower class (e.g., medium), which would be clearly unacceptable.

Beyond offering a new methodology for explaining or constructing composite indicators, our decision-rule-based approach introduces a novel conception of the composite indicator: one that, in addition to providing an overall assessment of the evaluated units, is also expected to offer clear and transparent justifications for those evaluations. This enables all interested stakeholders to assess the soundness and fairness of the outcomes, which is essential for the critical evaluation of composite indicators.

Moreover, the proposed methodology naturally extends to continuous composite indicators, which are common in practice. This is made possible by an efficient new algorithm that induces all minimal decision rules in a single run for a non-predefined number of classes, allowing each distinct score assigned by the composite indicator to be treated as an ordered class. Importantly, the approach can also handle datasets with missing values, enabling the construction of composite indicators in such cases in a natural and transparent way.



Having introduced the methodological foundations and advantages of decision-rule-based composite indicators, it is instructive to situate this approach within the broader interpretability literature in both machine learning and MCDA, to highlight its theoretical underpinnings and potential contributions. Ante-hoc interpretability, which refers to the inherent transparency of a method, is naturally supported in MCDA through explicit criteria, weighting schemes, and preference structures. Our decision rules embody this principle by providing understandable and transparent reasoning for each composite indicator outcome. Post-hoc interpretability, which focuses on generating explanations after the assessment, can likewise enrich MCDA practices by clarifying the contribution of individual criteria, assessing sensitivity, and highlighting robustness. By integrating both perspectives, MCDA not only benefits from established interpretability frameworks in AI but also contributes its own value-driven and participatory perspective, emphasizing stakeholder understanding and the fairness of outcomes. 

A key limitation of the proposed methodology concerns its dependence on the input data from which decision rules are induced. The validity and generalizability of the resulting rules are closely tied to the representativeness and quality of the scored or classified units used as input. When the available data are sparse, unbalanced, or not reflective of the diversity of real-world cases, the induced rules may fail to capture the full rationale underlying the composite indicator. Another limitation relates to scalability and complexity. While rule-based models are generally valued for their interpretability, in high-dimensional problems or in contexts with substantial heterogeneity, the number and length of the generated rules may grow considerably. Such rule proliferation can undermine the very goal of transparency, making the explanations difficult to read and interpret for decision makers. These aspects suggest that careful attention should be paid both to the quality of the input data and to strategies for simplifying or pruning the rule sets in order to preserve their intelligibility. In this perspective, observe that, although a large number of rules may be generated for continuous indicators, explainability is preserved by presenting only the rules satisfied by the unit of interest.

We believe that the proposed methodology can make a significant contribution to promoting the sound and fair use of composite indicators. Nevertheless, the construction of decision-rule-based composite indicators should be further tested in real-world applications to assess how well the approach adapts to the specific needs of different domains. In particular, it is important to evaluate this methodology on datasets of lower quality, such as small samples or imprecise and ill-determined information. A key advantage of DRSA is that, unlike many statistical methods, it does not impose requirements such as minimum sample size or specific probability distributions (e.g., normality). One of its distinguishing features is its capacity to handle inconsistent classification data, in which the dominance principle is violated. A violation of this principle occurs when a unit that dominates another unit on the considered elementary indicators is assigned a lower class or score. Decision rules induced from such inconsistent data either have ambiguous class or score assignments or are univocal but probabilistic \citep{BSSz_2011}. Moreover, DRSA can be directly applied to ordinal qualitative data, without translating them into quantitative values, as well as to mixed datasets combining numerical and qualitative variables. Finally, as demonstrated in this paper, DRSA can also handle datasets with missing values, offering an advantage over aggregation procedures that usually require ad hoc treatments for missing data.

Future developments of this approach may also involve the integration of unconventional data sources. Recent advances in natural language processing and computer vision, for instance, enable the extraction of structured quantitative variables from unstructured data such as text corpora (e.g., sentiment or thematic indicators from social media or survey responses) and images (e.g., land-use or pollution proxies from satellite imagery). Once standardized, such variables could be incorporated into composite indicators, thereby expanding their scope beyond traditional numerical datasets. Furthermore, the proposed framework has potential applications in domains characterized by the need to synthesize massive and heterogeneous information streams, such as smart cities, healthcare, and environmental monitoring, where AI and data analytics play a crucial role in generating and processing input variables. In this way, composite indicators can serve as an effective tool for bridging statistical methodologies with emerging data-intensive fields, thereby enhancing their practical value for policy-making and decision support. This further highlights the versatility of our framework, which accommodates continuous composite indicators, preserves explainability on demand, and can handle datasets with missing values, making it applicable to diverse and complex real-world scenarios.

\section*{Acknowledgments} 
\noindent This study was funded by the European Union - NextGenerationEU, in the framework of the GRINS - Growing Resilient, INclusive and Sustainable project (GRINS PE00000018 CUP E63C22002120006). The views and opinions expressed are solely those of the authors and do not necessarily reflect those of the European Union, nor can the European Union be held responsible for them.  The research of Roman S{\l}owi{\'n}ski was funded by the National Science Centre, Poland, under the MAESTRO programme (grant number 2023/50/A/HS4/00499).


\bibliographystyle{plainnat}
\bibliography{Full_bibliography} 

@inproceedings{greco2000algorithm,
  title={An algorithm for induction of decision rules consistent with the dominance principle},
  author={Greco, S. and Matarazzo, B. and S{\l}owi\'{n}ski, R. and Stefanowski, J.},
  booktitle={International conference on rough sets and current trends in computing},
  pages={304--313},
  year={2000},
  organization={Springer}
}

@misc{oecd_pisa2022,
  author       = {{OECD}},
  title        = {PISA 2022 Database [data set]},
  year         = {2024},
  publisher    = {OECD Publishing},
  howpublished = {\url{https://doi.org/10.1787/52df09c4-en}},
  doi          = {10.1787/52df09c4-en},
  note         = {Accessed: \today}
}

@inproceedings{greco1999handling,
  title={Handling missing values in rough set analysis of multi-attribute and multi-criteria decision problems},
  author={Greco, S. and Matarazzo, B. and S{\l}owi\'{n}ski, R.},
  booktitle={International Workshop on Rough Sets, Fuzzy Sets, Data Mining, and Granular-Soft Computing},
  pages={146--157},
  year={1999},
  organization={Springer}
}

@book{joint2008handbook,
  title={Handbook on constructing composite indicators: methodology and user guide},
  author={Joint Research Centre},
  year={2008},
  publisher={OECD publishing}
}

@article{donders2006gentle,
  title={A gentle introduction to imputation of missing values},
  author={Donders, A. and {Van Der Heijden}, G. and Stijnen, T. and Moons, K.},
  journal={Journal of Clinical Epidemiology},
  volume={59},
  number={10},
  pages={1087--1091},
  year={2006},
  publisher={Elsevier}
}

@article{BSSz_2011,
  title={Sequential covering rule induction algorithm for variable consistency rough set approaches},
  author={B{\l}aszczy\'{n}ski, J. and S{\l}owi\'{n}ski, R. and Szel\k{a}g, M.},
  journal={Information Sciences},
  volume={181},
  pages={987--1002},
  year={2011},
  publisher={Elsevier}
}

@article{morais2011evaluation,
  title={{Evaluation of performance of European cities with the aim to promote quality of life improvements}},
  author={Morais, P. and Camanho, A.S.},
  journal={Omega},
  volume={39},
  number={4},
  pages={398--409},
  year={2011},
  publisher={Elsevier}
}

@article{ruiz2020mrp,
  title={{MRP-WSCI: Multiple reference point based weak and strong composite indicators}},
  author={Ruiz, F. and El Gibari, S. and Cabello, J.M. and G{\'o}mez, T.},
  journal={Omega},
  volume={95},
  pages={102060},
  year={2020},
  publisher={Elsevier}
}

@article{despotis2005measuring,
  title={{Measuring human development via data envelopment analysis: the case of Asia and the Pacific}},
  author={Despotis, D.K.},
  journal={Omega},
  volume={33},
  number={5},
  pages={385--390},
  year={2005},
  publisher={Elsevier}
}

@article{greco2024fifty,
  title={Fifty years of multiple criteria decision analysis: From classical methods to robust ordinal regression},
  author={Greco, S. and S{\l}owi{\'n}ski, R. and Wallenius, J.},
  journal={European Journal of Operational Research}, 
  volume={323},
  pages={351--377},
  year={2025},
  publisher={Elsevier}
}

@article{kuc2020quantitative,
  title={Quantitative storytelling in the making of a composite indicator},
  author={Kuc-Czarnecka, M. and Lo Piano, S. and Saltelli, A.},
  journal={Social Indicators Research},
  volume={149},
  number={3},
  pages={775--802},
  year={2020},
  publisher={Springer}
}

@article{cherchye2001using,
  title={Using data envelopment analysis to assess macroeconomic policy performance},
  author={Cherchye, L.},
  journal={Applied Economics},
  volume={33},
  number={3},
  pages={407--416},
  year={2001},
  publisher={Taylor \& Francis}
}

@article{munda2009noncompensatory,
  title={Noncompensatory/nonlinear composite indicators for ranking countries: a defensible setting},
  author={Munda, G. and Nardo, M.},
  journal={Applied Economics},
  volume={41},
  number={12},
  pages={1513--1523},
  year={2009},
  publisher={Taylor \& Francis}
}

@book{hazelkorn2015rankings,
  title={Rankings and the reshaping of higher education: The battle for world-class excellence},
  author={Hazelkorn, E.},
  year={2015},
  publisher={Springer}
}

@article{haakenstad2022assessing,
  title={Assessing performance of the Healthcare Access and Quality Index, overall and by select age groups, for 204 countries and territories, 1990--2019: a systematic analysis from the Global Burden of Disease Study 2019},
  author={Haakenstad, A.  and others},
  journal={The Lancet global health},
  volume={10},
  number={12},
  pages={e1715--e1743},
  year={2022},
  publisher={Elsevier}
}

@article{barrington2018measuring,
  title={Measuring progress and well-being: A comparative review of indicators},
  author={Barrington-Leigh, Ch. and Escande, A.},
  journal={Social Indicators Research},
  volume={135},
  pages={893--925},
  year={2018},
  publisher={Springer}
}

@book{schwab2019global,
  title={The global competitiveness report 2019},
  author={Schwab, K.},
  year={2019},
  organization={World Economic Forum}
}

@article{pawlak1982rough,
  title={Rough sets},
  author={Pawlak, Z.},
  journal={International Journal of Computer \& Information Sciences},
  volume={11},
  pages={341--356},
  year={1982},
  publisher={Springer}
}

@article{ali2023explainable,
  title={{Explainable Artificial Intelligence (XAI): What we know and what is left to attain Trustworthy Artificial Intelligence}},
  author={Ali, S. and Abuhmed, T. and El-Sappagh, S. and Muhammad, K. and Alonso-Moral, J. M and Confalonieri, R. and Guidotti, R. and Del Ser, J. and D{\'\i}az-Rodr{\'\i}guez, N. and Herrera, F.},
  journal={Information Fusion},
  volume={99},
  pages={101805},
  year={2023},
  publisher={Elsevier}
}

@article{rudin2019stop,
  title={Stop explaining black box machine learning models for high stakes decisions and use interpretable models instead},
  author={Rudin, C.},
  journal={Nature machine intelligence},
  volume={1},
  number={5},
  pages={206--215},
  year={2019},
  publisher={Nature Publishing Group UK London}
}

@article{christoph2020interpretable,
  title={Interpretable machine learning: A guide for making black box models explainable},
  author={Molnar, Ch.},
  year={2020},
  publisher={Leanpub}
}

@book{saltelli2023politics,
  title={The politics of modelling: Numbers between science and policy},
  author={Saltelli, A. and Di Fiore, M.},
  year={2023},
  publisher={Oxford University Press}
}

@article{saltelli2020five,  
title={Five ways to ensure that models serve society: a manifesto},
  author={Saltelli, A. and Bammer, G. and Bruno, I. and Charters, E. and Di Fiore, M. and Didier, E. and Nelson Espeland, W. and Kay, J. and Lo Piano, S. and Mayo, D. and others},
  journal={Nature},
  volume={582},
  number={7813},
  pages={482--484},
  year={2020},
  publisher={Nature Publishing Group UK London}
}

@article{BanaCostaVansnick1994,
title={{MACBETH} -- {a}n interactive path towards the construction of cardinal value functions},
author={Bana e Costa, C. and Vansnick, J.-C.},
journal={International Transactions in Operational Research},
volume={1},
number={4},
pages={489--500},
year={1994},
publisher={Elsevier}
}

@BOOK{BeltonStewart2002,
author={Belton, V. and Stewart, T.J.},
title="Multiple criteria decision analysis: an integrated approach",
publisher = {Springer},
booktitle = {Multiple criteria decision analysis: an integrated approach},
year = "2002",
}

@article{CinelliEtAl2020,
title={{How to support the application of Multiple Criteria Decision Analysis? Let us start with a comprehensive taxonomy}},
author={Cinelli, M. and Kadzi\'{n}ski, M. and Gonzalez, M. and S{\l}owi\'{n}ski, R.},
journal={Omega},
volume={102261},
year={2020},
publisher={Elsevier}
}

@article{FigueiraGrecoRoy2022,
title={{\textsc{ELECTRE}-Score: A first outranking based method for scoring actions}},
author={Figueira, J.R. and Greco, S. and Roy, B.},
journal={European Journal of Operational Research},
volume={297},
number={3},
pages={986--1005},
year={2022},
publisher={Elsevier}
}

@ARTICLE{FigueiraRoy2002,
AUTHOR="Figueira, J.R. and Roy, B.",
TITLE="Determining the weights of criteria in the {ELECTRE} type methods with a revised {S}imos' procedure",
JOURNAL="European Journal of Operational Research",
VOLUME="139",
YEAR="2002",
PAGES="317-326",
}

@BOOK{GrecoEhrgottFigueira2016,
author={Greco, S. and Ehrgott, M. and Figueira, J.R.},
title="Multiple Criteria Decision Analysis: State of the Art Surveys",
publisher="Springer, New York",
year="2016",
}

@article{GrecoEtAl2018,
title={{Stochastic multi-attribute acceptability analysis (SMAA): an application to the ranking of Italian regions}},
author={Greco, S. and Ishizaka, A. and Matarazzo, B. and Torrisi, G.},
journal={Regional Studies},
volume={52},
number={4},
pages={585--600},
year={2018},
publisher={Taylor \& Francis}
}

@article{GrecoEtAl2019,
title={{On the methodological framework of composite indices: A review of the issues of weighting, aggregation, and robustness}},
author={Greco, S. and Ishizaka, A. and Tasiou, M. and Torrisi, G.},
journal={Social Indicators Research},
year={2019},
volume={141},
number={1},
pages={61-94}, 
publisher={Springer}
}

@article{GrecoMatarazzoSlowinski2001,
title={Rough sets theory for multicriteria decision analysis},
author={Greco, S. and Matarazzo, B. and S{\l}owi\'{n}ski, R.},
journal={European Journal of Operational Research},
volume={129},
number={1},
pages={1--47},
year={2001},
publisher={Elsevier}
}

@article{NemeryLamboray2008,
title={Flow{S}ort: a flow-based sorting method with limiting or central profiles},
author={Nemery, P. and Lamboray, C.},
journal={{TOP}},
volume={16},
pages={90--113},
year={2008}
}

@BOOK{Pawlak1991,
AUTHOR="Pawlak, Z.",
TITLE="Rough Sets - Theoretical Aspects of Reasoning about Data",
PUBLISHER="Kluwer Academic, Dordrecht",
YEAR=1991
}

@article{Rezaei2015,
title={Best-worst multi-criteria decision-making method},
author={Rezaei, J.},
journal={Omega},
volume={53},
pages={49--57},
year={2015},
publisher={Elsevier}
}

@book{Roy1996,
author="Roy, B.",
title={{Multicriteria Methodology for Decision Aiding}},
publisher="Kluwer Academic Publishers",
address="Dordrecht",
year="1996",
}

@article{Saaty1977,
title={A scaling method for priorities in hierarchical structures},
author={Saaty, T.},
journal={Journal of Mathematical Psychology},
volume={15},
number={3},
pages={234-281},
year={1977}
}

@article{Yu1992,
title={Aide multicrit\`{e}re \`{a} la d\'{e}cision dans le cadre de la probl\'{e}matique du tri: m\'{e}thodes et applications. {P}h.{D}. thesis},
author={Yu, W.},
journal={{LAMSADE}},
volume={Universit\'{e} {P}aris {D}auphine, {P}aris},
year={1992}
}

@article{TeasdaleJennett1976,
  title = {Assessment and prognosis of coma after head injury},
  author = {Teasdale,  G. and Jennett,  B.},
  journal = {Acta Neurochirurgica},
  volume = {34},
  number = {1–4},
  pages = {45–55},
  year = {1976},
  publisher = {Springer Science and Business Media LLC},
}

@article{Emamat2022,
  title = {{Using ELECTRE-TRI and FlowSort methods in a stock portfolio selection context}},
  volume = {8},
  number = {1},
  pages={1-35},
  journal = {Financial Innovation},
  publisher = {Springer Science and Business Media LLC},
  author = {Emamat,  M.S.M.M. and Mota,  C.M.M. and Mehregan,  M.R. and Sadeghi Moghadam, M.R. and Nemery,  P.},
  year = {2022},
}

@article{chakraborty2023comprehensive,
  title={A comprehensive and systematic review of multi-criteria decision-making methods and applications in healthcare},
  author={Chakraborty, S. and Raut, R.D. and Rofin, T.M. and Chakraborty, S.},
  journal={Healthcare Analytics},
  volume={4},
  pages={100232},
  year={2023},
  publisher={Elsevier}
}

@article{greco2016measures,
  title={{Measures of rule interestingness in various perspectives of confirmation}},
  author={Greco, S. and S{\l}owi{\'n}ski, R. and Szcz{\c{e}}ch, I.},
  journal={Information Sciences},
  volume={346},
  pages={216--235},
  year={2016},
  publisher={Elsevier}
}

@article{CorrenteGrecoZappalà2025,
  title={{Deck of cards method for hierarchical, robust and stochastic ordinal regression}},
  author={Corrente, S. and Greco, S. and Zappalà, S.},
  journal={European Journal of Operational Research},
  year={2025},
  publisher={Elsevier},
  doi={10.1016/j.ejor.2025.05.025}
}

@article{BarbatiGrecoLami2024,
  title={{The deck-of-cards-based ordinal regression method and its application for the development of an ecovillage}},
  author={Barbati, M. and Greco, S. and Lami, I. M.},
  journal={European Journal of Operational Research},
  volume={319},
  number={3},
  pages={845--861},
  year={2024},
  publisher={Elsevier}
}

@misc{HDR,
  author       = {{HDR}},
  title        = {{Human Development Reports}},
  year         = {2025},
  url          = {https://hdr.undp.org/}
}
\clearpage
\appendix
\section{}
\label{sec:appendix_1}
\begin{longtable}
{cm{0.15\textwidth}|>{\centering}m{0.15\textwidth}>{\centering}m{0.15\textwidth}>{\centering}m{0.15\textwidth}>{\centering}m{0.15\textwidth}|>{\centering\arraybackslash}m{0.05\textwidth}}\\
\caption{Human Development Index (HDI) criteria and the corresponding class for each country (\href{https://hdr.undp.org/}{source})}
\label{tab:hdi_index}\\
\toprule
ID & Unit & Life expectancy at birth (years) & Expected years of schooling (years) & Mean years of schooling (years) & Gross national income (GNI) per capita (2017 PPP \$) & Class \\ 
\midrule
\endfirsthead
\multicolumn{7}{c}{Table \ref{tab:hdi_index} (Continued)} \\
\toprule
ID & Unit & Life expectancy at birth (years) & Expected years of schooling (years) & Mean years of schooling (years) & Gross national income (GNI) per capita (2017 PPP \$) & Class \\ 
\midrule
\endhead
1 & Switzerland & 84.255 & 16.5837 & 13.9041 & 69432.7867 & 4 \\ 
2 & Norway & 83.393 & 18.6385 & 13.0623 & 69189.7617 & 4 \\ 
3 & Iceland & 82.815 & 19.1067 & 13.7672 & 54688.3792 & 4 \\ 
4 & Hong Kong, China (SAR) & 84.315 & 17.8496 & 12.3478 & 62485.5052 & 4 \\ 
5 & Denmark & 81.882 & 18.774 & 12.9605 & 62018.9569 & 4 \\ 
6 & Sweden & 83.505 & 19.0368 & 12.6737 & 56995.848 & 4 \\ 
7 & Germany & 80.989 & 17.3433 & 14.2559 & 55340.1972 & 4 \\ 
8 & Ireland & 82.716 & 19.1261 & 11.657 & 87467.5139 & 4 \\ 
9 & Singapore & 84.133 & 16.9027 & 11.9249 & 88761.1456 & 4 \\ 
10 & Australia & 83.579 & 21.08 & 12.7268 & 49257.1352 & 4 \\ 
11 & Netherlands & 82.451 & 18.5835 & 12.5816 & 57278.3101 & 4 \\ 
12 & Belgium & 82.293 & 18.9457 & 12.5286 & 53644.0385 & 4 \\ 
13 & Finland & 82.351 & 19.2286 & 12.9292 & 49522.0985 & 4 \\ 
14 & Liechtenstein & 84.656 & 15.4664 & 12.3512 & 146673.2415 & 4 \\ 
15 & United Kingdom & 82.156 & 17.6343 & 13.4061 & 46623.9027 & 4 \\ 
16 & New Zealand & 83.006 & 19.6823 & 12.9447 & 43665.4985 & 4 \\ 
17 & United Arab Emirates & 79.196 & 17.2081 & 12.7738 & 74103.7149 & 4 \\ 
18 & Canada & 82.847 & 15.9607 & 13.8688 & 48444.3932 & 4 \\ 
19 & Korea (Republic of) & 84.024 & 16.5096 & 12.6141 & 46026.4542 & 4 \\ 
20 & Luxembourg & 82.591 & 14.1971 & 12.9601 & 78554.2364 & 4 \\ 
21 & United States & 78.203 & 16.4127 & 13.5755 & 65564.938 & 4 \\ 
22 & Austria & 82.412 & 16.3675 & 12.3057 & 56529.6633 & 4 \\ 
23 & Slovenia & 82.133 & 17.4066 & 12.8802 & 41586.8985 & 4 \\ 
24 & Japan & 84.82 & 15.4563 & 12.6692 & 43643.8642 & 4 \\ 
25 & Israel & 82.601 & 15.0298 & 13.4428 & 43588.2573 & 4 \\ 
26 & Malta & 83.704 & 15.8613 & 12.2103 & 44464.0306 & 4 \\ 
27 & Spain & 83.912 & 17.8091 & 10.6054 & 40043.3377 & 4 \\ 
28 & France & 83.229 & 15.9876 & 11.6886 & 47378.743 & 4 \\ 
29 & Cyprus & 81.889 & 16.2431 & 12.4402 & 40136.8945 & 4 \\ 
30 & Italy & 84.057 & 16.6581 & 10.7401 & 44284.1574 & 4 \\ 
31 & Estonia & 79.155 & 15.943 & 13.5484 & 37151.6338 & 4 \\ 
32 & Czechia & 78.129 & 16.3473 & 12.9161 & 39944.6668 & 4 \\ 
33 & Greece & 80.614 & 20.0264 & 11.4085 & 31381.667 & 4 \\ 
34 & Bahrain & 79.246 & 16.2997 & 11.0466 & 48731.4456 & 4 \\ 
35 & Andorra & 83.552 & 12.7838 & 11.6134 & 54233.4495 & 4 \\ 
36 & Poland & 76.996 & 15.9345 & 13.1648 & 35150.9519 & 4 \\ 
37 & Latvia & 75.927 & 16.5561 & 13.3333 & 32082.981 & 4 \\ 
38 & Lithuania & 74.293 & 16.3998 & 13.4981 & 38131.2359 & 4 \\ 
39 & Croatia & 79.236 & 15.5727 & 12.3287 & 34323.8066 & 4 \\ 
40 & Qatar & 81.559 & 13.2643 & 10.1317 & 95944.3775 & 4 \\ 
41 & Saudi Arabia & 77.905 & 15.172 & 11.3105 & 50620.4371 & 4 \\ 
42 & Portugal & 82.24 & 16.8214 & 9.5759 & 35314.9983 & 4 \\ 
43 & San Marino & 83.433 & 12.403 & 10.5207 & 57686.5433 & 4 \\ 
44 & Chile & 79.519 & 16.7679 & 11.1114 & 24430.9959 & 4 \\ 
45 & Slovakia & 75.33 & 14.7215 & 13.0072 & 32171.2461 & 4 \\ 
46 & Türkiye & 78.475 & 19.6849 & 8.8113 & 32833.5351 & 4 \\ 
47 & Hungary & 74.958 & 15.0733 & 12.2496 & 34195.5406 & 4 \\ 
48 & Argentina & 76.064 & 18.9795 & 11.1441 & 22047.9713 & 4 \\ 
49 & Kuwait & 80.264 & 15.6911 & 7.4417 & 56729.1815 & 4 \\ 
50 & Montenegro & 76.845 & 15.0848 & 12.6162 & 22513.2631 & 4 \\ 
51 & Saint Kitts and Nevis & 72.027 & 18.407 & 10.8494 & 28441.6807 & 4 \\ 
52 & Uruguay & 78 & 17.3518 & 9.0582 & 22206.9903 & 4 \\ 
53 & Romania & 74.117 & 14.5075 & 11.3602 & 31641.3837 & 4 \\ 
54 & Antigua and Barbuda & 79.236 & 15.5123 & 10.5462 & 18783.9654 & 4 \\ 
55 & Brunei Darussalam & 74.551 & 13.6986 & 9.22 & 59245.6349 & 4 \\ 
56 & Russian Federation & 70.116 & 15.6617 & 12.4115 & 26991.8496 & 4 \\ 
57 & Bahamas & 74.358 & 11.8914 & 12.7316 & 32534.8878 & 4 \\ 
58 & Panama & 76.826 & 13.2138 & 10.687 & 32029.3603 & 4 \\ 
59 & Oman & 73.935 & 12.964 & 11.8927 & 32967.4383 & 4 \\ 
60 & Georgia & 71.587 & 16.7268 & 12.7022 & 15952.0245 & 4 \\ 
61 & Trinidad and Tobago & 74.708 & 14.1011 & 11.738 & 22473.0388 & 4 \\ 
62 & Barbados & 77.706 & 16.5307 & 9.8921 & 14810.2378 & 4 \\ 
63 & Malaysia & 76.26 & 12.9325 & 10.7483 & 27295.4122 & 4 \\ 
64 & Costa Rica & 77.32 & 16.092 & 8.8041 & 20248.3803 & 4 \\ 
65 & Serbia & 74.137 & 14.5044 & 11.5048 & 19494.0089 & 4 \\ 
66 & Thailand & 79.68 & 15.5813 & 8.8346 & 16886.5111 & 4 \\ 
67 & Kazakhstan & 69.489 & 14.8163 & 12.4334 & 22586.7989 & 4 \\ 
68 & Seychelles & 71.738 & 13.8918 & 11.1999 & 28385.7481 & 4 \\ 
69 & Belarus & 73.246 & 13.9815 & 12.2412 & 18425.0106 & 4 \\ 
70 & Bulgaria & 71.528 & 13.868 & 11.4132 & 25920.8037 & 3 \\ 
71 & Palau & 65.362 & 17.2199 & 13.0479 & 19343.812 & 3 \\ 
72 & Mauritius & 73.975 & 14.6101 & 9.992 & 23251.6207 & 3 \\ 
73 & Grenada & 75.335 & 16.5823 & 9.858 & 13593.2496 & 3 \\ 
74 & Albania & 76.833 & 14.4875 & 10.1211 & 15293.3265 & 3 \\ 
75 & China & 78.587 & 15.2179 & 8.1069 & 18024.8875 & 3 \\ 
76 & Armenia & 73.372 & 14.4056 & 11.3303 & 15388.2998 & 3 \\ 
77 & Mexico & 74.832 & 14.5053 & 9.2211 & 19138.0073 & 3 \\ 
78 & Iran (Islamic Republic of) & 74.556 & 14.1172 & 10.7462 & 14770.3231 & 3 \\ 
79 & Sri Lanka & 76.61 & 13.6418 & 11.2496 & 11899.4985 & 3 \\ 
80 & Bosnia and Herzegovina & 75.293 & 13.2771 & 10.5363 & 16571.4127 & 3 \\ 
81 & Saint Vincent and the Grenadines & 68.972 & 16.253 & 10.9963 & 14049.148 & 3 \\ 
82 & Dominican Republic & 74.17 & 13.557 & 9.1505 & 18653.2684 & 3 \\ 
83 & Ecuador & 77.894 & 14.8648 & 8.9698 & 10693.2348 & 3 \\ 
84 & North Macedonia & 73.888 & 13.0048 & 10.2282 & 16395.7517 & 3 \\ 
85 & Cuba & 78.155 & 14.4693 & 10.5481 & 7953.4484 & 3 \\ 
86 & Moldova (Republic of) & 68.621 & 14.906 & 11.8338 & 12963.6183 & 3 \\ 
87 & Maldives & 80.839 & 12.178 & 7.7624 & 18846.7922 & 3 \\ 
88 & Peru & 73.385 & 14.8049 & 10.017 & 11916.3595 & 3 \\ 
89 & Azerbaijan & 73.488 & 12.7106 & 10.5561 & 15018.0544 & 3 \\ 
90 & Brazil & 73.425 & 15.5788 & 8.2783 & 14615.8924 & 3 \\ 
91 & Colombia & 73.659 & 14.4375 & 8.8629 & 15013.9272 & 3 \\ 
92 & Libya & 72.151 & 13.9771 & 7.7786 & 19751.5676 & 3 \\ 
93 & Algeria & 77.129 & 15.4879 & 6.9874 & 10978.4057 & 3 \\ 
94 & Turkmenistan & 69.41 & 13.2406 & 11.143 & 12859.8741 & 3 \\ 
95 & Guyana & 65.989 & 13.0221 & 8.6274 & 35782.9096 & 3 \\ 
96 & Mongolia & 72.667 & 14.5172 & 9.4237 & 10350.8614 & 3 \\ 
97 & Dominica & 72.981 & 13.5505 & 9.1863 & 12467.8555 & 3 \\ 
98 & Tonga & 71.27 & 16.2875 & 10.8803 & 6360.1789 & 3 \\ 
99 & Jordan & 74.215 & 12.6266 & 10.4467 & 9294.8015 & 3 \\ 
100 & Ukraine & 68.564 & 13.328 & 11.1233 & 11416.2168 & 3 \\ 
101 & Tunisia & 74.263 & 14.6195 & 7.9534 & 10296.6496 & 3 \\ 
102 & Marshall Islands & 65.146 & 16.3861 & 12.8184 & 6855.2346 & 3 \\ 
103 & Paraguay & 70.475 & 13.9106 & 8.8618 & 13161.0752 & 3 \\ 
104 & Fiji & 68.312 & 13.8487 & 10.371 & 11233.6588 & 3 \\ 
105 & Egypt & 70.159 & 12.9118 & 9.8481 & 12360.8163 & 3 \\ 
106 & Uzbekistan & 71.674 & 11.9929 & 11.9111 & 8055.9103 & 3 \\ 
107 & Viet Nam & 74.58 & 13.0529 & 8.4553 & 10813.9827 & 3 \\ 
108 & Saint Lucia & 71.294 & 12.7103 & 8.577 & 14778.3463 & 3 \\ 
109 & Lebanon & 74.416 & 12.0511 & 8.603 & 12313.4226 & 3 \\ 
110 & South Africa & 61.48 & 14.2636 & 11.607 & 13185.5642 & 3 \\ 
111 & Palestine, State of & 73.444 & 13.1547 & 9.9385 & 6936.2587 & 3 \\ 
112 & Indonesia & 68.25 & 13.9757 & 8.5565 & 12045.5654 & 3 \\ 
113 & Philippines & 72.187 & 12.7834 & 8.9714 & 9058.8406 & 3 \\ 
114 & Botswana & 65.913 & 11.4257 & 10.42 & 14841.5784 & 3 \\ 
115 & Jamaica & 70.629 & 12.4598 & 9.2482 & 9694.5228 & 3 \\ 
116 & Samoa & 72.598 & 12.4364 & 11.3674 & 4970.2321 & 3 \\ 
117 & Kyrgyzstan & 70.484 & 12.9949 & 11.9613 & 4781.692 & 3 \\ 
118 & Belize & 70.962 & 12.4266 & 8.8476 & 9242.0823 & 3 \\ 
119 & Venezuela (Bolivarian Republic of) & 71.105 & 13.5021 & 9.6282 & 6184.136 & 2 \\ 
120 & Bolivia (Plurinational State of) & 64.928 & 15.0224 & 9.8278 & 7987.8422 & 2 \\ 
121 & Morocco & 74.973 & 14.5944 & 6.05 & 7954.5225 & 2 \\ 
122 & Nauru & 64.014 & 12.5609 & 9.248 & 14938.5566 & 2 \\ 
123 & Gabon & 65.694 & 12.4343 & 9.55 & 11194.2188 & 2 \\ 
124 & Suriname & 70.289 & 10.9591 & 8.3886 & 12309.9966 & 2 \\ 
125 & Bhutan & 72.229 & 13.0645 & 5.8358 & 10624.8739 & 2 \\ 
126 & Tajikistan & 71.288 & 10.861 & 11.2896 & 4807.2139 & 2 \\ 
127 & El Salvador & 71.475 & 11.9099 & 7.1509 & 8886.1735 & 2 \\ 
128 & Iraq & 71.336 & 12.2228 & 6.8119 & 9091.8654 & 2 \\ 
129 & Bangladesh & 73.698 & 11.9463 & 7.3791 & 6511.1222 & 2 \\ 
130 & Nicaragua & 74.615 & 12.5835 & 7.26 & 5426.5062 & 2 \\ 
131 & Cabo Verde & 74.722 & 11.5307 & 6.09 & 7601.0854 & 2 \\ 
132 & Tuvalu & 64.854 & 12.1016 & 10.6446 & 4754.4537 & 2 \\ 
133 & Equatorial Guinea & 61.19 & 12.1313 & 8.2813 & 10662.6592 & 2 \\ 
134 & India & 67.744 & 12.5837 & 6.5707 & 6950.5268 & 2 \\ 
135 & Micronesia (Federated States of) & 70.925 & 12.615 & 7.3325 & 3709.1549 & 2 \\ 
136 & Guatemala & 68.674 & 10.7672 & 5.6686 & 8996.4158 & 2 \\ 
137 & Kiribati & 67.661 & 11.803 & 9.1286 & 3440.4167 & 2 \\ 
138 & Honduras & 70.728 & 9.9617 & 7.2962 & 5271.605 & 2 \\ 
139 & Lao People's Democratic Republic & 68.999 & 10.1875 & 5.947 & 7744.8373 & 2 \\ 
140 & Vanuatu & 70.492 & 11.8116 & 7.1832 & 3243.9805 & 2 \\ 
141 & Sao Tome and Principe & 68.794 & 12.6637 & 5.9158 & 4054.1069 & 2 \\ 
142 & Eswatini (Kingdom of) & 56.36 & 14.9269 & 5.73 & 8391.8551 & 2 \\ 
143 & Namibia & 58.059 & 11.76 & 7.2435 & 9200.0266 & 2 \\ 
144 & Myanmar & 67.256 & 12.0609 & 6.5182 & 4037.7051 & 2 \\ 
145 & Ghana & 63.945 & 11.5863 & 6.4323 & 5380.2715 & 2 \\ 
146 & Kenya & 62.055 & 11.3845 & 7.6864 & 4807.7163 & 2 \\ 
147 & Nepal & 70.484 & 12.6445 & 4.4856 & 4025.5547 & 2 \\ 
148 & Cambodia & 69.896 & 11.5609 & 5.1971 & 4291.1132 & 2 \\ 
149 & Congo & 63.053 & 12.4165 & 8.2506 & 2902.808 & 2 \\ 
150 & Angola & 61.929 & 12.1676 & 5.8443 & 5327.7883 & 2 \\ 
151 & Cameroon & 60.958 & 13.381 & 6.5374 & 3681.47 & 2 \\ 
152 & Comoros & 63.68 & 13.0377 & 6.2091 & 3260.5555 & 2 \\ 
153 & Zambia & 61.803 & 11.018 & 7.2849 & 3157.3586 & 2 \\ 
154 & Papua New Guinea & 65.958 & 11.1274 & 4.9336 & 3710.3269 & 2 \\ 
155 & Timor-Leste & 69.056 & 13.2418 & 6.0201 & 1629.1604 & 2 \\ 
156 & Solomon Islands & 70.742 & 10.3007 & 5.8797 & 2273.3111 & 2 \\ 
157 & Syrian Arab Republic & 72.3 & 7.4188 & 5.7368 & 3594.1068 & 2 \\ 
158 & Haiti & 63.728 & 11.1376 & 5.6095 & 2801.7125 & 2 \\ 
159 & Uganda & 63.638 & 11.4955 & 6.2355 & 2240.5854 & 2 \\ 
160 & Zimbabwe & 59.391 & 11.0256 & 8.8078 & 2078.9181 & 2 \\ 
161 & Nigeria & 53.633 & 10.5141 & 7.586 & 4754.8364 & 1 \\ 
162 & Rwanda & 67.129 & 11.3835 & 4.8783 & 2316.8093 & 1 \\ 
163 & Togo & 61.588 & 12.9624 & 5.5903 & 2214.2336 & 1 \\ 
164 & Mauritania & 64.691 & 8.0519 & 4.7541 & 5343.5641 & 1 \\ 
165 & Pakistan & 66.431 & 7.8951 & 4.4223 & 5374.2704 & 1 \\ 
166 & Côte d'Ivoire & 58.916 & 10.1017 & 4.237 & 5376.3961 & 1 \\ 
167 & Tanzania (United Republic of) & 66.782 & 8.5857 & 5.64 & 2578.156 & 1 \\ 
168 & Lesotho & 53.036 & 11.0705 & 7.5451 & 2708.7315 & 1 \\ 
169 & Senegal & 67.913 & 9.1446 & 2.8874 & 3463.8061 & 1 \\ 
170 & Sudan & 65.578 & 8.5017 & 3.87 & 3514.7659 & 1 \\ 
171 & Djibouti & 62.859 & 8.0284 & 3.89 & 4874.5183 & 1 \\ 
172 & Malawi & 62.898 & 11.5049 & 5.2437 & 1432.4728 & 1 \\ 
173 & Benin & 59.954 & 10.3026 & 3.1373 & 3406.0729 & 1 \\ 
174 & Gambia & 62.906 & 8.9779 & 4.5124 & 2089.6268 & 1 \\ 
175 & Eritrea & 66.604 & 7.3188 & 5.0914 & 1957.0465 & 1 \\ 
176 & Ethiopia & 65.645 & 9.9387 & 2.3909 & 2368.7584 & 1 \\ 
177 & Liberia & 61.1 & 10.4659 & 5.3181 & 1330.4199 & 1 \\ 
178 & Madagascar & 65.23 & 9.2134 & 4.5925 & 1463.5479 & 1 \\ 
179 & Guinea-Bissau & 59.861 & 10.5317 & 3.673 & 1879.8536 & 1 \\ 
180 & Congo (Democratic Republic of the) & 59.743 & 9.5795 & 7.2062 & 1080.1383 & 1 \\ 
181 & Guinea & 58.985 & 10.2001 & 2.3591 & 2404.1658 & 1 \\ 
182 & Afghanistan & 62.879 & 10.7054 & 2.5148 & 1335.2057 & 1 \\ 
183 & Mozambique & 59.625 & 10.7253 & 3.8601 & 1219.2414 & 1 \\ 
184 & Sierra Leone & 60.411 & 9.0163 & 3.5194 & 1612.6665 & 1 \\ 
185 & Burkina Faso & 59.766 & 8.0916 & 2.3187 & 2036.9972 & 1 \\ 
186 & Yemen & 63.72 & 7.9353 & 2.7768 & 1105.7634 & 1 \\ 
187 & Burundi & 61.977 & 9.9664 & 3.3052 & 712.026 & 1 \\ 
188 & Mali & 59.417 & 7.0372 & 1.6267 & 2043.6724 & 1 \\ 
189 & Chad & 52.997 & 8.1855 & 2.2828 & 1388.8975 & 1 \\ 
190 & Niger & 62.08 & 7.1861 & 1.3414 & 1283.3092 & 1 \\ 
191 & Central African Republic & 54.477 & 7.2859 & 3.9532 & 869.1124 & 1 \\ 
192 & South Sudan & 55.567 & 5.6348 & 5.7261 & 690.6608 & 1 \\ 
193 & Somalia & 56.107 & 7.6472 & 1.9003 & 1072.2015 & 1 \\ 
\bottomrule
\end{longtable}
\begin{longtable}
{lm{0.18\textwidth}|>{\centering}m{0.06\textwidth}>{\centering}m{0.05\textwidth}>{\centering}m{0.1\textwidth}>{\centering}m{0.06\textwidth}>{\centering}m{0.07\textwidth}>{\centering}m{0.08\textwidth}>{\centering}m{0.05\textwidth}>{\centering\arraybackslash}m{0.05\textwidth}}
\setlength{\tabcolsep}{0.5pt}\\ 
    \caption{Stock portfolio evaluations on eight criteria \citep{Emamat2022}} \label{tab:ptf_evaluation_criteria}\\
    \toprule
Unit & Name & Return ($\uparrow$)& Beta ($\uparrow$)& Net Profit Margin ($\uparrow$)& ROA ($\uparrow$)& ROE ($\uparrow$)& EPS ($\uparrow$)& P/E ($\uparrow$)& P/BV ($\uparrow$)\\ 
\midrule
\endfirsthead
\multicolumn{10}{c}{Table \ref{tab:ptf_evaluation_criteria} (Continued)} \\
\toprule
Unit & Name & Return ($\uparrow$)& Beta ($\uparrow$)& Net Profit Margin ($\uparrow$)& ROA ($\uparrow$)& ROE ($\uparrow$)& EPS ($\uparrow$)& P/E ($\uparrow$)& P/BV ($\uparrow$)\\
\midrule
\endhead
A1 & Azarab & 86.692 & 1.146 & 10.327 & 4.94 & 22.855 & 402 & 7.169 & 2.183 \\ 
A2 & Mobile Telecommunication Company of Iran & 23.666 & 0.322 & 31.513 & 21.389 & 29.745 & 9331.333 & 6.701 & 1.905 \\ 
A3 & Electric Khodro Shargh & 24.695 & 1.794 & 2.418 & 2.644 & 9.642 & 127.333 & 16.885 & 1.007 \\ 
A4 & Iran Transfo & 24.145 & 1.489 & 30.193 & 6.101 & 21.223 & 469.667 & 11.951 & 1.742 \\ 
A5 & Iran Khodro & 0.279 & 1.42 & -1.236 & -0.081 & -39.005 & -58.333 & 100 & 3.051 \\ 
A6 & Iran Yasa & 103.495 & 1.379 & 12.536 & 14.722 & 59.18 & 2603.667 & 7.738 & 2.981 \\ 
A7 & Bama & 35.747 & 1.005 & 52.491 & 34.533 & 44.535 & 1565.333 & 7.145 & 3.172 \\ 
A8 & Behceram & 121.102 & 1.103 & 2.073 & 0.758 & -28.316 & 58.667 & 69.733 & 20.786 \\ 
A9 & Pars Khodro & 31.643 & 1.146 & -10.159 & -6.553 & -47.223 & -505.333 & 250 & 0.828 \\ 
A10 & Kharg Petrochemical Company & 67.379 & 1.276 & 72.002 & 61.141 & 85.927 & 7527.333 & 4.853 & 4.228 \\ 
A11 & Shazand Petrochemical Company & 204.761 & 1.302 & 17.371 & 24.637 & 59.911 & 3313.333 & 8.091 & 3.121 \\ 
A12 & Shiraz Petrochemical Company & 97.489 & 0.593 & 45.938 & 14.533 & 45.104 & 1171.667 & 8.333 & 3.108 \\ 
A13 & Fanavaran Petrochemical Company & 100.181 & 1.391 & 64.639 & 43.411 & 68.788 & 4802.333 & 6.142 & 4.075 \\ 
A14 & Techinco & 46.36 & 0.939 & 14.166 & 11.607 & 26.443 & 743.667 & 14.388 & 1.911 \\ 
A15 & Behshahr Industrial Development Corp & 46.993 & 1.159 & 96.066 & 14.912 & 19.765 & 374 & 9.979 & 1.765 \\ 
A16 & Chadormalu Industrial Company & 54.133 & 1.186 & 62.478 & 41.887 & 56.698 & 1506.667 & 5.569 & 3.994 \\ 
A17 & North Drilling & 82.361 & 1.023 & 28.614 & 14.808 & 34.276 & 664 & 11.682 & 2.669 \\ 
A18 & Informatics Services Corporation & 61.004 & 0.471 & 144.01 & 29.242 & 50.169 & 3060.333 & 9.156 & 7.513 \\ 
A19 & Jaber Ebne Hayyan Pharmacy & 49.54 & 1.967 & 29.94 & 18.873 & 36.125 & 1005 & 10.14 & 2.547 \\ 
A20 & RAZAK Pharmaceutical Company & 115.57 & 0.964 & 31.935 & 23.115 & 60.48 & 2673.667 & 8.684 & 3.436 \\ 
A21 & Rayan Saipa & -9.145 & 1.141 & 37.453 & 4.916 & 25.06 & 628.333 & 2.726 & 1.694 \\ 
A22 & Zamyad & -4.476 & 1.953 & -1.696 & -1.009 & -4.421 & -46 & 130 & 0.773 \\ 
A23 & Saipa & -2.406 & 0.955 & -9.333 & -2.995 & -36.529 & -209.667 & 200 & 2.246 \\ 
A24 & Saipa Azin & 36.293 & 1.944 & -0.547 & -0.863 & -10.613 & -72 & 105 & 1.771 \\ 
A25 & Tehran Cement & 62.824 & 0.749 & 39.808 & 7.556 & 27.533 & 653.333 & 8.608 & 1.929 \\ 
A26 & Khazar Cement & 129.896 & 1.148 & 17.002 & 11.337 & 29.042 & 569.333 & 11.715 & 1.78 \\ 
A27 & Shahroud Cement & 78.941 & 1.499 & 29.087 & 17.412 & 36.373 & 755 & 8.886 & 1.81 \\ 
A28 & Gharb Cement & 148.506 & 0.617 & 32.868 & 24.777 & 45.046 & 1088.333 & 7.491 & 2.203 \\ 
A29 & Shahid Ghandi Corporation Complex & 11.437 & 0.95 & 3.559 & 2.428 & 11.377 & 188 & 10.856 & 1.306 \\ 
A30 & Kermanshah Petrochemical Industries Company & 105.115 & 0.1 & 64.46 & 27.252 & 58.391 & 1484 & 6.147 & 3.021 \\ 
A31 & Iran Refractories Company & 263.307 & 2.563 & 24.245 & 25.116 & 57.525 & 2080.667 & 9.479 & 2.521 \\ 
A32 & Faravari Mavad Madani Iran & -1.221 & 0.739 & 21.956 & 26.824 & 38.109 & 831.333 & 5.797 & 2.823 \\ 
A33 & Khouzestan Steel Company & 139.535 & 0.739 & 26.339 & 31.301 & 62.479 & 3465.667 & 2.71 & 4.585 \\ 
A34 & Mobarakeh Steel Company & 50.781 & 1.223 & 27.082 & 16.003 & 34.194 & 669.667 & 6.455 & 1.936 \\ 
A35 & Khorasan Steel Company & 75.047 & 0.463 & 28.002 & 24.719 & 36.402 & 737.667 & 13.066 & 4.583 \\ 
A36 & Calcimine & 32.199 & 1.017 & 49.163 & 32.073 & 41.901 & 1034.333 & 4.15 & 2.015 \\ 
A37 & Piazar Agro industry & 67.954 & 0.913 & 12.533 & 12.901 & 23.099 & 404 & 9.757 & 2.814 \\ 
A38 & Chemi Darou & 50.416 & 3.421 & 19.395 & 13.415 & 31.608 & 621.667 & 13.388 & 2.398 \\ 
A39 & Bahman Group & 12.917 & 0.881 & 43.592 & 10.996 & 19.705 & 503.667 & 4.888 & 0.843 \\ 
A40 & MAPNA Group & 67.118 & 1.652 & 26.533 & 3.878 & 14.733 & 545.333 & 15.937 & 1.758 \\ 
A41 & Golgohar Mining and Industrial Company & 79.686 & 1.124 & 61.344 & 36.966 & 54.492 & 1554.667 & 6.645 & 3.617 \\ 
A42 & Sahand Rubber Industries Company & 171.557 & 1.419 & 49.806 & 25.332 & 38.578 & 1207.667 & 7.635 & 1.895 \\ 
A43 & Telecommunication Company of Iran & 9.669 & 0.762 & 103.157 & 13.723 & 18.348 & 436.667 & 7.754 & 2.494 \\ 
A44 & Shahid Bahonar Copper Industries Company & 62.877 & 1.628 & 2.492 & 3.057 & 16.364 & 247.333 & 13.35 & 1.267 \\ 
A45 & Bafgh Mining & 150.822 & 0.788 & 37.259 & 31.046 & 36.09 & 2201.667 & 12.932 & 4.325 \\ 
A46 & Iran Zinc Mines Development & 23.018 & 0.87 & 101.85 & 23.686 & 27.024 & 570 & 4.323 & 1.715 \\ 
A47 & National Iranian Lead \& Zinc Company & 1.738 & 1.107 & 9.433 & 9.698 & 15.379 & 160 & 13.85 & 1.649 \\ 
A48 & National Iranian Copper Industry Company & 18.465 & 0.64 & 42.077 & 25.043 & 37.398 & 879.667 & 3.623 & 1.415 \\ 
A49 & Mehr Cam Pars & 31.448 & 1.6 & -2.41 & -2.643 & -16.397 & -114.333 & 180 & 1.554 \\ 
A50 & Behran Oil Company & 90.647 & 1.31 & 20.255 & 20.052 & 68.113 & 3219.333 & 13.216 & 5.363 \\ 
\bottomrule
\end{longtable}
\end{document}